\RequirePackage{fix-cm}
\documentclass{antgroup}
\usepackage[T1]{fontenc}
\usepackage{antgroup}
\usepackage{algorithm}
\usepackage{algpseudocode}
\usepackage{fvextra}
\usepackage{CJKutf8}
\usepackage{needspace}
\usepackage{float}
\usepackage{pifont}
\usepackage{colortbl}
\usepackage{amsmath,amsfonts,bm}

\def\eqref#1{equation~\ref{#1}}
\def\1{\bm{1}}

\DeclareMathAlphabet{\mathsfit}{\encodingdefault}{\sfdefault}{m}{sl}
\SetMathAlphabet{\mathsfit}{bold}{\encodingdefault}{\sfdefault}{bx}{n}

\newcommand{\cmark}{\ding{51}}
\newcommand{\xmark}{\ding{55}}
\title{ZoomGain: Mining Global-to-Zoom Perception Gaps for Fine-Grained Visual Supervision}
\author{\begin{tabular}{@{}c@{}}
Haozhan Shen$^{1,2 *}$\quad Liangbo He$^1$\quad Jun Lan$^{1 \dagger}$\quad Huijia Zhu$^1$\\[3pt]
Weiqiang Wang$^1$\quad Tiancheng Zhao$^{3,4 \dagger}$\quad Jianwei Yin$^2$
\end{tabular}}
\affiliation{\begin{tabular}{@{}c@{}}
$^1$Ant Group\quad $^2$Zhejiang University\\[2pt]
$^3$Om AI Research\quad $^4$Binjiang Institute of Zhejiang University
\end{tabular}}

\begin{document}
\maketitle
\begingroup
\renewcommand{\thefootnote}{}
\makeatletter
\renewcommand{\@makefntext}[1]{\noindent#1}
\makeatother
\begin{NoHyper}
\footnotetext{\textsuperscript{*}Work done during an internship at Ant Group, \textsuperscript{$\dagger$}Corresponding authors\\
Contact: \href{mailto:hz_shen@zju.edu.cn}{hz\_shen@zju.edu.cn}, \href{mailto:yelan.lj@antgroup.com}{yelan.lj@antgroup.com}, \href{mailto:tianchez@zju-bj.com}{tianchez@zju-bj.com}.}
\end{NoHyper}
\endgroup

\begin{abstract}
Fine-grained visual perception requires Vision-Language Models (VLMs) to locate and interpret subtle evidence within complex scenes.
Constructing supervision for this ability is challenging:
each question must unambiguously identify its target in the full image, and answering it correctly must require fine-grained visual evidence.
Checking both requirements makes manual annotation labor-intensive.
Automatic data generation is also difficult because VLMs often ignore the very details in the full image, 
but the same models can often recognize such details once the relevant region is zoomed in.
We exploit this global-to-zoom perception gap and introduce \textbf{ZoomGain},
a data construction pipeline that constructs questions about the full image
from visual information \textbf{gain}ed \textit{only} by \textbf{zoom}ing in.
To resolve ambiguity when referring to a target in the full image, 
ZoomGain checks the uniqueness of the target's referring expression via multiple models grounding.
It also verifies answers on the zoomed view, filters out language-prior questions, and specifically constructs distractors.
With this pipeline and human-only screening (no manual QA editing), 
we build ZoomGainBench, a 1,000-question bilingual benchmark on which the best of 28 VLMs achieves only 61.08.
Failure analysis reveals that models favor a particular type of distractor.
Beyond evaluation, training on just 9K ZoomGain samples
substantially improves the perception capabilities of our 4B/9B models,
enabling them to outperform several hundred-billion-parameter models.
% The trained 9B model achieves SoTA performance on several benchmarks, such as 90.25 on HR-Bench 4K.
These results establish global-to-zoom perception gaps as a practical basis for scaling fine-grained visual supervision for both evaluation and training.
Pipeline code and data will be released after publication.
Our code is available at \href{https://github.com/inclusionAI/ZoomGain}{https://github.com/inclusionAI/ZoomGain}.
\end{abstract}

\section{Introduction}
\label{sec:intro}

\begin{figure}[t]
    \centering
    \includegraphics[width=1\textwidth]{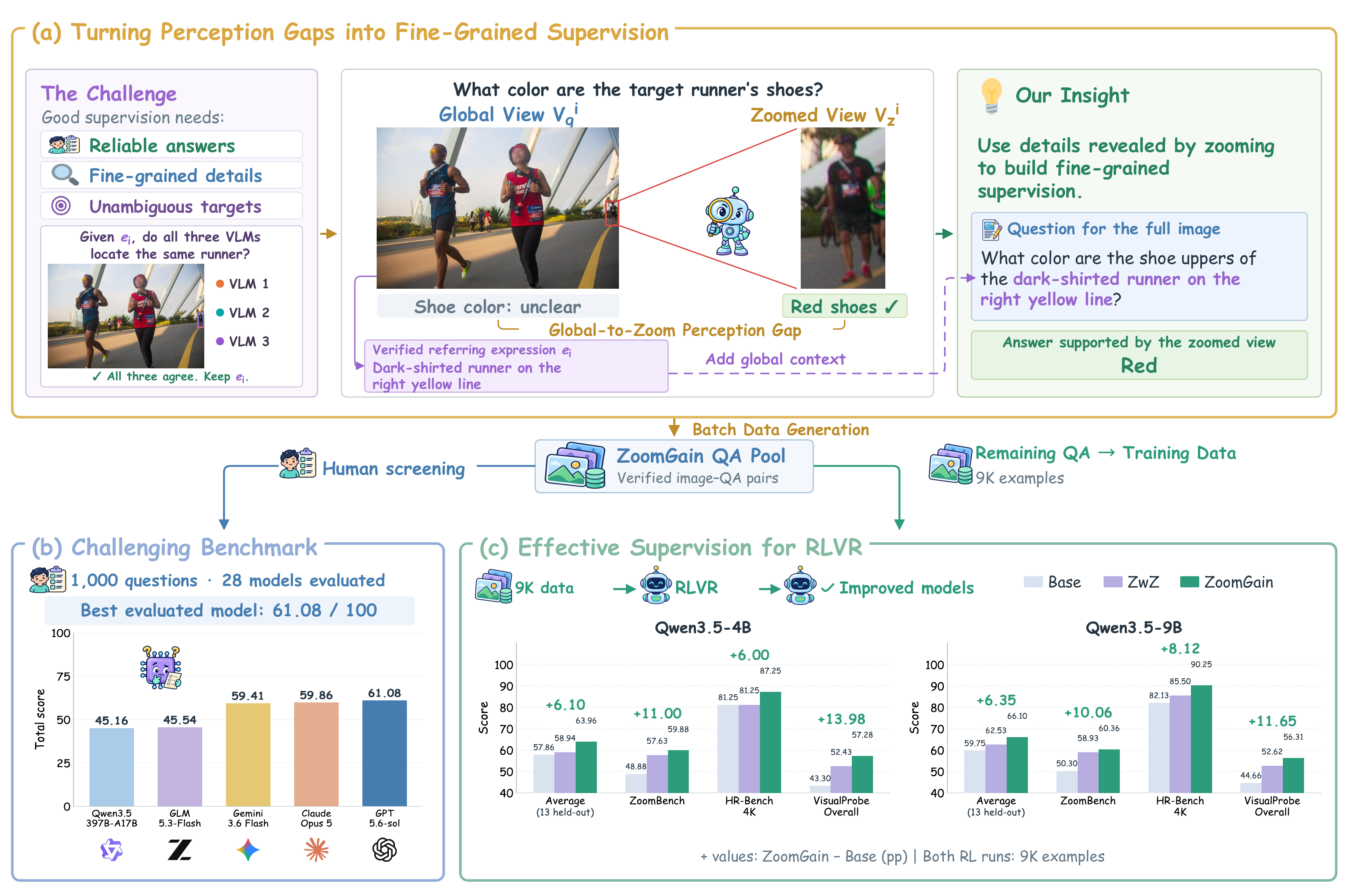}
    
    \caption{\textbf{ZoomGain: turning global-to-zoom perception gaps into fine-grained visual supervision.}
    (a) The runner's shoe color is unclear in the global view but discernible in the zoomed view;
    ZoomGain asks about such details in the full image,
    using a verified global referring expression to identify the target.
    The paired views $V_g^i$ and $V_z^i$, and the referring $e_i$ are defined in Section~\ref{sec:incremental-qa}.
    (b) ZoomGainBench, obtained by human screening without QA editing, challenges all 28 evaluated VLMs
    (c) RLVR on 9K training QA pairs improves Qwen3.5-4B and Qwen3.5-9B
    over their base models and ZwZ-trained counterparts~(ICML'26~\cite{wei2026zooming}).
    }
    \label{fig:zoomgain-overview}
    
\end{figure}

Fine-grained visual perception is essential in many practical applications, 
from detecting small vehicles in aerial imagery~(\cite{Xia_2018_CVPR}) and 
identifying subtle defects in industrial inspection~(\cite{Bergmann_2019_CVPR}) 
to reading product labels for people with visual impairments~(\cite{Gurari_2018_CVPR}). 
Addressing such tasks with Vision-Language Models~(VLMs)~(\cite{qwen35blog,Qwen3-VL,zhu2025internvl3,liu2023visual}) requires locating and interpreting 
small details that are easily overlooked in the full image.
To develop and benchmark such fine-grained capabilities, high-quality supervision is essential: 
answers must be correct and unambiguous,
and questions must genuinely require fine-grained visual evidence. 

Existing fine-grained benchmarks such as V* Bench~(\cite{wu2023vstar}), HR-Bench~(\cite{wang2024hrbench}),
and TreeBench~(\cite{wang2025treebench}) rely on manual annotation or heavy human editing of model drafts
(Table~\ref{tab:bench-compare}), keeping dataset sizes small (only hundreds of samples)
and impractical for large-scale RL training.
Meanwhile, directly prompting VLMs to generate question-answer pairs~(QAs) from full images is unreliable:
if a detail is already easy for the VLM to see in the full image, the resulting question is too simple;
if a detail is truly tiny, the VLM itself often fails to see it clearly and generates wrong answers.
(\textit{Issue 1: manual curation is costly, while full-image synthesis is unreliable.}) 

To avoid this, recent scalable pipelines such as ZwZ~(\cite{wei2026zooming}) and Vision-OPD~(\cite{yuan2026vision})
instead synthesize QAs solely from local crops.
Since these questions are generated from the crop alone without seeing the full image,
it is often unclear which object they refer to in the full image.
Consequently, these pipelines must overlay red bounding boxes on training images to specify targets,
creating a train-test visual mismatch since no oracle boxes exist at inference.
We further find that this red-box training paradigm binds the model's perception to box prompts~(detailed in Appendix~\ref{app:red-box}).
(\textit{Issue 2: crop-based synthesis suffers from referring ambiguity and binds perception to bounding-box.})

Moreover, existing fine-grained benchmarks and data pipelines
neither construct specific distractors nor apply blind shortcut filtering (Table~\ref{tab:bench-compare}),
even though language-prior leakage is well documented in general VLM evaluation~(\cite{cadene2019rubi,mmstar}).
We find that some of their questions can be answered correctly without the image
(see examples in Appendix~\ref{app:language-prior}).  
This may allow models to eliminate generic multiple-choice options without actually perceiving the tiny detail,
leading to inflated benchmark scores and false-positive RL rewards from lucky guesses.
(\textit{Issue 3: existing methods lack blind-shortcut filtering and dedicated distractor construction.}) 

Details a model fails to resolve in the full image often become recognizable once zoomed in.
Unlike prior work that zooms in at inference time~(\cite{wu2023vstar,zheng2026deepeyes})
or distills zoomed-in perception during training~(\cite{wei2026zooming,yuan2026vision}),
we use this global-to-zoom perception gap to construct the supervision itself
(Fig.~\ref{fig:zoomgain-overview}(a)).

In this paper, we introduce \textbf{ZoomGain}, an automated data construction pipeline
that constructs questions about the full image
from visual information \textbf{gain}ed \textit{only} by \textbf{zoom}ing in~(\textbf{Contribution \#1}).
For Issue 1, by contrasting global and zoomed descriptions,
ZoomGain generates questions strictly from incremental details,
which are unclear or absent in the full-image description
but verified by multi-model consensus on the zoomed view,
ensuring both challenging questions and reliable answers.
For Issue 2, ZoomGain generates globally unique natural-language referring expressions
and verifies them via multi-model grounding consensus.
This enables RL training directly on clean, unmodified images without bounding-box prompts.
For Issue 3, ZoomGain applies blind filtering,
and builds multi-source distractors from visual evidence, blind guesses, and full-image errors
that could mislead the queried VLMs.

Using ZoomGain, we curate \textbf{ZoomGainBench}, a bilingual benchmark of 1,000 questions
on attribute recognition, counting, and OCR, whose questions and answers are entirely model-authored:
two human reviewers only accept/reject candidates (\textbf{Contribution \#2}).
Among 28 evaluated VLMs, the best model, GPT-5.6-Sol, scores only 61.08,
and only five exceed 50~(Fig.~\ref{fig:zoomgain-overview}(b)).
Error analysis further reveals that incorrect choices concentrate on distractors
mined from full-image mispredictions during construction (Sec.~\ref{sec:verification-distractors}).
This indicates that when details are unresolved, 
models tend to fill the information gap with similar scene-plausible guesses,
suggesting that the global-to-zoom perception gap is a common weakness across current VLMs, 
rather than an issue to specific models.

Beyond evaluation, ZoomGain data also serves as effective training supervision~(\textbf{Contribution \#3}).
With identical source images, Qwen3.5-4B and Qwen3.5-9B~(\cite{qwen35blog}) trained on 9K ZoomGain QA pairs
outperform their ZwZ-trained counterparts~(\cite{wei2026zooming}) on all 14 benchmarks (Fig.~\ref{fig:zoomgain-overview}(c)).
The gains extend to general perception and difficult perception-and-reasoning tasks,
e.g., 11.01 points on O3-Bench~(\cite{li2026insight_o3}) for the 9B model, which is on par with GLM-5.3-Flash~(\cite{glm5team2026glm5vibecodingagentic}).
Ablations confirm that both incremental evidence and multi-source distractors are essential,
and further analysis shows that training improves recognition of zoom-revealed details in the full image (Sec.~\ref{sec:caption-gap}).

\section{Related Work}
\label{sec:releted}

\paragraph{Fine-grained perception benchmarks.}
A variety of benchmarks specifically evaluate fine-grained perception in high-resolution images, 
including~(\cite{wu2023vstar,wang2024hrbench,zhang2024mmerealworld,wang2025treebench,Gavrikov_2026_visualoverload,lai2025mini-o3,zhang2026finers,wei2026zooming});
O3-Bench~(\cite{li2026insight_o3}) further requires multi-step reasoning beyond perception.
However, questions in these benchmarks are typically annotated manually or generated by models with extensive manual editing (Table~\ref{tab:bench-compare}), making scalable cruation costly.
Moreover, existing benchmarks rarely filter out language-prior shortcuts~(\cite{cadene2019rubi,mmstar,yue-etal-2025-mmmu})
or design targeted distractors. In contrast, ZoomGain fully automates QA authoring with zoom-view verification, 
blind shortcut filtering, and distractor mining from model errors.

\paragraph{Improving fine-grained perception.}
Zooming has been widely adopted to enhance fine-grained perception, 
either at inference time via visual search~(\cite{wu2023vstar,shen2025zoomeye,lai2025mini-o3,li2026insight_o3,wangretrieval}) 
and agentic cropping tools~(\cite{zheng2026deepeyes,hong2026deepeyesv2,su2026pixel,zhang2025thymethinkimages}), 
or during training through zoomed-to-global distillation~(\cite{wei2026zooming,yuan2026vision,li2026visual,zhang2026vad}). 
Unlike prior works that focus on inference procedures or training objectives, 
ZoomGain targets data construction itself: 
it leverages the global-to-zoom perception gap to identify what to ask (strictly on zoom-revealed details) 
and reliably verify ground truth, making it complementary to existing approaches.

\section{ZoomGain}
\label{sec:pipeline}

\begin{figure}[t]
    \centering
    \includegraphics[width=1\linewidth]{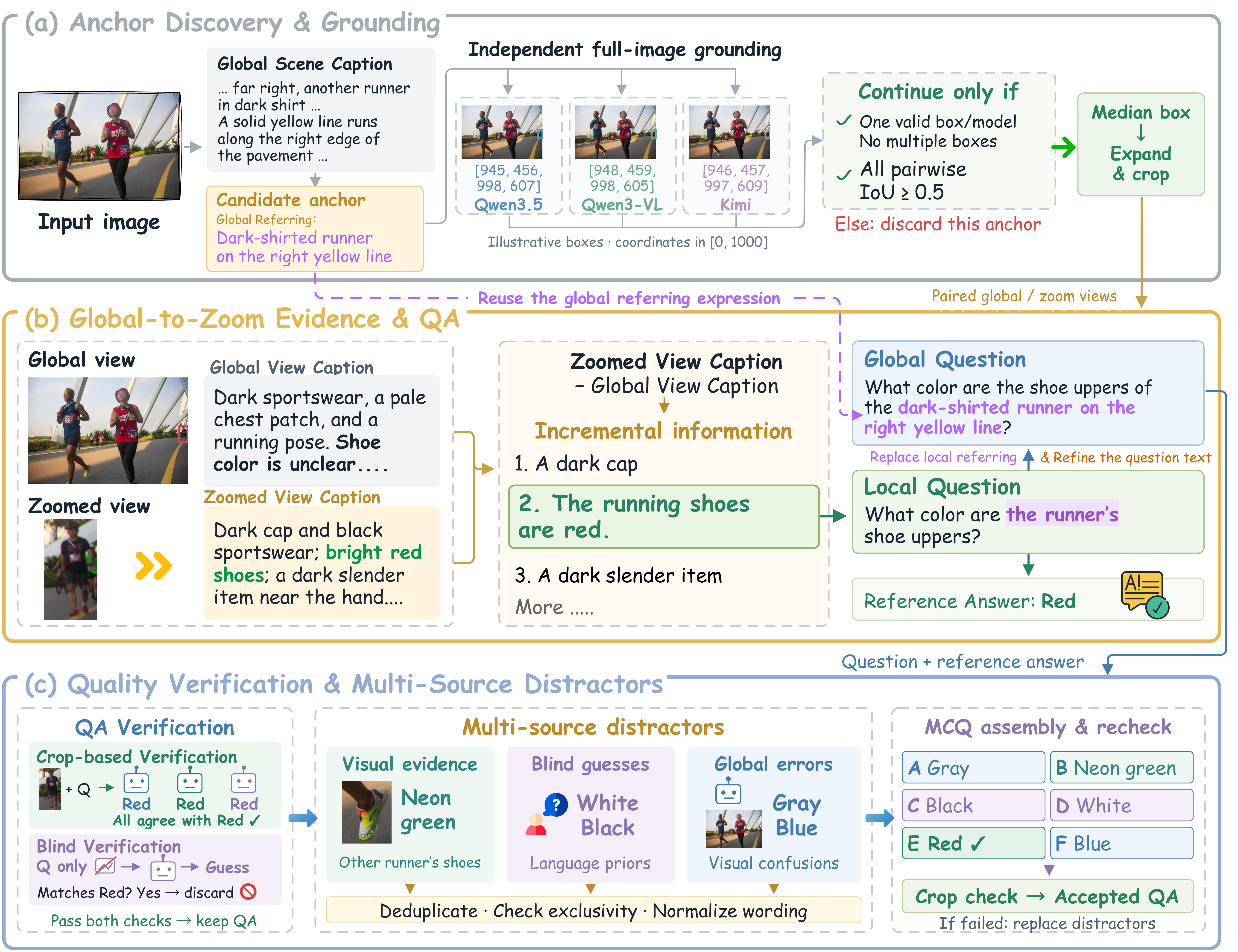}
    
    \caption{\textbf{Overview of ZoomGain.}
    (a) Discover anchors and obtain consensus regions through independent grounding.
    (b) Mine global-to-zoom incremental evidence from paired views to construct questions and reference answers.
    (c) Verify answers and visual dependence, then assemble and recheck
    multiple-choice questions with multi-source distractors.}
    \label{fig:zoomgain-pipeline}
    
\end{figure}

ZoomGain exploits the perception gap between global and zoomed-in views:
fine-grained details that are difficult to discern in the full image
may become clear once the target region is zoomed in.
As illustrated in Figure~\ref{fig:zoomgain-pipeline},
ZoomGain first discovers and grounds target anchors (Sec.~\ref{sec:anchor-grounding}),
then constructs questions and answers from details newly revealed by zooming (Sec.~\ref{sec:incremental-qa}),
and finally verifies answer reliability and visual dependence
and builds multi-source distractors that capture plausible model errors (Sec.n~\ref{sec:verification-distractors}).
The complete algorithm and additional implementation details are provided in
Appendix~\ref{app:construction-algorithm}.

\subsection{Anchor Discovery and Grounding}
\label{sec:anchor-grounding}

Given an input image, denoted as the global view $V_g$,
we first use a VLM to generate a global scene caption $C_g$.
A text-only LLM then takes $C_g$ as its sole input
and proposes a set of candidate anchors.
Each anchor $i$ is represented by a referring expression $e_i$
that describes the target's appearance, spatial location, or relations to other objects.
To ensure that $e_i$ can identify a localizable instance unambiguously in the full image,
three VLMs independently ground $e_i$ in $V_g$.
We retain $e_i$ only if each model returns exactly one valid bounding box,
since a missing box or multiple boxes indicate that the referent cannot be found or may be ambiguous,
and all pairwise box IoUs are at least $0.5$,
which provides cross-model evidence that $e_i$ refers to a consistent region.
For each retained anchor, we take the coordinate-wise median of the predicted boxes,
expand the resulting region to include more surrounding context,
yielding the crop box $b_i$, and crop $V_g$ using $b_i$ to obtain a zoomed view $V_z^i$.
The quadruples $(V_g, V_z^i, b_i, e_i)$
serve as inputs to the subsequent question construction stage.

\subsection{Global-to-Zoom Evidence and QA Construction}
\label{sec:incremental-qa}

For each retained anchor $i$, we use a strong VLM $M_{\mathrm{cap}}$
to describe the target region from both views.
Specifically, we draw a red bounding box at $b_i$ in $V_g$
to obtain an annotated global view $V_g^i$.
Figure~\ref{fig:zoomgain-overview}(a) illustrates the paired views $(V_g^i, V_z^i)$.
We feed $V_g^i$ into $M_{\mathrm{cap}}$ and prompt it to describe
only the contents of the red box in as much detail as possible,
yielding the global-view caption $C_g^i$.
In a separate, independent call,
we provide the zoomed view $V_z^i$ and request a detailed caption $C_z^i$.
This produces two descriptions of the same region $b_i$ at different viewing scales.
We then use an LLM to compare $C_g^i$ and $C_z^i$,
extracting details that are explicitly described in $C_z^i$
but remain unclear or unmentioned in $C_g^i$.
These details form a candidate fact set $\Delta_i$.
For example, in Figure~\ref{fig:zoomgain-pipeline}(b),
the runner's shoe color is unclear in the global-view caption
but identified as red in the zoomed-view caption,
so it becomes a candidate fact in $\Delta_i$.

We generate three types of questions:
attribute recognition, text recognition (OCR), and counting.
An LLM selects one fact from $\Delta_i$ that supports a definite answer
and generates a local question $q_{\mathrm{local}}^i$
and a reference answer $a_i$ based on this fact.
The local question is phrased for the zoomed view
with a simple reference such as ``the runner'',
without specifying the target's location in the full image.
The LLM then replaces this reference with the global referring expression $e_i$
and adjusts the wording to produce a global-view question $q_{\mathrm{global}}^i$,
keeping the queried detail and reference answer unchanged,
so that questions about details revealed by zooming
clearly identify the intended target in the full image.
For example, ``What color are the runner's shoe uppers?''
becomes ``What color are the shoe uppers of
the dark-shirted runner on the right yellow line?''
Appendix~\ref{app:dual-caption-examples} illustrates this construction process
with attribute, counting, and OCR examples.

\subsection{Quality Verification and Multi-Source Distractors}
\label{sec:verification-distractors}

We verify each candidate QA pair in two ways.
(1) For \emph{crop-based verification}, we provide several VLMs
with $q_{\mathrm{local}}^i$ and the zoomed view $V_z^i$,
and sample multiple answers from each model.
We retain the QA pair only if all models meet a predefined accuracy threshold on their sampled answers.
Checking agreement among multiple models answering based on the zoomed-view helps ensure answer correctness,
thereby avoiding sole reliance on a single model's caption for answer annotation, which may contain hallucinations~(\cite{rohrbach-etal-2018-object,li-etal-2023-evaluating,jing-etal-2024-faithscore}).
(2) For \emph{blind verification}, models are asked to answer $q_{\mathrm{global}}^i$ without an image.
We discard a QA if all blind models exceed a prescribed correct-answer threshold.
This step effectively filters out questions that can be answered through language priors alone rather than actual image content~(\cite{cadene2019rubi,mmstar,yue-etal-2025-mmmu}).

For verified QA pairs suited to multiple-choice evaluation,
we construct distractors from three complementary sources.
(1)\;\emph{Visual evidence} supplies plausible alternatives from image content,
such as the neon-green shoes of another runner in Figure~\ref{fig:zoomgain-pipeline}(c),
which can expose errors in target selection or attribute binding.
(2)\;\emph{Blind guesses} provide incorrect answers produced without the image,
capturing language priors.
(3)\;\emph{Global-view errors} provide incorrect answers produced from the full image,
capturing confusions when local details are difficult to discern.
We remove candidates equivalent to the reference answer or also valid for the question,
merge semantically redundant candidates,
and normalize their wording to form distinct answer options.

We combine the reference answer with the selected distractors
and recheck the resulting MCQ using the zoomed view.
This final check tests whether the reference answer remains reliably distinguishable
after competing options are introduced.
If the per-model agreement thresholds are not met,
we replace or remove problematic distractors.
OCR and counting questions retain an open-ended format
and bypass distractor construction and MCQ rechecking.
Appendix~\ref{app:verification-examples} provides examples of candidate QA pairs
rejected by crop-based and blind verification.

\section{Implementation Details and Data Preparation}
\label{sec:data-preparation}

\paragraph{Model Roles and Verification Settings.}
We use Qwen3.5-397B-A17B~(\cite{qwen35blog}) as $M_{\mathrm{cap}}$
and the initial global scene captioner.
GLM-5.2~(\cite{glm5team2026glm5vibecodingagentic}) performs incremental fact extraction and QA generation.
Global-view error distractors are mined from incorrect answers produced by
Qwen3.5-397B-A17B, Qwen3-VL-235B-A22B-Instruct~(\cite{Qwen3-VL}),
and Kimi-K2.6~(\cite{kimiteam2026kimik25visualagentic})
to the global question on the full image,
with 10 answers sampled per model (30 in total).
All verification stages likewise sample 10 answers per model and apply per-model accuracy thresholds.
Complete model assignments, verification thresholds, and distractor repair settings are provided in
Appendix~\ref{app:construction-settings}.

\paragraph{Data Generation and Benchmark Curation.}
We randomly sample 12K images from ZwZ~(\cite{wei2026zooming}),
which collects high-resolution images from several sources~(\cite{kirillov2023segment,chuang2025meta,chen2022stpls3d,changpinyo2021conceptual,krishna2017visual,schuhmann2022laion}),
and generate a pool of 23,610 QA pairs with ZoomGain.
GLM-5.2 jointly generates Chinese and English versions of both the local and global questions,
and the two language versions of a question are counted as one QA pair.
From this pool, two human reviewers select 1,000 questions for the benchmark
without writing or editing any questions or answers.

\paragraph{Training Data Selection and Matching.}
\label{subsec:training-data}
We exclude all QAs pairs associated with the images shown in ZoomGainBench,
so the training pool and the benchmark have no same images.
For a fair comparison with ZwZ, for each image, we take the smaller number of available unique questions
between the two sides and sample that many from each side without replacement,
yielding 8,746 image-matched QA pairs.
We then duplicate 254 randomly chosen pairs on both sides to reach 9,000 samples per dataset.
Both training sets thus use identical source images (ZwZ retains its bounding-box overlays)
and differ solely in the questions.

\begin{table}[t]
    \centering
    \renewcommand{\cmark}{\textcolor{green!70!black}{\ding{51}}}
    \renewcommand{\xmark}{\textcolor{black!60}{\ding{55}}}
    \caption{\textbf{Comparison with existing fine-grained perception benchmarks.}
    \emph{QA authoring}: who writes the questions and answers; 
    Model+Human denotes model-generated questions refined by human annotators,
    and Model denotes questions authored entirely by models,
    with humans only deciding which candidates to include.
    \emph{Blind filter}: questions answerable without the image are removed during construction.
    \emph{Local verif.}: answers are automatically verified on the zoomed evidence region.
    \emph{Dual-view}: both the full image and the evidence region are provided in metadata;
    in standard evaluation, models are fed with only the full image.
    \emph{Dedicated distractor generation}: a separate stage for constructing distractor options
    (Section~\ref{sec:verification-distractors}).
    In this column, \xmark{} indicates that no dedicated stage is reported,
    and N/A denotes open-ended evaluation.
    ``--'' denotes not provided or not specified.}
    \label{tab:bench-compare}
    \small
    \setlength{\tabcolsep}{4.5pt}
    \resizebox{\linewidth}{!}{%
    \begin{tabular}{lrccccccc}
        \toprule
        Benchmark & \#Questions & QA authoring & Language & Evidence & Blind filter & Local verif. & Dual-view & \shortstack{Dedicated distractor\\generation} \\
        \midrule
        ZoomBench~(\cite{wei2026zooming})   & 845                & Model+Human & EN     & Auto  & \xmark & \xmark & \cmark & \xmark \\
        V* Bench~(\cite{wu2023vstar})        & 191                & Human       & EN     & Human & \xmark & \xmark & \xmark & \xmark \\
        HR-Bench~(\cite{wang2024hrbench})    & 200          & Model+Human       & EN     & \xmark & \xmark & \xmark & \xmark & \xmark \\
        VisualProbe~(\cite{lai2025mini-o3})   & 515          & --          & EN     & \xmark & \xmark & \xmark & \xmark & N/A \\
        TreeBench~(\cite{wang2025treebench}) & 405                & Model+Human & EN     & Human & \xmark & \xmark & \xmark & \xmark \\
        FINERS-4k~(\cite{zhang2026finers})   & 2,427        & Human       & EN     & --    & \xmark & \xmark & \xmark & \xmark \\
        \midrule
        \rowcolor[HTML]{F3F0FF}
        \textbf{ZoomGainBench (ours)}            & 1,000              & Model       & EN/ZH  & Auto  & \cmark & \cmark & \cmark & \cmark \\
        \bottomrule
    \end{tabular}}

\end{table}

\begin{table}[t]
    \centering
    \caption{\textbf{Evaluation results on ZoomGainBench.}
    Scores are reported as percentages for English and Chinese questions.
    Each language-specific Overall is weighted by the number of questions in each category.
    \textbf{Bold} and \underline{underlined} scores indicate the best and second-best results
    across all models in each column, respectively.}
    \label{tab:zoomgain-results}
    \definecolor{benchChineseOverall}{HTML}{EAF2FF}
    \definecolor{benchEnglishOverall}{HTML}{EDF7ED}
    \definecolor{benchAllOverall}{HTML}{F3F0FF}
    \fontsize{7}{6.2}\selectfont
    \setlength{\tabcolsep}{4pt}
    \renewcommand{\arraystretch}{1.08}
    \begin{tabular*}{\linewidth}{@{\extracolsep{\fill}}lccc>{\columncolor{benchEnglishOverall}}cccc>{\columncolor{benchChineseOverall}}c>{\columncolor{benchAllOverall}}c}
        \toprule
        & \multicolumn{4}{c}{\textbf{English}} & \multicolumn{4}{c}{\textbf{Chinese}} & \textbf{Overall} \\
        \cmidrule(lr){2-5}\cmidrule(lr){6-9}
        \textbf{Model} & Attr & Count & OCR & Overall & Attr & Count & OCR & Overall & \\
        \midrule
        \multicolumn{10}{c}{\textcolor{black!55}{\textbf{\textit{Closed-source Frontier VLMs}}}} \\
        \midrule
        GPT-5.5~(\cite{openai_gpt55}) & \underline{52.75} & 57.45 & 60.40 & 56.51 & 52.12 & 63.83 & 60.96 & 57.06 & 56.79 \\
        Gemini-3.6-Flash~(\cite{google_gemini36flash}) & 51.27 & 42.55 & \textbf{74.71} & \underline{60.62} & 48.52 & 45.74 & \textbf{71.42} & 58.20 & 59.41 \\
        Claude-Opus-5~(\cite{anthropic_claudeopus5}) & 48.09 & 63.83 & \underline{71.86} & 59.89 & 48.73 & 62.77 & \underline{71.29} & \underline{59.84} & \underline{59.86} \\
        GPT-5.6-Sol~(\cite{openai_gpt56sol}) & \textbf{56.57} & 62.77 & 67.93 & \textbf{62.08} & \textbf{54.03} & 63.83 & 65.84 & \textbf{60.07} & \textbf{61.08} \\
        \midrule
        \multicolumn{10}{c}{\textcolor{black!55}{\textbf{\textit{Agentic VLMs}}}} \\
        \midrule
        DeepEyes-7B~(\cite{zheng2026deepeyes}) & 24.79 & 20.21 & 41.05 & 31.42 & 24.36 & 22.34 & 34.46 & 28.55 & 29.99 \\
        Thyme-7B~(\cite{zhang2025thymethinkimages}) & 26.69 & 34.04 & 43.88 & 34.85 & 22.88 & 31.91 & 40.31 & 31.30 & 33.07 \\
        DeepEyesV2-7B~(\cite{hong2026deepeyesv2}) & 22.25 & 41.49 & 43.70 & 33.37 & 22.67 & 34.04 & 43.62 & 32.83 & 33.10 \\
        PixelReasoner-7B~(\cite{su2026pixel}) & 27.12 & 45.74 & 44.83 & 36.56 & 25.42 & 23.40 & 40.13 & 31.62 & 34.09 \\
        \midrule
        \multicolumn{10}{c}{\textcolor{black!55}{\textbf{\textit{Fine-grained Enhanced Single Forward VLMs}}}} \\
        \midrule
        ZwZ-4B~(\cite{wei2026zooming}) & 30.08 & 39.36 & 29.68 & 30.78 & 33.47 & 50.00 & 34.45 & 35.45 & 33.12 \\
        Vision-OPD-4B~(\cite{yuan2026vision}) & 35.59 & 59.57 & 45.69 & 42.23 & 39.62 & 59.57 & 43.80 & 43.31 & 42.77 \\
        Vision-OPD-9B~(\cite{yuan2026vision}) & 46.40 & 55.32 & 47.18 & 47.58 & 47.67 & 60.64 & 37.97 & 44.68 & 46.13 \\
        ZwZ-8B~(\cite{wei2026zooming}) & 42.58 & 55.32 & 50.53 & 47.23 & 44.70 & 52.13 & 51.77 & 48.47 & 47.85 \\
        \midrule
        \multicolumn{10}{c}{\textcolor{black!55}{\textbf{\textit{Open-source General VLMs}}}} \\
        \midrule
        MiMo-VL-7B-RL~(\cite{coreteam2025mimovltechnicalreport}) & 23.73 & 34.04 & 20.06 & 23.11 & 23.31 & 41.49 & 32.06 & 28.82 & 25.96 \\
        GLM-4.6V~(106B)~(\cite{vteam2025glm45vglm41vthinkingversatilemultimodal}) & 24.36 & 38.30 & 31.57 & 28.80 & 22.46 & 36.17 & 28.51 & 26.37 & 27.59 \\
        Qwen3-VL-4B~(\cite{Qwen3-VL}) & 21.82 & 43.62 & 31.14 & 27.92 & 22.88 & 41.49 & 33.63 & 29.30 & 28.61 \\
        Qwen3-VL-8B~(\cite{Qwen3-VL}) & 15.89 & 41.49 & 39.17 & 28.40 & 19.92 & 30.85 & 41.64 & 30.37 & 29.38 \\
        Qwen3.5-4B~(\cite{qwen35blog}) & 21.40 & 51.06 & 34.40 & 29.83 & 22.67 & 46.81 & 34.87 & 30.23 & 30.03 \\
        Qwen3.5-9B~(\cite{qwen35blog}) & 21.82 & 41.49 & 37.79 & 30.60 & 22.67 & 47.87 & 42.52 & 33.66 & 32.13 \\
        Qwen3.5-35B-A3B~(\cite{qwen35blog}) & 28.18 & 44.68 & 46.10 & 37.51 & 27.33 & 51.06 & 47.95 & 38.51 & 38.01 \\
        Qwen3.6-35B-A3B~(\cite{qwen36_35b_a3b}) & 26.91 & 54.26 & 50.21 & 39.59 & 26.69 & 52.13 & 46.17 & 37.54 & 38.57 \\
        Qwen3-VL-235B-A22B~(\cite{Qwen3-VL}) & 29.45 & 52.13 & 48.01 & 39.64 & 30.93 & 50.00 & 50.90 & 41.39 & 40.51 \\
        Qwen3.5-27B~(\cite{qwen35blog}) & 33.05 & 51.06 & 45.92 & 40.33 & 34.75 & 58.51 & 48.67 & 43.02 & 41.68 \\
        Qwen3.6-27B~(\cite{qwen3.6-27b}) & 33.47 & 56.38 & 47.47 & 41.70 & 37.71 & \textbf{70.21} & 45.33 & 44.07 & 42.89 \\
        Qwen3.8-27B~(\cite{qwen3.8}) & 40.25 & 54.26 & 45.49 & 43.84 & 44.92 & 62.77 & 43.05 & 45.78 & 44.81 \\
        Qwen3.5-397B-A17B~(\cite{qwen35blog}) & 33.47 & \underline{65.96} & 51.95 & 44.55 & 38.14 & 62.77 & 50.41 & 45.78 & 45.16 \\
        GLM-5.3-Flash~(320B)~(\cite{glm5team2026glm5vibecodingagentic}) & 44.07 & 54.26 & 43.77 & 44.90 & 46.61 & 55.32 & 43.74 & 46.18 & 45.54 \\
        Kimi-K3~(2.8T)~(\cite{kimiteam2026kimik3openfrontier}) & 49.15 & 64.89 & 45.48 & 49.04 & 46.82 & 64.89 & 46.22 & 48.26 & 48.65 \\
        Qwen3.8-Flash-Next~(125B)~(\cite{qwen2026design,qwen3.8flashnext}) & 47.46 & \textbf{67.02} & 56.97 & 53.42 & \underline{52.97} & \underline{67.02} & 58.45 & 56.67 & 55.05 \\
        \bottomrule
    \end{tabular*}
    
\end{table}

\section{Experiments on ZoomGainBench}
\label{sec:benchmark}

To examine whether questions mined from global-to-zoom perception gaps
pose a genuine challenge to current VLMs,
we evaluate 28 widely-used VLMs on ZoomGainBench,
spanning closed-source frontier, agentic, fine-grained enhanced single-forward,
and open-source general VLMs (Table~\ref{tab:zoomgain-results}).

\vspace{-4pt}
\paragraph{Statistics.} 
ZoomGainBench contains 1,000 questions over 941 images:
472 attribute recognition, 94 counting, and 434 OCR questions.
Attribute questions have 4 to 6 answer options; counting and OCR questions are open-ended.
Each question has English and Chinese versions (2,000 evaluation instances in total)
and an automatically obtained evidence region specified by its crop box.
Appendix~\ref{app:benchmark-categories} shows examples of all three question categories.
Table~\ref{tab:bench-compare} compares ZoomGainBench with existing benchmarks:
it is the only one that combines fully model-authored QAs, automated verification,
and dedicated distractor construction,
and Section~\ref{sec:error-analysis} shows how these distractors reveal errors shared across models.

\vspace{-4pt}
\paragraph{Evaluation protocol.}
Multiple-choice questions are scored by exact match on the selected option,
and counting questions by exact match on the predicted integer.
OCR questions receive their ANLS score if it is at least $0.7$,
and zero otherwise.
Prompts, decoding temperature, and thinking-mode settings
for each model are provided in Appendix~\ref{app:benchmark-eval-settings}.

\subsection{Benchmark results.}

\textbf{ZoomGain challenges even frontier VLMs.}
The strongest model, GPT-5.6-Sol, reaches an overall score of only 61.08 (Table~\ref{tab:zoomgain-results}), and only five models exceed 50.
All four closed-source models outperform every open-source general VLM:
even GPT-5.5 (56.79) surpasses Qwen3.8-Flash-Next (55.05),
the best open-source general model.

\textbf{Fine-grained enhanced VLMs are highly competitive.}
ZwZ-8B and Vision-OPD-9B improve over their respective base models,
Qwen3-VL-8B and Qwen3.5-9B, by 18.47 and 14.00 points.
Despite its smaller size, ZwZ-8B (47.85) also surpasses Qwen3.5-397B-A17B (45.16).

\textbf{Attribute recognition is the most difficult category.}
Attribute recognition yields the lowest score averaged over both languages for 23 of 28 models,
despite being the only multiple-choice category, with a uniform-guessing accuracy of about 21.5\%.
Qwen3-VL-8B, for example, falls below this level in both languages (15.89 in English and 19.92 in Chinese),
motivating an analysis of which distractors attract incorrect choices.

\textbf{The evidence region makes questions solvable.}
We also evaluate the models conditioned on the provided evidence regions, 
with detailed results reported in Appendix~\ref{app:dual-view}.
Once the evidence region is given, these challenging questions become solvable,
confirming that the difficulty lies in finding and resolving the detail in the full image

\subsection{Error analysis}
\label{sec:error-analysis}

\begin{figure}[t]
    \centering
    \includegraphics[width=\linewidth]{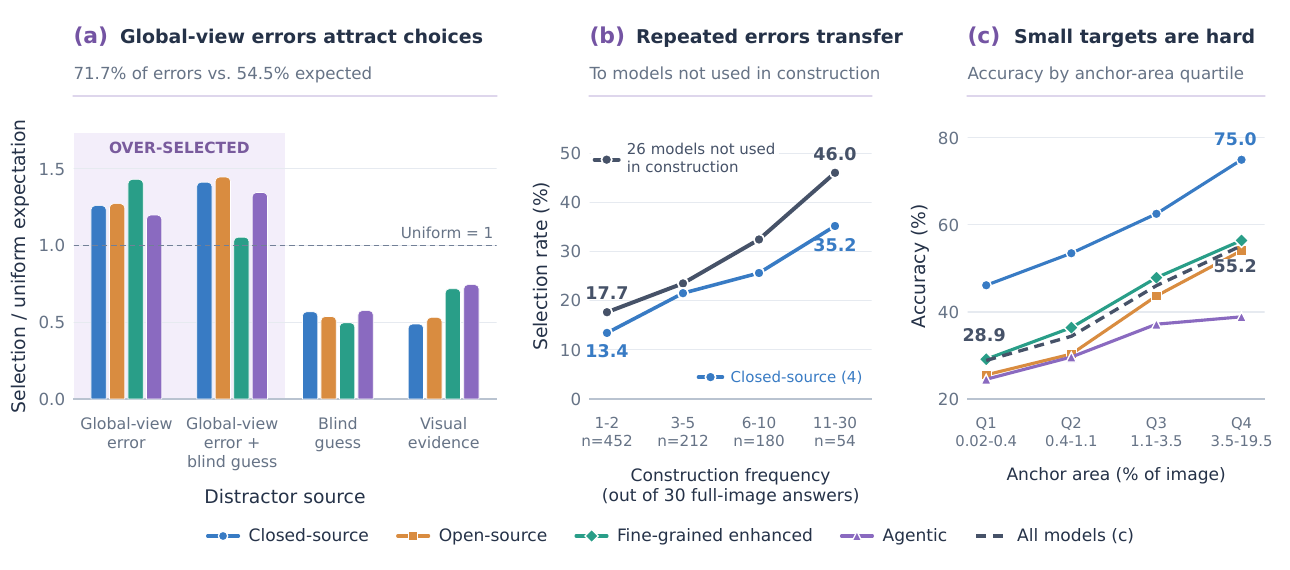}
    
    \caption{\textbf{Where models fail on ZoomGainBench.}
    (a) How often each model group chooses each type of distractor
    compared with the random baseline, considering only incorrect answers.
    Values above 1 indicate over-selection.
    \emph{Global-view error + blind guess} denotes distractors produced both as incorrect answers
    when models view the full image and as guesses based on language priors without an image.
    (b) How often the 26 models not used in construction
    (all evaluated models except Qwen3.5-397B-A17B and Qwen3-VL-235B-A22B-Instruct)
    choose global-view errors, grouped by each error's frequency
    in the 30 full-image answers collected during construction.
    (c) Accuracy by quartile of the anchor's area in the full image.}
    \label{fig:error-analysis}
    
\end{figure}

\paragraph{Models are misled by global-view errors.}
Since every distractor retains its source from construction
(Section~\ref{sec:verification-distractors}),
we can trace where incorrect choices go.
Across all 28 models, global-view-error distractors
receive 71.7\% of incorrect choices, exceeding the random baseline of 54.5\%,\footnote{The baseline assumes that whenever a model answers incorrectly,
it selects uniformly at random among that question's distractors.}
whereas blind guesses and visual evidence receive only 0.54 and 0.59 times their random-baseline
(Figure~\ref{fig:error-analysis}(a)).
This over-selection holds for every model, including those not involved in construction
Moreover, wrong answers that appear more often during construction are also chosen more often by other models.
Among the 26 models not used in construction,
errors that appeared 1--2 times in the 30 full-image answers are chosen 17.7\% of the time,
versus 46.0\% for those that appeared more than 10 times
(Figure~\ref{fig:error-analysis}(b)).
That is, models reproduce the global-view confusions observed during construction,
guessing a scene-plausible answer instead of resolving the queried detail,
such as naming a hockey equipment brand for the logo on a hockey player's pants
(shown in Figure~\ref{fig:error-cases}(a) in Appendix~\ref{app:error-cases}).
\textbf{In other words, distractors mined from global-view errors capture confusions shared across models,
rather than idiosyncrasies of the models used in construction.}

\paragraph{Smaller targets are harder.}
Accuracy rises with the fraction of the image occupied by the anchor for every model group
(Figure~\ref{fig:error-analysis}(c)):
averaged over all models, it increases from 28.9\% on the smallest quartile,
where anchors occupy less than 0.4\% of the image, to 55.2\% on the largest.

Appendix~\ref{app:error-analysis} further reports the error analysis regarding open-ended questions,
including how incorrect counts differ from the ground truth,
as well as common OCR error patterns. 

\section{Experiments on ZoomGain Training Data}
\label{sec:training}

\subsection{Main Results}
\paragraph{Setup.}
We conduct the experiment on the image-matched ZoomGain and ZwZ sets described in Section~\ref{subsec:training-data},
each with 9,000 samples.
We select Qwen3.5-4B and Qwen3.5-9B~(\cite{qwen35blog}) as our base models,
applying RL with GRPO~(\cite{shao2024deepseekmath}) using each set under identical settings
(batch size 64, 8 rollouts per question; other details are shown in Appendix~\ref{app:training-settings}).
For ZwZ, we retain its official recipe (ZwZ Original):
red bounding-box overlays, box-referring prompts, and multiple-choice rewards for counting and OCR.
To eliminate both visual-input and reward differences as potential confounders,
we additionally train a strictly controlled baseline on the same 9K samples, which is named as ZwZ Align:
(1)~we remove all red bounding-box overlays and box-referring instructions,
so that ZwZ Align uses the same full images as ZoomGain;
(2)~we convert ZwZ's counting and OCR questions from multiple-choice to open-ended
and train using the exact same reward functions as ZoomGain.
We evaluate on four groups of benchmarks.
\emph{Held-in}: ZoomGainBench.
\emph{Held-out fine-grained perception}: ZoomBench~(\cite{wei2026zooming}), V* Bench~(\cite{wu2023vstar}),
HR-Bench 4K and 8K~(\cite{wang2024hrbench}), MME-RealWorld-Lite~(\cite{zhang2024mmerealworld}),
TreeBench~(\cite{wang2025treebench}), and VisualProbe~(\cite{lai2025mini-o3}).
\emph{General perception}: CV-Bench~(\cite{tong2024cambrian1}), MMStar~(\cite{mmstar}), and MMVP~(\cite{tong2024eyes}).
\emph{Difficult perception and reasoning}: BabyVision~(\cite{chen2026babyvision}),
PerceptionBench~(\cite{lin2026perceptionbench}),
and O3-Bench~(\cite{li2026insight_o3}). 
We also compare with some large parameter VLMs, including open-source and closed-source~(\cite{google_gemini3flash,openai_gpt51}).
We report the average score over the 13 held-out benchmarks, excluding ZoomGainBench.
Because ZoomGainBench and the ZoomGain training data are generated by the same generation pipeline,
including it could favor ZoomGain model.

\begin{table}[t]
    \centering
    \caption{\textbf{Results of RL training on ZoomGain and ZwZ data.}
    $\Delta_{\mathrm{ZoomGain-ZwZ}}$ and $\Delta_{\mathrm{ZoomGain-Base}}$ denote score gains
    over the ZwZ-trained and base models, respectively;
    Avg.\ is computed over the 13 held-out benchmarks, \textbf{excluding ZoomGainBench}.
    \emph{w/ ZwZ 9K} follows the official ZwZ recipe (ZwZ Original);
    \emph{w/ ZwZ Align 9K} uses the same clean images and reward functions as ZoomGain.
    $\Delta_{\mathrm{ZoomGain-ZwZ}}$ is computed against ZwZ Original.}
    \label{tab:training-results}
    \definecolor{deltaup}{HTML}{16A34A}
    \definecolor{deltadown}{HTML}{DC2626}
    \definecolor{deltabg}{HTML}{EFEFEF}
    \newcommand{\deltaUp}[1]{\textcolor{deltaup}{#1}}
    \newcommand{\deltaDown}[1]{\textcolor{deltadown}{#1}}
    \newcommand{\deltaTie}[1]{\textcolor{black!50}{#1}}
    \setlength{\tabcolsep}{1.5pt}
    \renewcommand{\arraystretch}{1.12} 
    \resizebox{\linewidth}{!}{%
    \begin{tabular}{l|c|ccccccc|ccc|ccc|c}
        \toprule
        & \textbf{Held-in}
        & \multicolumn{7}{c|}{\textbf{Fine-grained Perception}}
        & \multicolumn{3}{c|}{\textbf{General Perception}}
        & \multicolumn{3}{c|}{\textbf{Difficult Percept \& Reason}}
        & \\
        \cmidrule(lr){2-2}\cmidrule(lr){3-9}\cmidrule(lr){10-12}\cmidrule(lr){13-15}
        \textbf{Model}
        & \shortstack{ZoomGain\\Bench}
        & \shortstack{Zoom\\Bench} & \shortstack{V*\\Bench} & \shortstack{HR\\4K} & \shortstack{HR\\8K} & \shortstack{MME-RW \\ Lite} & \shortstack{Tree\\Bench} & \shortstack{Visual\\Probe}
        & \shortstack{CV-\\Bench} & \raisebox{0.5\normalbaselineskip}{MMStar} & \raisebox{0.5\normalbaselineskip}{MMVP}
        & \shortstack{Baby\\Vision} & \shortstack{Perception\\Bench} & \shortstack{O3-\\Bench}
        & \shortstack{\textbf{Avg.}\\{\scriptsize Only Held-out}} \\
        \midrule
        \multicolumn{16}{c}{\textcolor{black!55}{\textbf{\textit{Several Large Parameter VLMs}}}} \\
        \midrule
        GPT-5.1 & 23.22 & 47.22 & 70.16 & 67.00 & 65.25 & 45.86 & 47.65 & 29.90 & 84.22 & 71.60 & 32.67 & 13.92 & 28.97 & 31.01 & 48.88 \\
        Qwen3-VL-235B-A22B & 40.51 & 49.11 & 87.96 & 84.50 & 81.62 & 58.16 & 53.58 & 46.41 & 86.72 & 76.33 & 64.00 & 18.30 & 35.50 & 60.58 & 61.75 \\
        Qwen3.5-122B-A10B & 41.11 & 51.83 & 84.82 & 86.50 & 83.75 & 57.89 & \underline{54.07} & 44.85 & 88.06 & 74.60 & 70.00 & 19.07 & 38.80 & 65.22 & 63.04 \\
        Qwen3.5-397B-A17B & 45.16 & 54.91 & 87.43 & 86.50 & \underline{85.12} & 57.06 & 53.33 & 49.51 & \textbf{89.57} & 76.00 & 72.67 & 20.36 & \textbf{47.50} & \textbf{72.17} & 65.55 \\
        GLM-5.3-Flash~(320B) & 45.54 & 56.80 & 89.01 & 87.00 & 84.75 & \underline{62.12} & \textbf{60.25} & 48.35 & \underline{88.94} & \underline{79.47} & \underline{74.67} & 20.10 & 41.90 & 65.51 & 66.07 \\
        Gemini-3-Flash & 51.27 & 59.29 & 86.39 & \underline{87.88} & 85.00 & 55.86 & 48.64 & 49.71 & \textbf{89.57} & \textbf{83.60} & \textbf{82.67} & \textbf{34.51} & \underline{43.50} & \underline{71.01} & \textbf{67.51} \\
        \midrule
        \multicolumn{16}{c}{\textcolor{black!55}{\textbf{\textit{Our Experimental Models}}}} \\
        \midrule
        Qwen3.5-4B & 30.03 & 48.88 & 81.68 & 81.25 & 77.63 & 57.89 & 49.38 & 43.30 & 85.99 & 66.80 & 57.33 & 14.69 & 33.10 & 54.20 & 57.86 \\
        \quad w/ ZwZ 9K & 40.99 & 57.63 & 86.91 & 81.25 & 79.00 & 57.43 & 45.19 & 52.43 & 86.58 & 68.93 & 53.33 & 10.82 & 30.43 & 56.23 & 58.94 \\
        \quad w/ ZwZ Align 9K & 37.09 & 54.08 & 86.39 & 83.50 & 79.12 & 59.41 & 46.67 & 51.26 & 87.15 & 70.73 & 58.67 & 15.21 & 33.30 & 52.46 & 59.84 \\
        \quad w/ ZoomGain 9K & \textbf{55.81} & \underline{59.88} & 90.05 & 87.25 & 83.88 & 61.59 & 49.14 & \textbf{57.28} & 88.02 & 72.53 & 61.33 & \underline{21.13} & 36.23 & 63.19 & 63.96 \\
        \rowcolor{deltabg}
        \multicolumn{1}{c|}{\cellcolor{deltabg}$\Delta_{\mathrm{ZoomGain-ZwZ}}$} & \deltaUp{+14.83} & \deltaUp{+2.25} & \deltaUp{+3.14} & \deltaUp{+6.00} & \deltaUp{+4.88} & \deltaUp{+4.17} & \deltaUp{+3.95} & \deltaUp{+4.85} & \deltaUp{+1.44} & \deltaUp{+3.60} & \deltaUp{+8.00} & \deltaUp{+10.31} & \deltaUp{+5.80} & \deltaUp{+6.96} & \deltaUp{+5.03} \\
        \rowcolor{deltabg}
        \multicolumn{1}{c|}{\cellcolor{deltabg}$\Delta_{\mathrm{ZoomGain-Base}}$} & \deltaUp{+25.78} & \deltaUp{+11.00} & \deltaUp{+8.37} & \deltaUp{+6.00} & \deltaUp{+6.25} & \deltaUp{+3.70} & \deltaDown{-0.24} & \deltaUp{+13.98} & \deltaUp{+2.03} & \deltaUp{+5.73} & \deltaUp{+4.00} & \deltaUp{+6.44} & \deltaUp{+3.13} & \deltaUp{+8.99} & \deltaUp{+6.10} \\
        \midrule
        Qwen3.5-9B & 32.13 & 50.30 & 86.91 & 82.13 & 79.63 & 58.52 & 49.88 & 44.66 & 86.27 & 71.73 & 60.00 & 15.98 & 34.20 & 56.52 & 59.75 \\
        \quad w/ ZwZ 9K & 44.07 & 58.93 & 90.58 & 85.50 & 82.38 & 61.70 & 47.90 & 52.62 & 88.16 & 73.27 & 63.33 & 17.27 & 34.13 & 57.10 & 62.53 \\
        \quad w/ ZwZ Align 9K & 44.32 & 57.86 & \underline{91.62} & 85.25 & 82.38 & 61.23 & 49.38 & 51.26 & 88.27 & 73.27 & 67.33 & 18.04 & 34.07 & 58.26 & 62.94 \\
        \quad w/ ZoomGain 9K & \underline{55.48} & \textbf{60.36} & \textbf{93.19} & \textbf{90.25} & \textbf{85.50} & \textbf{63.31} & 53.33 & \underline{56.31} & 88.58 & 73.87 & 70.00 & 17.53 & 39.00 & 68.12 & \underline{66.10} \\
        \rowcolor{deltabg}
        \multicolumn{1}{c|}{\cellcolor{deltabg}$\Delta_{\mathrm{ZoomGain-ZwZ}}$} & \deltaUp{+11.41} & \deltaUp{+1.43} & \deltaUp{+2.62} & \deltaUp{+4.75} & \deltaUp{+3.13} & \deltaUp{+1.62} & \deltaUp{+5.43} & \deltaUp{+3.69} & \deltaUp{+0.42} & \deltaUp{+0.60} & \deltaUp{+6.67} & \deltaUp{+0.26} & \deltaUp{+4.87} & \deltaUp{+11.01} & \deltaUp{+3.58} \\
        \rowcolor{deltabg}
        \multicolumn{1}{c|}{\cellcolor{deltabg}$\Delta_{\mathrm{ZoomGain-Base}}$} & \deltaUp{+23.35} & \deltaUp{+10.06} & \deltaUp{+6.28} & \deltaUp{+8.12} & \deltaUp{+5.87} & \deltaUp{+4.79} & \deltaUp{+3.45} & \deltaUp{+11.65} & \deltaUp{+2.31} & \deltaUp{+2.14} & \deltaUp{+10.00} & \deltaUp{+1.55} & \deltaUp{+4.80} & \deltaUp{+11.60} & \deltaUp{+6.35} \\
        \bottomrule
    \end{tabular}}
    
\end{table}

\paragraph{Results.}
As shown in Table~\ref{tab:training-results},
models trained on ZoomGain data consistently outperform their ZwZ-trained counterparts on all 14 benchmarks.
Besides,
ZoomGain also outperforms ZwZ Align on all 13 held-out benchmarks for the 4B model
and on 12 of 13 for the 9B model.
Therefore, these gains can be attributed to the questions constructed by ZoomGain.
Moreover, the improvements brought by ZoomGain extend across all benchmark groups. 
Compared with the base models, the gains of ZwZ are mainly concentrated on fine-grained perception.
In contrast, ZoomGain improves all three groups.
The advantage of ZoomGain over ZwZ is most pronounced on difficult perception and reasoning,
where the 9B model gains 11.01 points on O3-Bench and the 4B model gains 10.31 points on BabyVision.
These results indicate that ZoomGain data
enables models to acquire transferable fine-grained perception capabilities.

Remarkably, with only 9K training samples,
the 9B model achieves an average score of 66.10,
surpassing Qwen3.5-397B-A17B (65.55) and performing on par with GLM-5.3-Flash (66.07),
while the 4B model (63.96) also surpasses Qwen3.5-122B-A10B and the ZwZ-trained 9B model.
Moreover, ZoomGain-trained models achieve the best results among all compared models on 6 of 7 fine-grained perception benchmarks.
However, we also notice that the strongest large models still lead on
general perception and difficult perception and reasoning,
suggesting that these tasks still rely heavily on model scale and reasoning capability.
We also show the results on Gemma-4-E4B-it~(\cite{gemmateam2026gemma4technicalreport}) in Appendix~\ref{app:gemma}.

\subsection{Ablation Study}
\label{sec:training-ablation}

\paragraph{Setup.}
We ablate two core designs of ZoomGain on Qwen3.5-4B.
All variants use 9K training samples,
with the same RL settings and evaluation protocol
as in Table~\ref{tab:training-results}.
\textbf{(1)} Remove incremental evidence:
instead of requiring questions to be based on incremental information
that appears only in the zoomed-view caption,
we generate questions directly from the zoomed-view caption.
The source images, candidate QA pool size, and 9K sample selection procedure
are otherwise identical to those of the full version.
\textbf{(2)} Remove dedicated distractors:
we keep each question, its answer, the correct option's position,
and the number of options unchanged,
and replace only the distractors of each multiple-choice question
with correct answers from other training questions in the same answer category
(e.g., color or material).
% If a replacement matches any of the question's original distractors,
% we replace it again until none of the new distractors matches an original one.
OCR and counting questions are open-ended and are unaffected by this ablation.

\begin{table}[t]
    \centering
    \definecolor{ablationgreen}{RGB}{0,183,0}
    \definecolor{ablationred}{RGB}{255,0,0}
    \caption{\textbf{Ablation of incremental evidence and dedicated distractors.} Avg.\ is computed over the 10 held-out benchmarks.}
    \label{tab:training-ablation}
    \setlength{\tabcolsep}{2.5pt}
    \renewcommand{\arraystretch}{1.12}
    \resizebox{\linewidth}{!}{%
    \begin{tabular}{lcc|c|ccccccc|ccc|c}
        \toprule
        \multirow{2}{*}[-2ex]{\textbf{Model}}
        & \multirow{2}{*}[-2.5ex]{\shortstack{\textbf{Incremental}\\\textbf{Evidence}}}
        & \multirow{2}{*}[-2.5ex]{\shortstack{\textbf{Dedicated}\\\textbf{Distractors}}}
        & \textbf{Held-in}
        & \multicolumn{7}{c|}{\textbf{Fine-grained Perception}}
        & \multicolumn{3}{c|}{\textbf{Difficult Perception \& Reason}}
        & \multirow{2}{*}[-2ex]{\shortstack{\textbf{Avg.}\\{\scriptsize Only Held-out}}} \\
        \cmidrule(lr){4-4}\cmidrule(lr){5-11}\cmidrule(lr){12-14}
        & &
        & \shortstack{ZoomGain\\Bench}
        & \shortstack{Zoom\\Bench} & \shortstack{V*\\Bench}
        & \shortstack{HR\\4K} & \shortstack{HR\\8K}
        & \shortstack{MME-RW\\Lite} & \shortstack{Tree\\Bench} & \shortstack{Visual\\Probe}
        & \shortstack{Baby\\Vision} & \shortstack{Perception\\Bench}
        & \shortstack{O3-\\Bench} & \\
        \midrule
        Qwen3.5-4B & -- & -- & 30.03 & 48.88 & 81.68 & 81.25 & 77.63 & 57.89 & \textbf{49.38} & 43.30 & 14.69 & 33.10 & 54.20 & 54.20 \\
        \quad +ZoomGain 9K & \textcolor{ablationred}{\xmark} & \textcolor{ablationgreen}{\cmark} & 49.71 & 56.45 & 88.48 & 85.50 & 81.75 & 60.24 & 48.40 & 52.82 & 18.56 & 34.70 & 58.84 & 58.57 \\
        \quad +ZoomGain 9K & \textcolor{ablationgreen}{\cmark} & \textcolor{ablationred}{\xmark} & 46.85 & 57.04 & \textbf{90.05} & 85.75 & 80.25 & \textbf{62.69} & 48.89 & 54.76 & 17.53 & 34.97 & 59.42 & 59.14 \\
        \quad +ZoomGain 9K & \textcolor{ablationgreen}{\cmark} & \textcolor{ablationgreen}{\cmark} & \textbf{55.81} & \textbf{59.88} & \textbf{90.05} & \textbf{87.25} & \textbf{83.88} & 61.59 & 49.14 & \textbf{57.28} & \textbf{21.13} & \textbf{36.23} & \textbf{63.19} & \textbf{60.96} \\
        \bottomrule
    \end{tabular}}
    
\end{table}

\paragraph{Results.}
As shown in Table~\ref{tab:training-ablation}, the results demonstrate that both designs contribute to training performance.
The full ZoomGain pipeline achieves the highest average score across the 10 held-out benchmarks.
Removing any component incude the performance losses on almost all benchmarks. Both ablations incur the largest losses in the
\textit{difficult perception and reasoning} group.
The full version improves over the base model by an average of 6.19 points
across these three benchmarks,
while each ablation retains only about half of this gain.
This indicates that the two designs mainly improve the capability of perception
and reasoning that transfer beyond the training distribution.

The effect of incremental evidence is most apparent on fine-grained benchmarks:
removing it reduces scores on ZoomBench and VisualProbe by 3.43 and 4.46 points, respectively.
This indicates that restricting questions to evidence revealed only after zooming
forces the model to learn to locate and recognize details hard to discern.
Meanwhile, removing dedicated distractors causes the largest drop
on the held-in ZoomGainBench, at 8.96 points.
It is expected because the benchmark's options are constructed using the same pipeline.
Using dedicated distractors introduces harder confusions,
forcing the model to attend more closely to visual details and thereby enhancing its perception capabilities under the same data budget.
Combining distractor sources also matters:
on a 5,050-sample subset, using only full-image errors as distractors
lowers the 13-benchmark average from 63.19 to 61.12 (Appendix~\ref{app:distractor-source}).
 
\subsection{\texorpdfstring{Do Caption Gaps Reflect Perception Gaps?}{Do Caption Gaps Reflect Perception Gaps?}}
\label{sec:caption-gap}

\begin{table}[t]
    \centering
    \caption{\textbf{Caption gaps correspond to perception gaps.}
    $\Delta$ = Crop $-$ Full w/ Box.}
    \label{tab:caption-gap}
    \scriptsize 
    \setlength{\tabcolsep}{7pt}
    \renewcommand{\arraystretch}{1}
    \begin{tabular}{lcccccc}
        \toprule 
        & \multicolumn{3}{c}{\textbf{Crop-only}} & \multicolumn{3}{c}{\textbf{ZoomGain incremental}} \\
        \cmidrule(lr){2-4}\cmidrule(lr){5-7}
        \textbf{Model} & Full w/ Box & Crop & $\Delta$ & Full w/ Box & Crop & $\Delta$ \\
        \midrule
        Qwen3.5-4B                 & 73.00 & 85.33 & +12.33 & 36.67 & 88.00 & +51.33 \\
        \quad w/ ZoomGain 9K        & 78.00 & 86.67 & +8.67  & 51.67 & 89.33 & +37.67 \\
        Qwen3.5-9B                 & 78.67 & 86.00 & +7.33  & 43.33 & 90.00 & +46.67 \\
        \quad w/ ZoomGain 9K        & 82.33 & 86.33 & +4.00  & 57.67 & 91.33 & +33.67 \\
        Qwen3.5-35B-A3B            & 75.00 & 85.33 & +10.33 & 51.00 & 91.00 & +40.00 \\
        GLM-5.3-Flash~(320B)       & 76.67 & 89.00 & +12.33 & 53.33 & 94.00 & +40.67 \\
        \bottomrule
    \end{tabular}
    \vspace{-10pt}
\end{table}

Since a caption may omit details its caption model can perceive,
we test whether caption gaps reflect actual perceptual gaps in visual-answering.
On 300 regions from ZoomGainBench, we build two sets of open-ended questions:
(1)\emph{Crop-only}, generated from the zoomed view alone following the ZwZ prompt~(\cite{wei2026zooming}),
and (2)\emph{ZoomGain incremental}, generated from the incremental facts $\Delta_i$.
Models answer each question on the full image with the target region marked by a red box (\emph{Full w/ Box})
and on the zoomed view (\emph{Crop}),
so the difference measures how much zooming helps perception once the target is located.

As shown in Table~\ref{tab:caption-gap}, both sets are well answered on Crop,
but on Full w/ Box crop-only questions are already largely solved (73.00--78.67)
while incremental questions drop to 36.67--53.33.
Zooming thus helps 3 to 6 times more on ZoomGain incremental questions,
demonstrating that the details found by captions gaps
are indeed diffcult for models to perceive in the full image.
After ZoomGain training, performance on ZoomGain incremental questions rises by about 15 points on Full w/ Box
while only 1.33 on Crop,
showing that the models learn to recognize these details in the full image.

\section{Conclusion}
\label{sec:conclusion}

To address the difficulty of constructing reliable fine-grained visual supervision at scale,
we propose ZoomGain, a data construction pipeline that exploits the perception gap
between global and zoomed-in views.
ZoomGain turns details revealed only by zooming into questions about the full image,
verifies their answers on the zoomed view with multiple models,
and mines multi-source distractors that could mislead VLMs.
With this pipeline, we construct ZoomGainBench, a challenging benchmark
whose questions and answers are entirely model-authored. Through the analysis of failure cases, we find that different VLMs share similar confusions
when fine details are unresolved in the full image.
Experiments further show the effectiveness of ZoomGain data as training supervision,
which consistently improves VLMs' capabilities of fine-grained perception,
general perception, and difficult perception and reasoning.

% Enable the bibliography when citations are added to the paper.
\bibliography{references}
\bibliographystyle{antgroup}

% \clearpage
\appendix
\clearpage
\section{Complete Algorithm and Additional Implementation Details}
\label{app:construction-algorithm}

Section~\ref{sec:pipeline} describes the core pipeline.
Here we add two implementation details, incorporated into Algorithm~\ref{alg:zoomgain-construction}.
Model assignments and thresholds are given in Appendix~\ref{app:construction-settings};
the prompts are collected in Appendix~\ref{app:construction-prompts}.

\begin{figure}[H]
    \centering
    \includegraphics[width=0.72\linewidth]{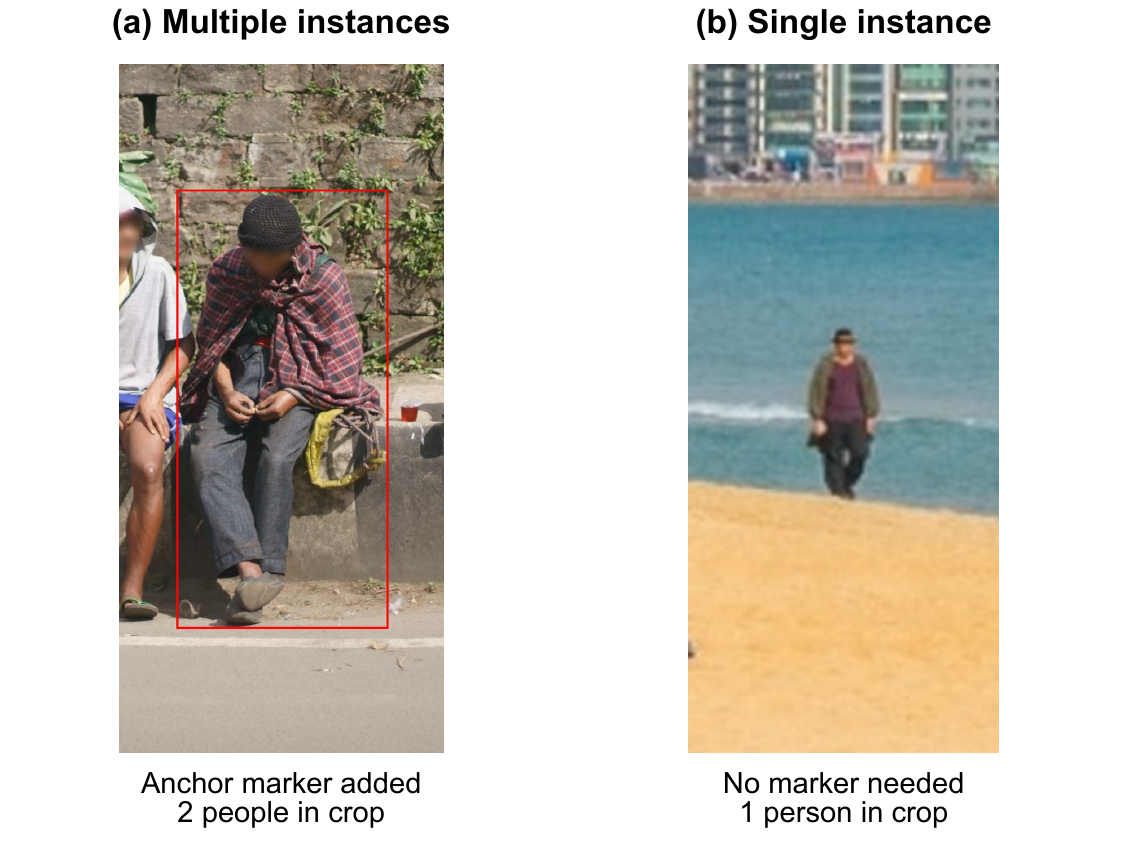}
    \caption{\textbf{Conditional anchor markers in the zoomed view.}
    (a) Two people are present, including the partially visible person on the left;
    the red rectangle selects the person wearing a red-and-black plaid shawl.
    (b) Only one person is present, so no marker is added.
    Both views are rendered from the original images and recorded crop/anchor coordinates
    using the pipeline's marker and resizing functions.
    The rectangle identifies the anchor rather than the answer-bearing detail;
    it is used only in the zoomed view and is not part of the photographed scene.}
    \label{fig:anchor-marker-examples}
\end{figure}

\paragraph{Anchor markers in the zoomed view.}
The expanded crop $V_z^i$ may contain several objects of the anchor's category.
A VLM enumerates these instances, including partially visible ones, and matches the selected anchor to its grounded box.
If multiple instances are present, we draw a red rectangle around the anchor itself;
if only one is present, the crop remains unmarked.
This rectangle is distinct from $b_i$, which marks the entire crop region in the global view.
The marked-view caption prompt uses it only to identify the anchor and otherwise follows the unmarked-view instructions.
It must not be described as photographed content or used as an answer.
Local questions may refer to the marker; global questions use natural referring expressions.
Figure~\ref{fig:anchor-marker-examples} shows both cases.

\paragraph{Joint crop/full-image audit and rewriting.}
After QA generation, \texttt{EDGE\_REWRITER} inspects the crop and the full photograph with $b_i$ marked,
together with both question forms and the reference answer.
It checks whether answer-bearing text or objects continue beyond the crop,
whether the global referent is unique, and whether both questions admit the same answer.
It returns \emph{keep}, \emph{rewrite}, or \emph{drop}.
A bounded rewrite can add answer-free qualifiers valid in both views while preserving the non-OCR answer.
For OCR, it may narrow the question and answer to a complete text block or specified prefix/suffix
contained in the original answer, but cannot add text read from the full image.
Counting questions whose requested set extends outside the crop are discarded.
Revisions are synchronized across the local/global and English/Chinese questions,
and target and answer constraints are checked again before crop-based verification
(Section~\ref{sec:verification-distractors}).
Both anchor enumeration and this audit use Qwen3.5-397B-A17B.

\clearpage
\begin{algorithm}[H]
    \caption{\textit{ZoomGain Data Construction}}
    \label{alg:zoomgain-construction}
    \footnotesize
    \algrenewcommand\algorithmiccomment[1]{\hfill\textcolor{black!50}{// #1}}
    \begin{algorithmic}[1]
        \Require Image pool $\mathcal{I}$; configured models; $N=10$ samples per model; $R=3$ MCQ verification rounds
        \Ensure Verified QA dataset $\mathcal{D}$ with evidence regions and distractor provenance
        \State $\mathcal{D} \gets \emptyset$
        \For{each full image $V_g \in \mathcal{I}$}
            \State $C_g \gets \Call{SceneCaption}{V_g}$; $\mathcal{E} \gets \Call{ProposeAnchors}{C_g}$
            \For{each referring expression $e_i \in \mathcal{E}$}
                \Statex \textcolor{black!50}{\textit{// Anchor discovery and grounding}}
                \State $\mathcal{B}_i \gets \Call{GroundIndependently}{V_g,e_i}$
                \State \textbf{if} $\neg\Call{ValidGrounding}{\mathcal{B}_i,0.5}$ \textbf{then continue}
                \State $b_i \gets \Call{Expand}{\mathrm{median}_{\mathrm{coord}}(\mathcal{B}_i)}$; $V_z^i \gets \Call{Crop}{V_g,b_i}$
                \State $L_i \gets \Call{EnumerateAnchorInstances}{V_z^i,e_i}$
                \State \textbf{if} no instance matches the grounded anchor \textbf{then continue}
                \State $\widetilde V_z^i \gets \Call{MarkAnchorIfMultiple}{V_z^i,L_i}$
                \Statex \textcolor{black!50}{\textit{// Global-to-zoom evidence and QA construction}}
                \State $V_g^i \gets \Call{DrawBox}{V_g,b_i}$
                \State $C_g^i \gets M_{\mathrm{cap}}(V_g^i,b_i)$; $C_z^i \gets M_{\mathrm{cap}}(\widetilde V_z^i)$ \Comment{Independent calls}
                \State $\Delta_i \gets \Call{IncrementalFacts}{C_g^i,C_z^i}$
                \State $(q_{\mathrm{local}}^i,q_{\mathrm{global}}^i,a_i,t_i) \gets \Call{GenerateQA}{\Delta_i,e_i,L_i}$
                \State \textbf{if} no supported, unambiguous QA is generated \textbf{then continue}
                \State $(v,q_{\mathrm{local}}^i,q_{\mathrm{global}}^i,a_i) \gets
                    \Call{EdgeAuditAndRewrite}{\widetilde V_z^i,V_g^i,q_{\mathrm{local}}^i,q_{\mathrm{global}}^i,a_i}$
                \State \textbf{if} $v=\mathrm{drop}$ or post-rewrite constraints fail \textbf{then continue}
                \Statex \textcolor{black!50}{\textit{// Quality verification and multi-source distractors}}
                \State $A_{\mathrm{crop}} \gets \Call{Sample}{\mathrm{crop},\widetilde V_z^i,q_{\mathrm{local}}^i,N}$
                \State \textbf{if} $\min_m \Call{Correct}{A_{\mathrm{crop}}^m,a_i}<6$ \textbf{then continue}
                \State $A_{\mathrm{blind}} \gets \Call{Sample}{\mathrm{blind},\emptyset,q_{\mathrm{global}}^i,N}$
                \State \textbf{if} $\min_m \Call{Correct}{A_{\mathrm{blind}}^m,a_i}\geq 3$ \textbf{then continue}
                \State $O_i \gets \emptyset$; $S_i \gets \emptyset$
                \If{$t_i=\mathrm{attribute}$}
                    \State $A_{\mathrm{global}} \gets \Call{Sample}{\mathrm{global},V_g,q_{\mathrm{global}}^i,N}$
                    \State $D_v \gets \Call{VisualAlternatives}{V_g,\widetilde V_z^i,q_{\mathrm{local}}^i,q_{\mathrm{global}}^i,a_i}$
                    \State $D_b \gets \Call{Incorrect}{A_{\mathrm{blind}},a_i}$; $D_g \gets \Call{Incorrect}{A_{\mathrm{global}},a_i}$
                    \State $(O_i,S_i) \gets \Call{CurateOptions}{a_i,D_v,D_b,D_g}$
                    \State $\mathrm{verified} \gets \mathrm{false}$
                    \For{$r=1,\ldots,R$}
                        \State $A_{\mathrm{mcq}} \gets \Call{Sample}{\mathrm{mcq},\widetilde V_z^i,(q_{\mathrm{local}}^i,q_{\mathrm{global}}^i,O_i),N}$
                        \If{$\min_m \Call{Correct}{A_{\mathrm{mcq}}^m,a_i}\geq 6$}
                            \State $\mathrm{verified} \gets \mathrm{true}$; \textbf{break}
                        \EndIf
                        \State \textbf{if} $r<R$ \textbf{then} $(O_i,S_i) \gets \Call{RepairDistractors}{O_i,S_i}$
                    \EndFor
                    \State \textbf{if} $\neg\mathrm{verified}$ \textbf{then continue}
                \EndIf
                \State $\mathcal{D} \gets \mathcal{D}\cup\{(V_g,V_z^i,b_i,q_{\mathrm{local}}^i,q_{\mathrm{global}}^i,a_i,t_i,O_i,S_i)\}$
            \EndFor
        \EndFor
        \State \Return $\mathcal{D}$
    \end{algorithmic}
\end{algorithm}

\paragraph{Notation and operations.}
$L_i$ records the anchor instances; $\widetilde V_z^i$ is the optionally marked zoomed view.
\textsc{EdgeAuditAndRewrite} returns a verdict $v$ and the retained or revised QA under the constraints above.
Both question forms have English and Chinese versions.
The remaining operations follow Section~\ref{sec:pipeline}, with verification settings in
Appendix~\ref{app:construction-settings}; $S_i$ preserves distractor provenance during repairs.

\clearpage
\section{Construction Settings}
\label{app:construction-settings}

Table~\ref{tab:construction-model-roles} provides the complete model assignments
for the ZoomGain pipeline described in Section~\ref{sec:data-preparation}.

\begin{table}[htbp]
    \centering
    \caption{\textbf{Model assignments in the ZoomGain construction pipeline.}}
    \label{tab:construction-model-roles}
    \small
    \renewcommand{\arraystretch}{1.15}
    \begin{tabular}{@{}p{0.32\linewidth}p{0.64\linewidth}@{}}
        \toprule
        Model & Roles \\
        \midrule
        Qwen3.5-397B-A17B~(\cite{qwen35blog})
        & Initial global scene captioning and paired-view captioning ($M_{\mathrm{cap}}$);
          grounding; crop anchor enumeration; joint crop/full-image audit and rewriting;
          blind answering; global-view error collection;
          visual distractor generation; final MCQ verification. \\
        GLM-5.2~(\cite{glm5team2026glm5vibecodingagentic})
        & Incremental fact extraction and QA generation. \\
        Qwen3-VL-235B-A22B-Instruct~(\cite{Qwen3-VL})
        & Grounding; crop-based verification; blind answering;
          global-view error collection; final MCQ verification. \\
        Kimi-K2.6~(\cite{kimiteam2026kimik25visualagentic})
        & Grounding; crop-based verification; blind answering;
          global-view error collection; final MCQ verification. \\
        Qwen3.5-122B-A10B~(\cite{qwen35blog})
        & Answer-equivalence judging. \\
        DeepSeek-V4-Pro~(\cite{deepseekai2026deepseekv4})
        & Candidate distractor clustering and option-pool curation. \\
        \bottomrule
    \end{tabular}
\end{table}

\paragraph{Verification Thresholds.}
All verification stages sample 10 answers per model.
Crop-based verification requires at least 6 correct answers from each of its two models,
Qwen3-VL-235B-A22B-Instruct and Kimi-K2.6.
Blind verification rejects a question only when all three models each answer correctly at least 3 times.
Final MCQ verification requires at least 6 correct selections from each of its three models.

\paragraph{Verification Rounds and Distractor Repair.}
Final MCQ verification uses at most three rounds:
an initial check followed by up to two further checks after distractor repairs.
Each round applies the per-model threshold described above.
Candidates that still fail after the final round are discarded.

\clearpage
\section{Benchmark Question Categories}
\label{app:benchmark-categories}

\begin{figure}[!htbp]
    \centering
    \includegraphics[width=0.88\linewidth]{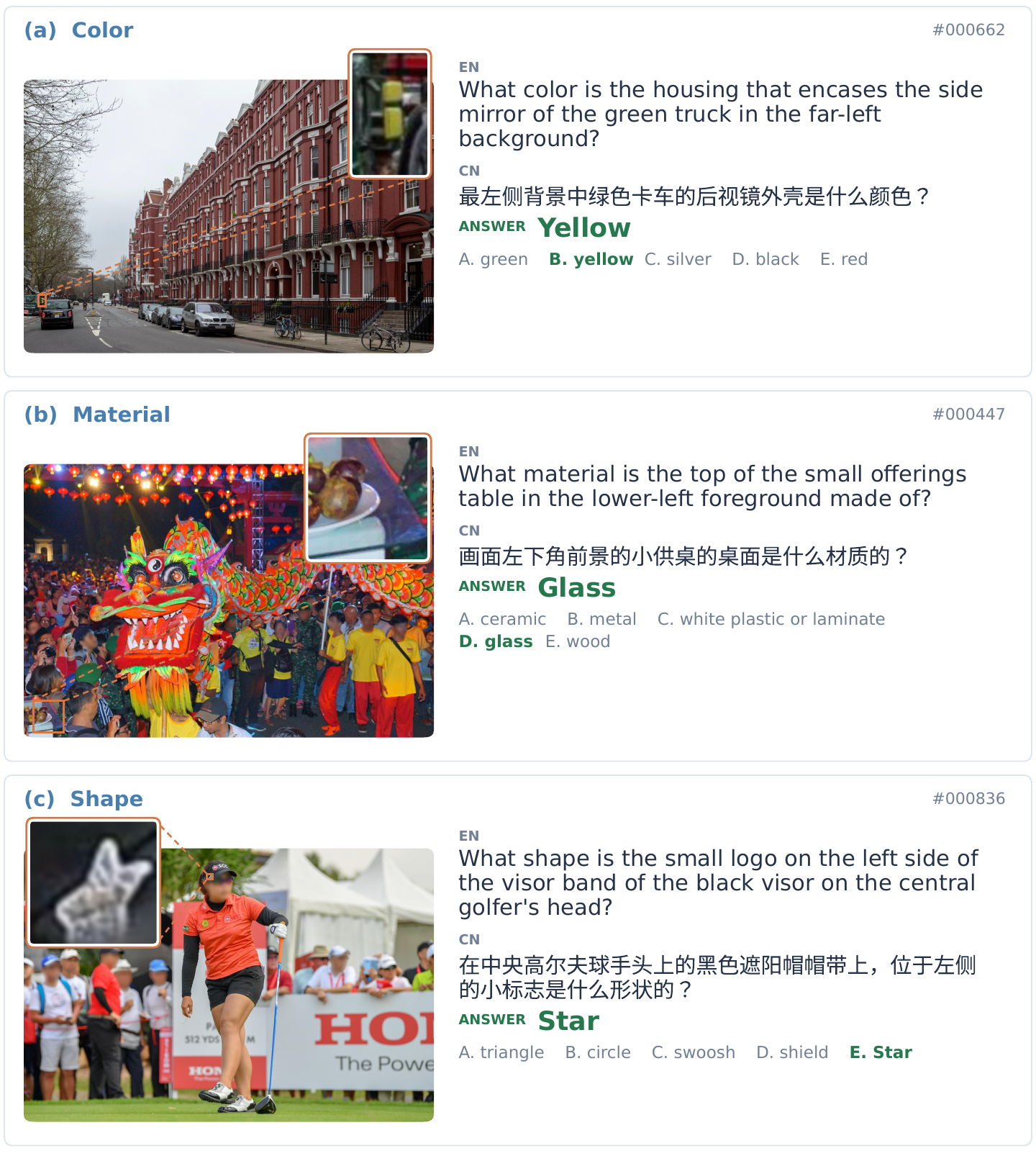}
    \caption{\textbf{Attribute recognition examples.}
    (a) Mirror housing color; (b) tabletop material; (c) logo shape.
    Correct answers and their corresponding options are highlighted in green.}
    \label{fig:category-attribute}
\end{figure}

\begin{figure}[!htbp]
    \centering
    \includegraphics[width=\linewidth]{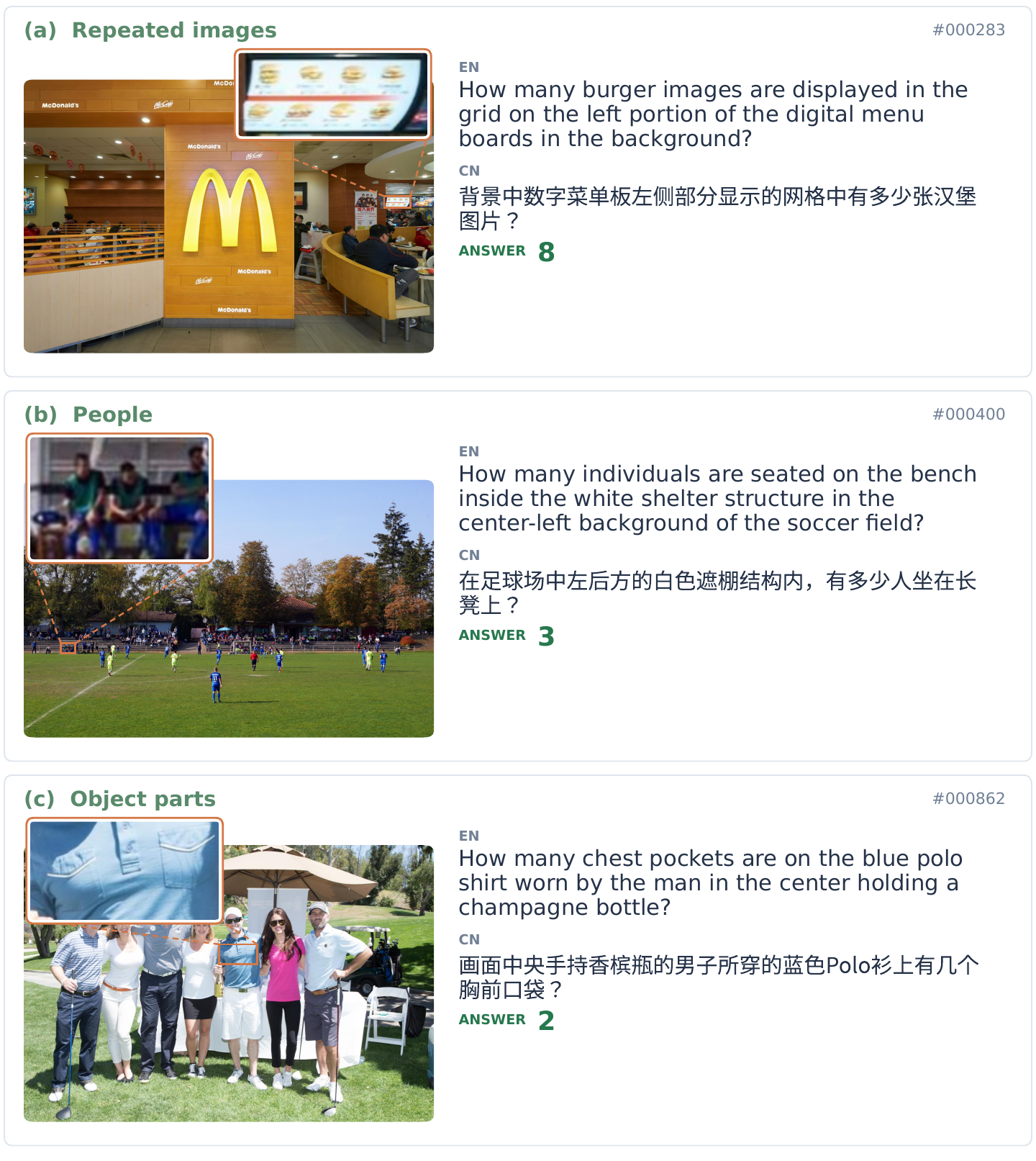}
    \caption{\textbf{Counting examples.}
    (a) Eight burger images on a menu; (b) three people seated inside a shelter;
    (c) two chest pockets on a shirt.
    Each question requires an open-ended numerical answer.}
    \label{fig:category-counting}
\end{figure}

\begin{figure}[!htbp]
    \centering
    \includegraphics[width=\linewidth]{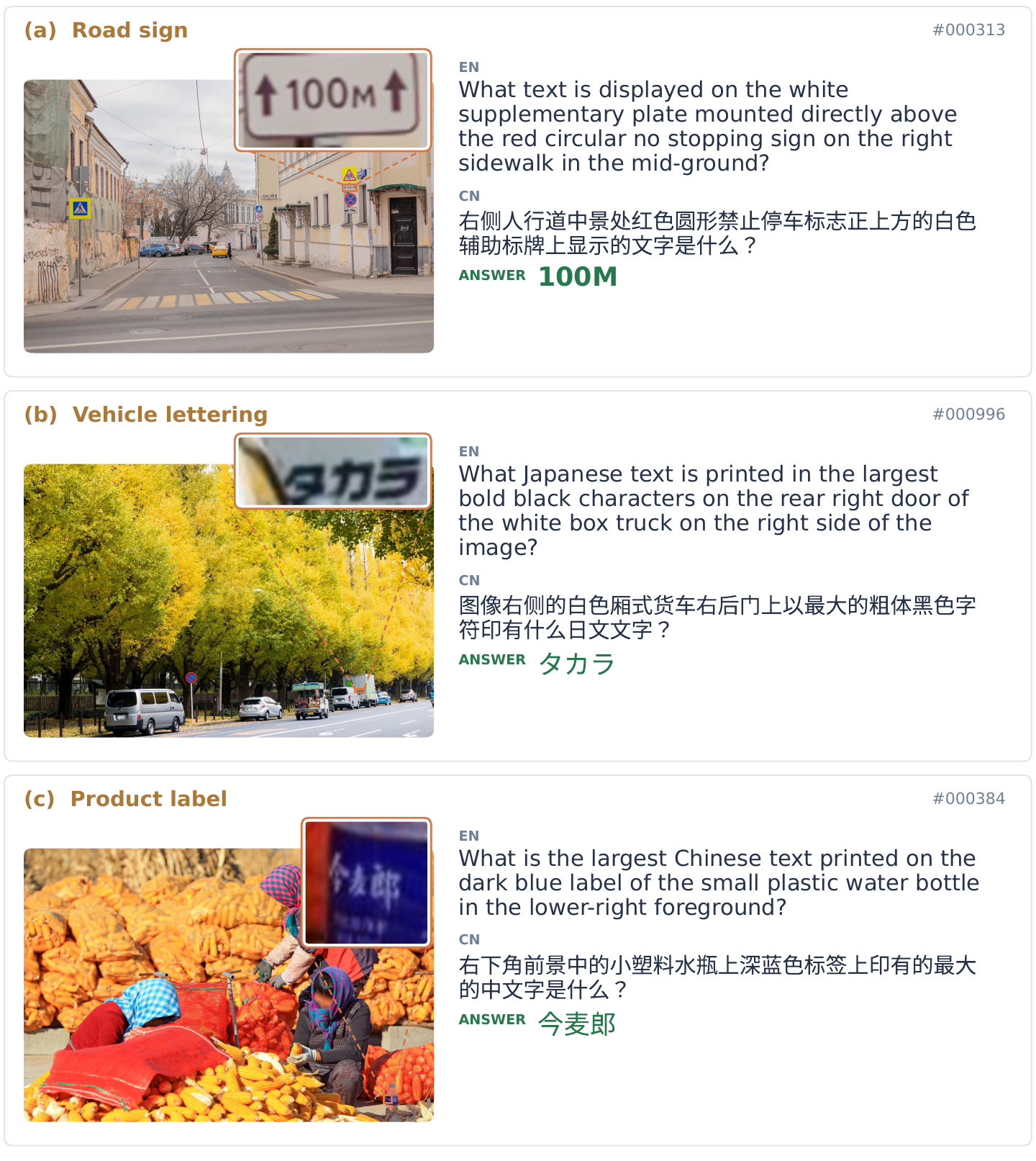}
    \caption{\textbf{Text recognition examples.}
    (a) A supplementary road sign; (b) lettering on a truck's rear door;
    (c) the largest Chinese text on a bottle label.
    Reference transcriptions are shown in green.}
    \label{fig:category-ocr}
\end{figure}

The three question categories in ZoomGainBench cover the following fine-grained details.
Figures~\ref{fig:category-attribute}--\ref{fig:category-ocr} show three examples per category,
with paired English and Chinese questions.
The orange boxes mark the specific details needed to answer each question,
with dashed lines connecting them to enlarged views.
These boxes are selected for illustration and are distinct from the pipeline's anchor boxes.

\paragraph{Attribute Recognition.}
Questions concern a target's color, material, shape, object identity,
actions, or relations.
The examples below ask about a truck's mirror housing, a glass tabletop,
and a small logo on a golfer's visor.

\clearpage
\paragraph{Counting.}
Questions ask for the number of entities, object parts, or repeated patterns.
The examples below cover repeated images on a menu, people seated on a bench,
and pockets on a shirt.

\clearpage
\paragraph{Text Recognition (OCR).}
Questions require transcribing text on signs, vehicles, clothing, printed materials, and product labels.
The examples below include a road sign, Japanese lettering on a truck,
and Chinese text on a bottle label.
Answers retain the original text visible in the image, regardless of the question language.

\clearpage
\section{\texorpdfstring{Language-Prior Shortcuts in Existing Benchmarks}{Language-Prior Shortcuts in Existing Benchmarks}}
\label{app:language-prior}

As verified from their official papers,
existing fine-grained benchmarks such as ZoomBench, V* Bench, HR-Bench, and TreeBench
do not include a blind-filtering stage during data construction.
To directly show that language-prior shortcuts,
we remove both the image and all multiple-choice options,
feeding only the question text to Qwen3.5-4B and Qwen3.5-9B in an open-ended blind setting.
As shown in Table~\ref{tab:language-prior},
even without seeing the image or options,
both models directly output the exact ground-truth answers through common-sense priors alone.
(these cases are illustrative examples rather than an exhaustive list.)
When multiple-choice options are further provided in standard evaluation,
guessing such questions without perceiving the visual detail becomes even easier.
By contrast, ZoomGain's blind verification (Section~\ref{sec:verification-distractors};
see rejected candidates in Figure~\ref{fig:blind-verification-examples})
explicitly removes such questions so that answering requires visual perception.

\begin{table}[htbp]
    \centering
    \caption{\textbf{Questions in existing benchmarks answerable from language priors.}
    Only the question text is given, without the image or options.}
    \label{tab:language-prior}
    \small
    \renewcommand{\arraystretch}{1.15}
    \begin{tabular}{@{}l p{0.48\linewidth} c c@{}}
        \toprule
        \textbf{Benchmark} & \textbf{Question stem} & \textbf{Ground truth} & \textbf{Blind output (4B / 9B)} \\
        \midrule
        ZoomBench & What type of jewelry is the person wearing on their finger? & Ring & Ring / Ring \\
        ZoomBench & What is the person doing with the watering can? & Water plants & \shortstack{Watering plants /\\Watering plants} \\
        V* Bench  & What is the cartoon character on the clock. & Mickey Mouse & \shortstack{Mickey Mouse /\\Mickey Mouse} \\
        V* Bench  & What is the color of the lock? & Silver & Silver / Silver \\
        HR-Bench  & Who is the founder of the company that makes the square white adapter shown in the image? & Steve Jobs & Steve Jobs / Steve Jobs \\
        HR-Bench  & What is the material of the ceiling beams in the room? & Wood & Wood / Wood \\
        \bottomrule
    \end{tabular}
\end{table}

\clearpage
\section{Additional Error Analysis}
\label{app:error-analysis}

This section complements Section~\ref{sec:error-analysis}.

\paragraph{Per-model consistency.}
Every evaluated model over-selects global-view errors:
they receive 65.0--78.2\% of each model's incorrect choices,
against 52.3--57.8\% expected under uniform choice among the distractors of each question.
For the four closed-source models, which took no part in constructing the benchmark,
the shares are 75.7\% versus 57.4\%.
Their selection rate of a global-view error also rises with how consistently it arose during construction,
from 13.4\% when given by one or two of the 30 full-image answers
to 35.2\% when given more than ten times.

\paragraph{Controlling for option position.}
Models prefer some option positions over others.
Normalizing by the average selection rate of each option position,
global-view errors are still selected 1.34 times as often as expected,
and 1.46 times when they also coincide with blind guesses,
compared with 0.52 and 0.55 times for blind guesses and visual evidence.

\paragraph{Below-chance accuracy of Qwen3-VL-8B.}
Only 1.5\% of Qwen3-VL-8B's answers to attribute questions cannot be parsed,
so its below-chance accuracy does not stem from formatting failures.
Instead, 67.5\% of its incorrect choices fall on global-view errors, against 54.0\% expected.
Its responses typically claim to inspect the target closely
and then state a plausible but incorrect attribute,
for example describing a dark blue backpack as black.

\paragraph{Errors on open-ended questions.}
Counting errors are mostly near misses:
71.6\% of incorrect numeric answers are off by exactly one,
and 67 of the 94 counting questions have a ground truth of at most three,
so the difficulty lies in resolving a few small instances rather than in enumerating many objects.
For OCR, among transcriptions without full credit,
29.1\% remain close to the reference, with a character-level similarity of at least 0.5;
10.1\% omit part of the target text or include adjacent text;
and at least 13.4\% transcribe other text in the scene instead of the target,
as identified by matching predictions against the scene descriptions generated during construction.
Most of the remaining transcriptions bear little resemblance to the reference.

\section{Representative Failure Cases}
\label{app:error-cases}

\begin{figure}[htbp]
    \centering
    \includegraphics[width=\linewidth]{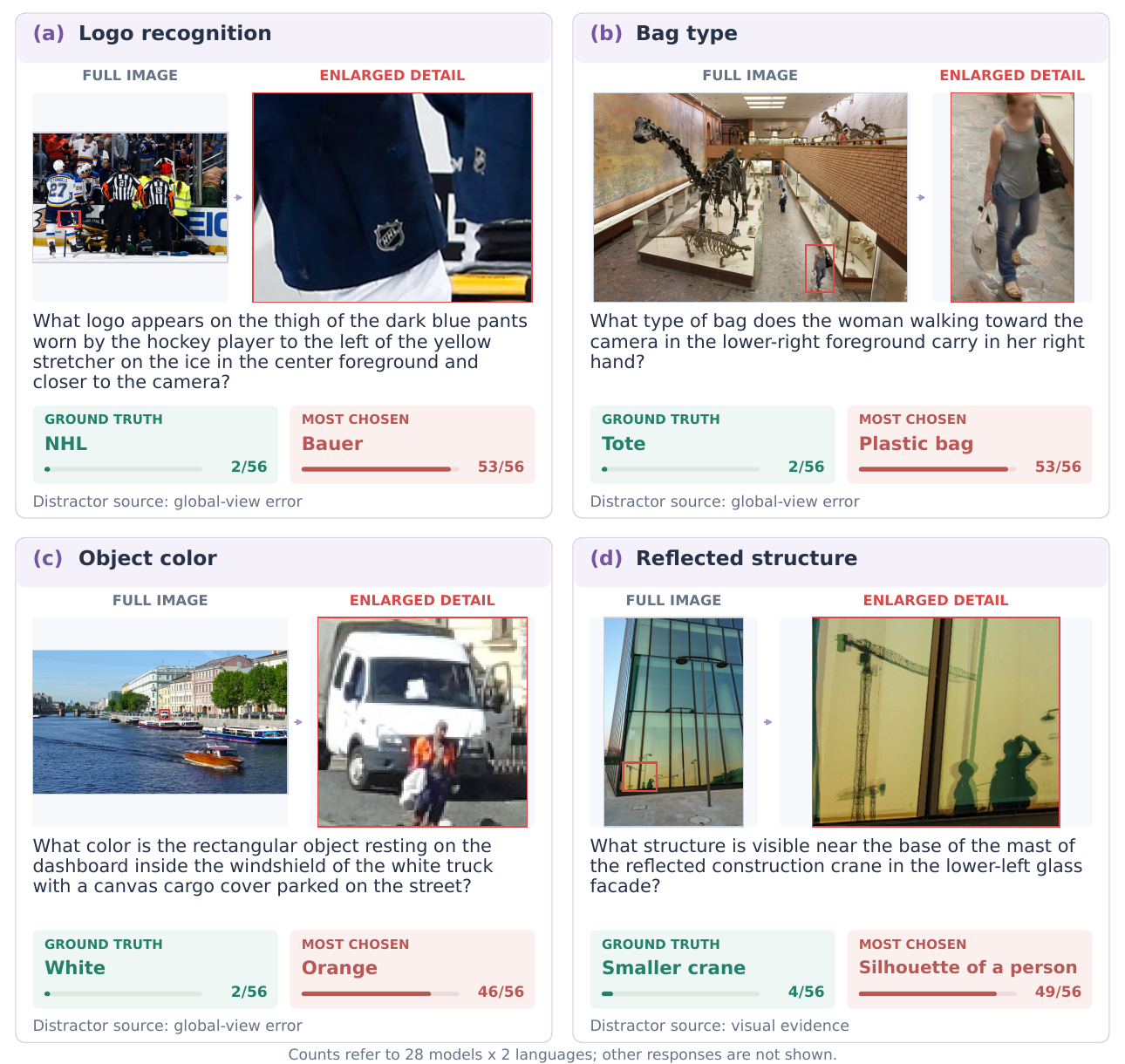}
    \caption{\textbf{Representative failures on ZoomGainBench.}
    Each case shows the full image with the queried detail outlined in red,
    an enlarged view of that region,
    and the responses of the 56 evaluations (28 models $\times$ 2 languages).
    (a) Nearly all evaluations name Bauer, a hockey equipment brand, instead of the NHL logo,
    while the blind guesses (Reebok, Nike) are never chosen.
    (b) The white tote bag is taken for a plastic bag;
    the visual-evidence distractors, drawn from the black bag on her shoulder
    and from another pedestrian's purse, are never chosen.
    (c) Most evaluations (46/56) call the white object on the truck's dashboard orange,
    likely the color of the vest worn by the worker in front of the truck.
    (d) The salient reflected silhouettes draw choices away from the smaller crane at the base of the mast.}
    \label{fig:error-cases}
\end{figure}

Figure~\ref{fig:error-cases} shows representative failures discussed in Section~\ref{sec:error-analysis}.

\section{ZoomGainBench Evaluation Settings}
\label{app:benchmark-eval-settings}

All non-agentic models use the same system prompt as ZwZ~(\cite{wei2026zooming}),
which asks the model to reason before placing its final answer within
\texttt{<answer>} and \texttt{</answer>} tags.
The user message contains the full image and the question with its options, if applicable.
Counting questions additionally request Arabic numerals, while OCR questions request
only the transcription within the answer tags, preserving capitalization, spacing,
punctuation, and symbols.
The Chinese version replaces the question stem while retaining the options and formatting instructions.

Qwen models, ZwZ, and Vision-OPD use a temperature of 0 and a maximum output length
of 2,048 tokens, with native thinking disabled via \texttt{enable\_thinking=false}.
MiMo-VL-7B-RL uses the same temperature and output limit, with \texttt{/no\_think}
appended to the question.
GLM-4.6V and GLM-5.3-Flash use a temperature of 0 and output limits of 4,096 and
8,192 tokens, respectively; their requests also set \texttt{enable\_thinking=false}.
Kimi-K3 is evaluated with thinking enabled, a temperature of 0, and an output limit
of 8,192 tokens.
GPT-5.5, GPT-5.6-Sol, Gemini-3.6-Flash, and Claude-Opus-5 use an output limit of
4,096 tokens and retain the provider defaults for temperature and thinking mode.

The four agentic models use their native reasoning and tool-use prompts with
multi-turn image interactions, all at a temperature of 0.
The per-turn output limits are 10,240 tokens for DeepEyes and PixelReasoner,
20,480 for DeepEyesV2, and 2,048 for Thyme.
The total output budget is 32,768 tokens per question.
DeepEyes, DeepEyesV2, and PixelReasoner allow up to 11 turns, while Thyme allows 5;
the tool-call limit is 10 for all except PixelReasoner, which allows 4.
Generic answer-format instructions that conflict with the native agentic protocols are removed.

\subsection{\texorpdfstring{Evaluation with the Evidence Region}{Evaluation with the Evidence Region}}
\label{app:dual-view}

All ZoomGainBench results in Table~\ref{tab:zoomgain-results} provide the full image and the global question to models.
To study how much the evidence region helps, we evaluate four models under three inputs:
(1)~\emph{Full image}, which is the default setting,
(2)~\emph{Full image + crop}, which combines the full image with the zoomed view of the evidence region along with the same global question,
and (3)~\emph{Crop + local question}, which provides only the zoomed view and the local question.
As shown in Table~\ref{tab:dual-view},
adding the crop raises accuracy by 43.70--48.11 points,
and providing the crop and the local question further reaches 82.65--93.86.
It could be observed that the questions are answerable once the region is zoomed in,
and the default setting mainly evaluates a model's ability to 
autonomously locate and resolve fine-grained details in the full image.

\begin{table}[htbp]
    \centering
    \caption{\textbf{ZoomGainBench accuracy under different inputs.}}
    \label{tab:dual-view}
    \small
    \setlength{\tabcolsep}{8pt}
    \renewcommand{\arraystretch}{1.1}
    \begin{tabular}{lccc}
        \toprule
        \textbf{Model} & Full image & Full image + crop & Crop + local question \\
        \midrule
        Qwen3.5-4B            & 30.03 & 73.73 & 82.65 \\
        Qwen3.5-9B            & 32.13 & 80.24 & 90.78 \\
        Qwen3.5-35B-A3B       & 38.01 & 83.79 & 90.51 \\
        GLM-5.3-Flash~(320B)  & 45.54 & 89.86 & 93.86 \\
        \bottomrule
    \end{tabular}
\end{table}

\section{Training and Evaluation Settings}
\label{app:training-settings}

\paragraph{Training.}
Both datasets are trained with direct RL using GRPO,
with a batch size of 64, 8 rollouts per question, and one epoch (140 steps),
and the checkpoint at step 140 is evaluated.
For both ZwZ and ZoomGain, we use a learning rate of $10^{-6}$,
no KL regularization, and a maximum response length of 2,046 tokens.
ZwZ images retain their native bounding-box overlays, whereas ZoomGain uses the original images.
For ZwZ, we follow the official reward definitions~(\cite{wei2026zooming}).
ZwZ Align instead uses the original images without box-referring instructions,
converts counting and OCR questions to open-ended format, and uses the same rewards as ZoomGain.
For ZoomGain, rewards follow the ZoomGainBench scoring rules:
multiple-choice and counting questions receive a reward of 1 for an exact match
on the selected option or predicted integer, respectively, and 0 otherwise.
OCR questions receive their ANLS score as the reward if it is at least $0.7$,
and 0 otherwise.

\paragraph{Evaluation.}
For the training experiments, all models are evaluated with chain-of-thought prompting and a temperature of 0.
Within each benchmark, base and trained models share the same prompt profile and output limit:
2,048 new tokens for all benchmarks except BabyVision, PerceptionBench, and O3-Bench,
which use 8,192 tokens and are scored by GPT-5.5 as the answer judge.
MMVP is scored by paired accuracy, CV-Bench by its official balanced score,
and MME-RealWorld on 1,919 questions.
VisualProbe is reported as the question-weighted average of its easy (141), medium (268), and hard (106) splits.

\clearpage
\section{\texorpdfstring{Additional Training Results}{Additional Training Results}}
\label{app:additional-training}

\subsection{\texorpdfstring{Training on Different Model Family}{Training on Different Model Family}}
\label{app:gemma}

To test whether the benefit of ZoomGain data could generalize beyond the Qwen family,
we train Gemma-4-E4B-it~(\cite{gemmateam2026gemma4technicalreport}) on the 9K ZoomGain and ZwZ sets
with the same RL settings and evaluation protocol described in Table~\ref{tab:training-results}.
As shown in Table~\ref{tab:gemma-results}, 
ZoomGain consistently outperforms the base model across all 13 held-out benchmarks, 
achieving an average gain of 3.93 points (from 46.09 to 50.02), compared to 2.39 points for ZwZ.
In contrast, ZwZ only slightly surpasses ZoomGain on ZoomBench, which is the held-in benchmark of ZwZ.

\begin{table}[htbp]
    \centering
    \caption{\textbf{RL training on Gemma-4-E4B-it.}
    Avg.\ is computed over the 13 held-out benchmarks.}
    \label{tab:gemma-results}
    \setlength{\tabcolsep}{2.5pt}
    \renewcommand{\arraystretch}{1.12}
    \resizebox{\linewidth}{!}{%
    \begin{tabular}{l|ccccccc|ccc|ccc|c}
        \toprule
        \textbf{Model}
        & \shortstack{Zoom\\Bench} & \shortstack{V*\\Bench} & \shortstack{HR\\4K} & \shortstack{HR\\8K} & \shortstack{MME-RW\\Lite} & \shortstack{Tree\\Bench} & \shortstack{Visual\\Probe}
        & \shortstack{CV-\\Bench} & MMStar & MMVP
        & \shortstack{Baby\\Vision} & \shortstack{Perception\\Bench} & \shortstack{O3-\\Bench}
        & \textbf{Avg.} \\
        \midrule
        Gemma-4-E4B-it    & 35.15 & 56.54 & 67.88 & 61.75 & 44.87 & 47.65 & 15.15 & 78.92 & 63.73 & 56.00 & 14.18 & 29.57 & 27.83 & 46.09 \\
        \quad w/ ZwZ 9K     & \textbf{42.96} & 64.92 & 69.25 & 65.00 & 48.88 & 45.68 & 20.97 & 80.29 & 63.93 & 54.00 & 14.95 & 29.53 & \textbf{29.86} & 48.48 \\
        \quad w/ ZoomGain 9K & 42.49 & \textbf{67.54} & \textbf{71.00} & \textbf{66.88} & \textbf{49.14} & \textbf{47.90} & \textbf{22.33} & \textbf{81.49} & \textbf{64.40} & \textbf{61.33} & \textbf{15.98} & \textbf{29.87} & \textbf{29.86} & \textbf{50.02} \\
        \bottomrule
    \end{tabular}}
\end{table}

\subsection{\texorpdfstring{Distractors from Full-Image Errors Only}{Distractors from Full-Image Errors Only}}
\label{app:distractor-source}

Section~\ref{sec:error-analysis} reveals that global-view errors occupies the largest proportion of incorrect choices.
We therefore investigate whether these errors alone suffice as effective training distractors.
From the 9K training set, we filter all 2,277 multiple-choice questions including at least two global-view errors, 
and retain the 2,773 open-ended OCR and counting questions, resulting in 5,050 training samples per setting.
In \emph{Full-image errors only}, each multiple-choice question only has distractors taken from full-image errors.
In \emph{Multi-source}, each question keeps its original distractors.
Both conditions share the same training settings, and train Qwen3.5-4B for one epoch.
As shown in Table~\ref{tab:distractor-source},
\emph{Multi-source} distractors yield a higher average (63.19 vs.\ 61.12)
and outperforms \emph{Full-image errors only} on 9 of the 13 benchmarks,
with the largest gains on O3-Bench, TreeBench, and PerceptionBench,
showing its overall better benefit for training.

\begin{table}[htbp]
    \centering
    \caption{\textbf{Distractors from full-image errors only versus multiple sources.}
    Qwen3.5-4B trained on the same 5,050 samples, differing only in the distractors of multiple-choice questions.
    Avg.\ is computed over the 13 held-out benchmarks.
    \textbf{Bold} indicates the better result in each column.}
    \label{tab:distractor-source}
    \setlength{\tabcolsep}{2.5pt}
    \renewcommand{\arraystretch}{1.12}
    \resizebox{\linewidth}{!}{%
    \begin{tabular}{l|ccccccc|ccc|ccc|c}
        \toprule
        \textbf{Model}
        & \shortstack{Zoom\\Bench} & \shortstack{V*\\Bench} & \shortstack{HR\\4K} & \shortstack{HR\\8K} & \shortstack{MME-RW\\Lite} & \shortstack{Tree\\Bench} & \shortstack{Visual\\Probe}
        & \shortstack{CV-\\Bench} & MMStar & MMVP
        & \shortstack{Baby\\Vision} & \shortstack{Perception\\Bench} & \shortstack{O3-\\Bench}
        & \textbf{Avg.} \\
        \midrule
        Full-image errors only & \textbf{57.40} & \textbf{88.48} & 85.88 & 82.50 & \textbf{61.39} & 48.64 & 50.49 & 87.65 & 70.87 & \textbf{66.00} & 17.53 & 34.20 & 43.48 & 61.12 \\
        Multi-source           & 55.50 & \textbf{88.48} & \textbf{87.88} & \textbf{82.75} & 60.55 & \textbf{51.85} & \textbf{52.23} & \textbf{87.74} & \textbf{72.33} & \textbf{66.00} & \textbf{18.81} & \textbf{36.73} & \textbf{60.58} & \textbf{63.19} \\
        \bottomrule
    \end{tabular}}
\end{table}

\subsection{\texorpdfstring{Dependence on Test-Time Red Boxes}{Dependence on Test-Time Red Boxes}}
\label{app:red-box}

\begin{figure}[H]
    \centering
    \begin{minipage}{\linewidth}
    % Source IDs (English): ZoomGainBench zoominc_benchmark_v2_3_000825; ZoomBench bb73f09c27c371b96fe1e7aec80b465d.
    \textbf{(a) Recognizing the border of a sign}\\[3pt]
    What does the white border of the red LOCAL sign behind the A-frame sign resemble?\par
    {\small A. Stamp\quad B. Candy canes\quad C. Speech bubble\quad D. Stop sign\quad E. License plate}\par\smallskip
    \makebox[.326\linewidth]{\small Clean}\hfill
    \makebox[.326\linewidth]{\small w/ Box}\hfill
    \makebox[.326\linewidth]{\small Detail for reference}\par
    \includegraphics[page=1,width=\linewidth]{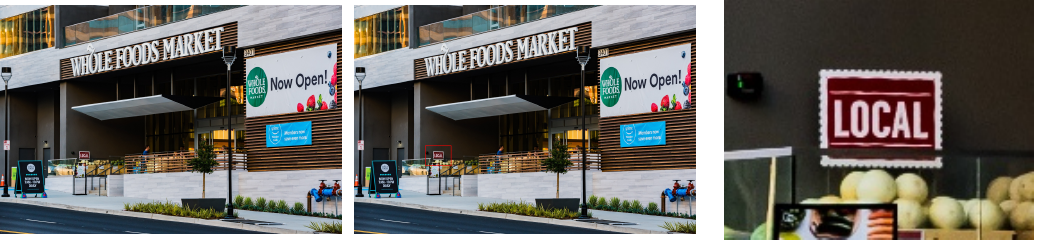}\par
    \makebox[.326\linewidth]{ZwZ: B. Candy canes}\hfill
    \makebox[.326\linewidth]{ZwZ: \textbf{A. Stamp}}\hfill
    \makebox[.326\linewidth]{Ground truth: \textbf{A. Stamp}}\par
    \vspace{1.5em}
    \textbf{(b) Counting swans on the water}\\[3pt]
    How many swans are visible in the water?\par\smallskip
    \makebox[.326\linewidth]{\small Clean}\hfill
    \makebox[.326\linewidth]{\small w/ Box}\hfill
    \makebox[.326\linewidth]{\small Detail for reference}\par
    \includegraphics[page=2,width=\linewidth]{figures/zwz_red_box_examples.pdf}\par
    \makebox[.326\linewidth]{ZwZ: 7}\hfill
    \makebox[.326\linewidth]{ZwZ: \textbf{8}}\hfill
    \makebox[.326\linewidth]{Ground truth: \textbf{8}}
    \end{minipage}
    \caption{\textbf{ZwZ answers a visual attribute question and a counting question correctly with a test-time red box.}
    Panel (a) is from ZoomGainBench and panel (b) from ZoomBench. In both, the ZwZ-trained Qwen3.5-9B answers the same question about the same full image in each condition. The boxed condition adds the ground-truth red box and an instruction to focus on its contents. The right column enlarges the evidence for the reader. In (b), the upper strip shows the four leftmost swans and the lower strip shows the other four.}
    \label{fig:zwz-red-box-examples}
\end{figure}

ZwZ overlays red bounding boxes on training images to specify targets.
To investigate how this affects inference, we compare the base Qwen3.5-9B with the ZwZ-trained 9B model
on ZoomBench and ZoomGainBench under two settings:
\emph{Clean} (standard unmodified images) and
\emph{w/ Box} (drawing the ground-truth red box around the target region on test images,
which aligns with the ZwZ training setting).
As shown in Table~\ref{tab:red-box},
adding ground-truth red boxes at inference time improves ZwZ more than the base model.
This confirms that ZwZ binds perception heavily to oracle red-box prompts,
which are unavailable on standard test setting.
Figure~\ref{fig:zwz-red-box-examples} shows a question from each benchmark that ZwZ answers correctly only in the boxed condition of this comparison.

\begin{table}[htbp]
    \centering
    \caption{\textbf{Effect of test-time red boxes.}
    $\Delta_{\mathrm{Box}}$ = w/ Box $-$ Clean.}
    \label{tab:red-box}
    \small
    \setlength{\tabcolsep}{6pt}
    \renewcommand{\arraystretch}{1.1}
    \begin{tabular}{lcccccc}
        \toprule
        & \multicolumn{3}{c}{\textbf{ZoomBench}} & \multicolumn{3}{c}{\textbf{ZoomGainBench}} \\
        \cmidrule(lr){2-4}\cmidrule(lr){5-7}
        \textbf{Model} & Clean & w/ Box & $\Delta_{\mathrm{Box}}$ & Clean & w/ Box & $\Delta_{\mathrm{Box}}$ \\
        \midrule
        Qwen3.5-9B          & 50.30 & 50.30 & +0.00 & 32.13 & 40.54 & +8.41 \\
        \quad w/ ZwZ 9K     & 58.93 & 61.30 & +2.37 & 44.07 & 56.69 & +12.62 \\
        \bottomrule
    \end{tabular}
\end{table}

\clearpage

\clearpage
\section{Global-to-Zoom Evidence and QA Construction Examples}
\label{app:dual-caption-examples}

\begin{figure}[H]
    \centering
    \includegraphics[width=\linewidth]{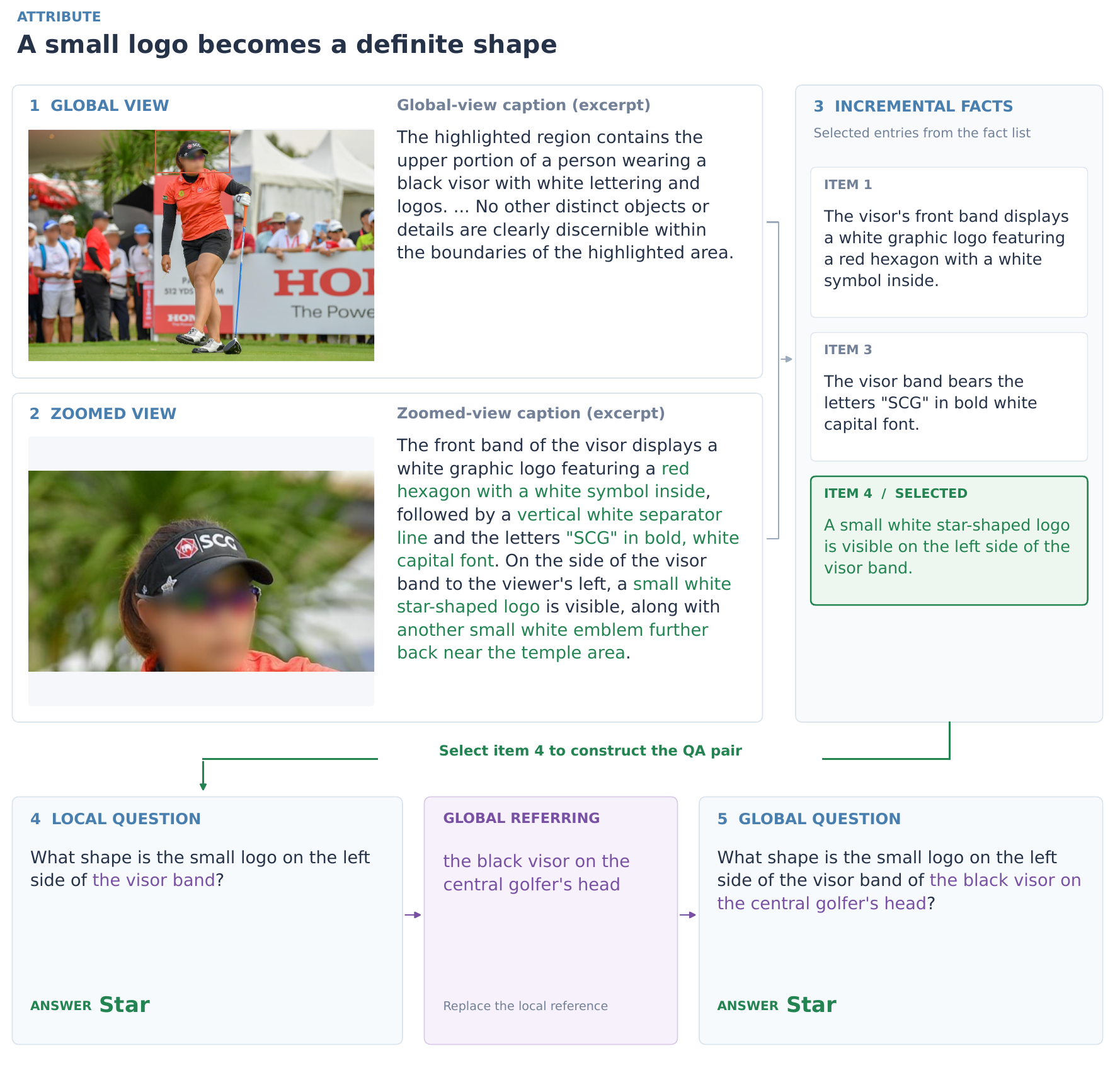}
    \caption{\textbf{Attribute QA construction example.}
    The global-view caption mentions logos on a visor without specifying their shapes.
    The zoomed-view caption identifies a small star-shaped logo.
    This fact supports a shape question, which is rewritten using the global referring expression
    while preserving the answer.}
    \label{fig:dual-caption-attribute}
\end{figure}

\begin{figure}[H]
    \centering
    \includegraphics[width=\linewidth]{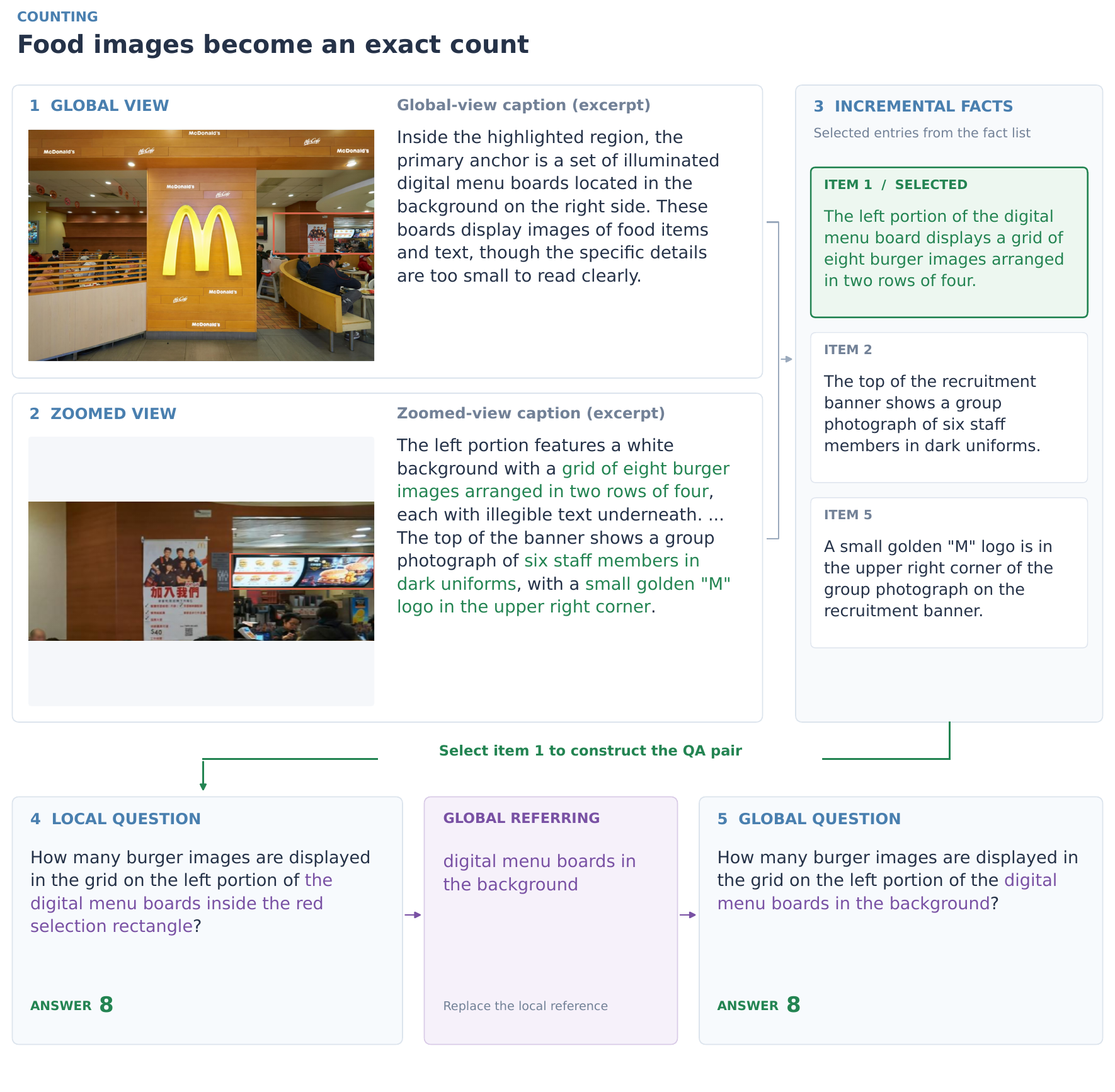}
    \caption{\textbf{Counting QA construction example.}
    The global-view caption describes food images on digital menu boards without giving a count.
    The zoomed-view caption identifies eight burger images arranged in two rows of four.
    The selected fact yields a counting question with the same answer in both views.}
    \label{fig:dual-caption-counting}
\end{figure}

\begin{figure}[H]
    \centering
    \includegraphics[width=\linewidth]{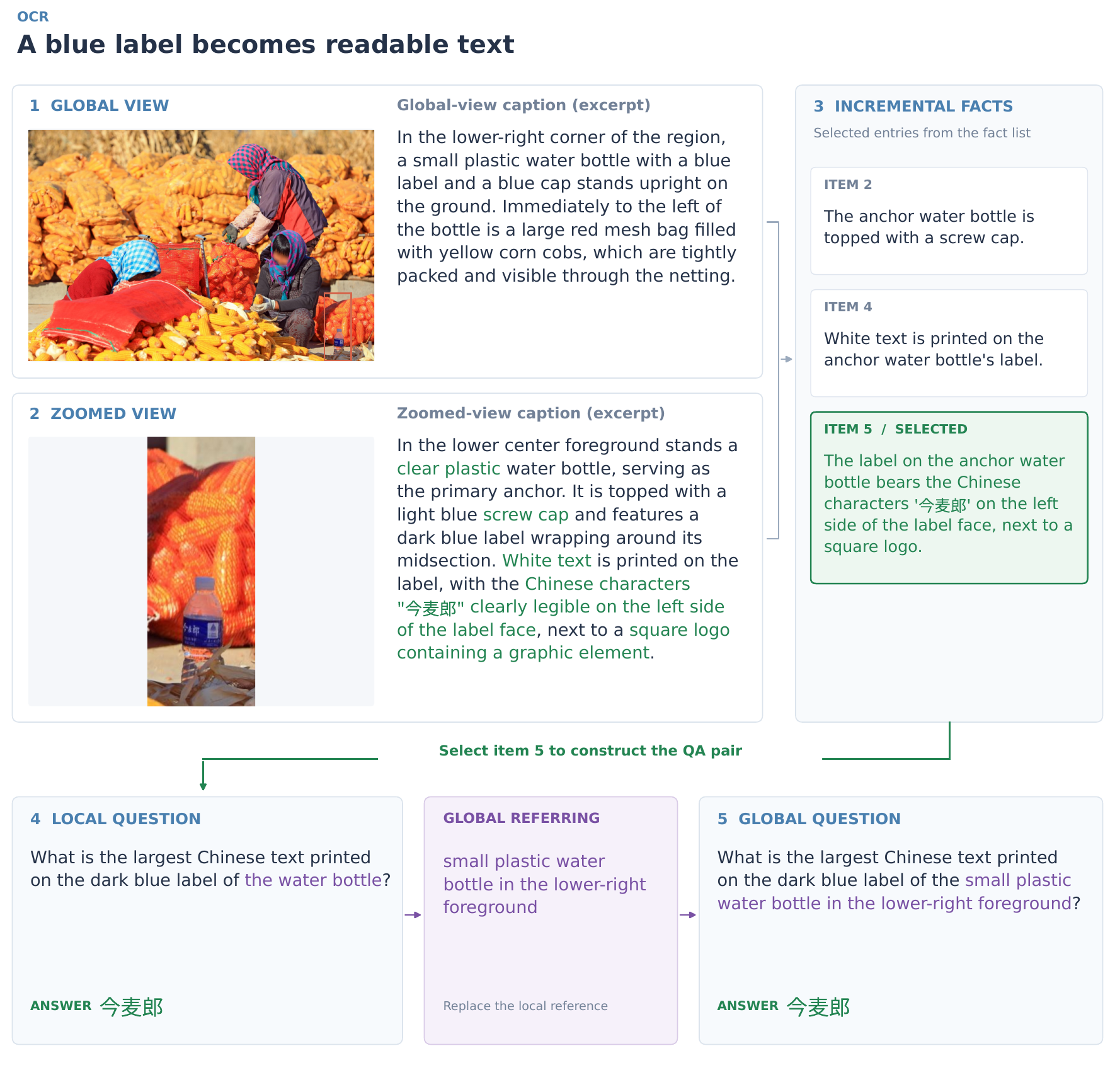}
    \caption{\textbf{OCR QA construction example.}
    The global-view caption describes a bottle with a blue label without transcribing its text.
    The zoomed-view caption reads the Chinese characters on the label.
    The selected text fact supports a local OCR question, which is rewritten to identify
    the bottle in the full image while retaining the original transcription as the answer.}
    \label{fig:dual-caption-ocr}
\end{figure}

Figures~\ref{fig:dual-caption-attribute}--\ref{fig:dual-caption-ocr}
illustrate the construction process in Section~\ref{sec:incremental-qa}
for the three question types.
Each example shows the paired views and caption excerpts,
selected entries from the original incremental fact list,
and how one fact leads to a local QA pair and then a global QA pair.
Green text marks details newly described in the zoomed-view caption;
purple text highlights the local and global target references.
The zoomed views use the original pipeline crops.

\clearpage

\clearpage

\clearpage
\section{Quality Verification Examples}
\label{app:verification-examples}

\begin{figure}[H]
    \centering
    \includegraphics[width=\linewidth]{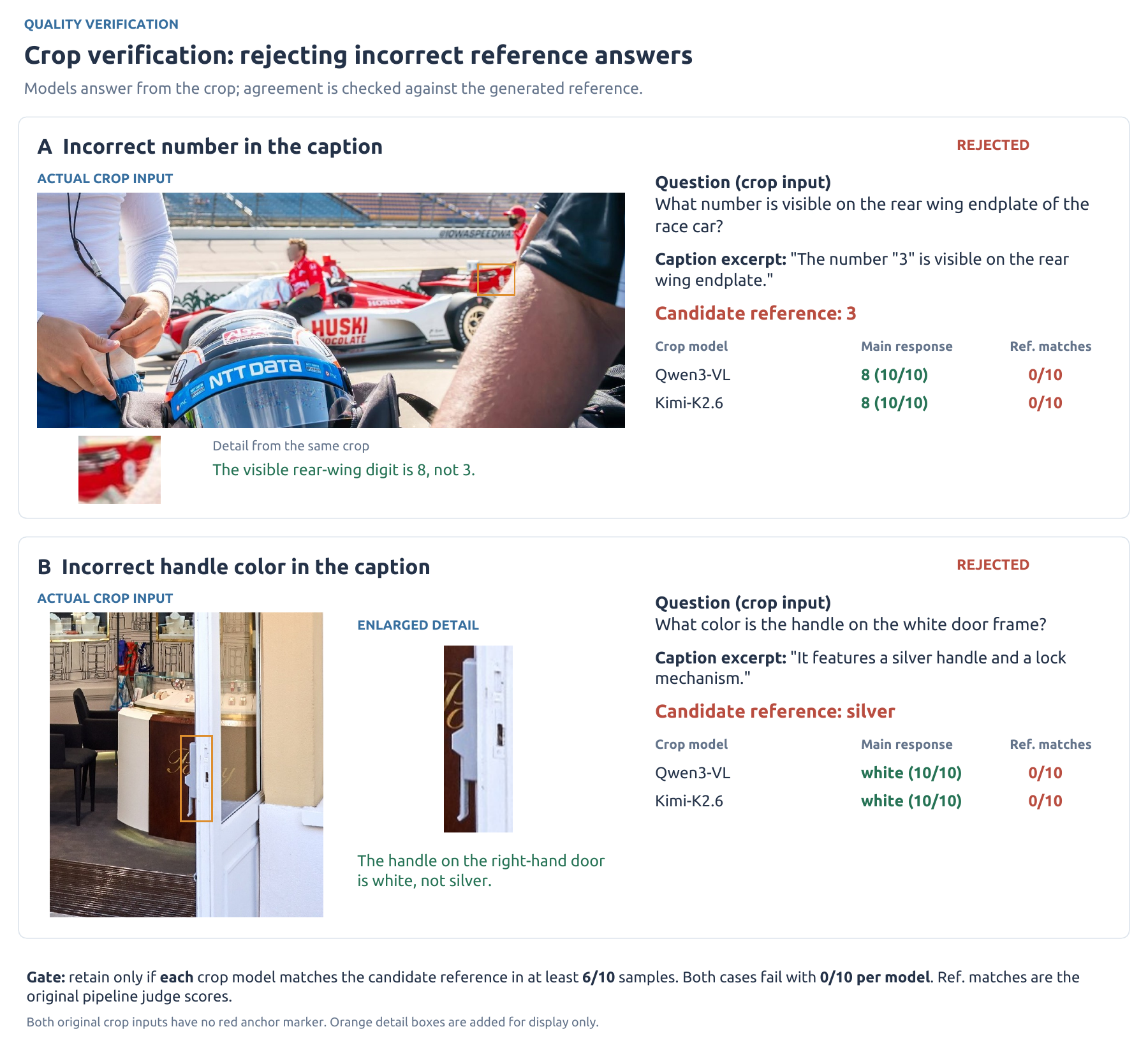}
    \caption{\textbf{Examples rejected by crop-based verification.}
    The captions incorrectly describe (A) the race-car digit as ``3'' instead of ``8''
    and (B) the door handle as silver instead of white.
    Both verification models give the visually supported answer in all 10 samples,
    yielding zero matches to the generated reference answer and failing the 6/10 threshold.
    The original crop inputs have no red anchor markers;
    orange boxes and enlarged details are added only for display.}
    \label{fig:crop-verification-examples}
\end{figure}

\begin{figure}[H]
    \centering
    \includegraphics[width=\linewidth]{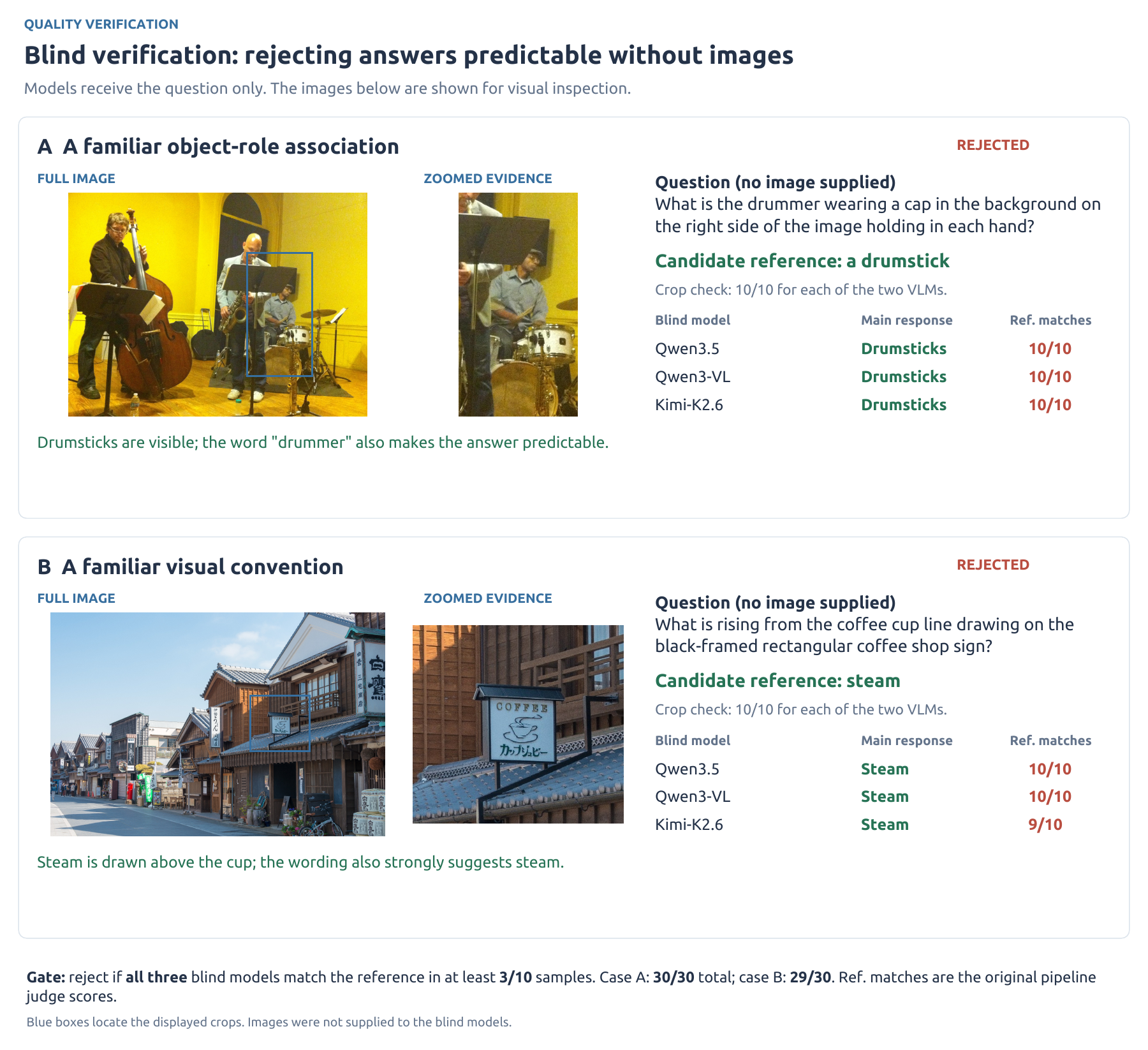}
    \caption{\textbf{Examples rejected by blind verification.}
    The reference answers, (A) drumsticks and (B) steam, are supported by the images,
    but are also predictable from the questions alone.
    Without images, the three models obtain 30/30 and 29/30 reference matches in total,
    respectively, with every model meeting the rejection threshold of at least 3/10 matches.
    The images and blue crop boxes are shown only for the reader;
    no images are supplied to the blind models.}
    \label{fig:blind-verification-examples}
\end{figure}

Figures~\ref{fig:crop-verification-examples} and~\ref{fig:blind-verification-examples}
show candidate QA pairs rejected by the two verification stages
in Section~\ref{sec:verification-distractors}.
Crop-based verification checks whether the image supports the generated reference answer,
while blind verification filters questions whose answers are predictable without an image.
All questions, candidate answers, and vote counts are taken from the original generation records.

\clearpage

\clearpage
\begingroup
\raggedbottom
\section{Prompts Used in Data Construction}
\label{app:construction-prompts}

This section provides the prompts for the roles described in Section~\ref{sec:pipeline},
plus anchor enumeration, marked/\allowbreak unmarked zoomed-view captioning, and
\texttt{EDGE\_REWRITER} from Appendix~\ref{app:construction-algorithm}.
Templates are taken from \texttt{zoom\_inc\_pipeline/prompts.py}; view-dependent instruction blocks
are taken from their callers in \texttt{pipeline.py}.
Sample-specific placeholders such as \texttt{\{anchor\}}, \texttt{\{question\}}, and \texttt{\{reference\}}
are filled at runtime. JSON braces are shown as literal braces rather than Python string escapes.
A line of the form \texttt{<<<INSERT: NAME>>>} denotes exact insertion of the named block below,
allowing shared instructions to be printed once without omitting their content.
The insertion notation is for presentation only and is not sent to the models.

\paragraph{View-specific inputs.}
\texttt{CAPTION1} receives the full image with the crop region marked.
\texttt{CAPTION2} receives an unmarked zoomed view, while \texttt{CAPTION2\_ANCHOR\_MARKED}
prepends marker-specific instructions to the same caption template.
The local anchor rule in QA generation is selected according to whether the zoomed view has an anchor marker.
\texttt{EDGE\_REWRITER} receives the crop and the full image with its crop region marked;
a candidate box on the crop can additionally locate the referent being audited.
Crop solvers answer the local question; full-image and blind solvers answer the global question,
with no image supplied for blind answering.
The Kimi format instruction is appended to its open-answer requests.
Visual distractor generation uses the appropriate crop/full-view rules.
MCQ verification sees the crop, local question, global question as context, and assembled choices.

\needspace{7\baselineskip}
\subsection{Shared instructions}
\needspace{5\baselineskip}
\paragraph{\texttt{SPATIAL\_FRAME\_RULES}.}
\begin{CJK*}{UTF8}{gbsn}
\begin{Verbatim}[fontsize=\small,breaklines=true,breaksymbolleft={},breaksymbolright={},xleftmargin=0.6em,xrightmargin=0.6em]
SPATIAL FRAME CONTRACT — apply to every direction word you read or write:
- LEFT/RIGHT BETWEEN OBJECTS uses the viewer's frame of the image: 'the person to the left of the sign' means left as seen by the viewer. Never mirror-compute the reference object's own left/right when locating one object relative to another. When such a phrase is load-bearing, prefer explicit wording such as 'on the left side of the image'.
- BODY PARTS are anatomical, never viewer-relative: 'right hand' always means the person's own right hand. Derive the side from the person's facing direction, never from raw image position: facing the camera -> the right hand appears on the viewer's LEFT; facing away from the camera -> the right hand appears on the viewer's RIGHT; facing image-left -> the left side of the body is toward the camera; facing image-right -> the right side of the body is toward the camera. If the facing direction is not clearly visible, never use left/right body-part wording; identify the part another way (by what it holds or touches, or with explicit viewer wording such as 'the hand on the left side of the image').
- FRONT/BEHIND is NEVER a depth word. Scene depth must be written only as 'closer to the camera (than X)' / 'farther from the camera (than X)' / 'beyond X from the camera's viewpoint'. The words front/behind are reserved exclusively for the intrinsic front/back sides of people, animals, and vehicles whose facing you have already stated: 'in front of the seated man' is the side his face and chest point toward, and 'behind him' is his back side, regardless of distance to the camera.
- DEPTH-FLIP TRAP (the most common mistake — check every sentence): when the camera sees a person's BACK, that person faces AWAY, so objects deeper in the scene are IN FRONT of that person, and objects between the camera and the person are behind them. Writing 'behind the man' for a sign that lies beyond a man seen from the back is WRONG — write 'in front of the man (farther from the camera)' or 'farther from the camera than the man'. The same flip applies to seated crowds watching something away from the camera.
- For objects with no inherent front (poles, boxes, trees, candles, signs), never use front/behind at all; use camera-distance or image-position wording.
- VEHICLE AND BUILDING PARTS use part terms fixed to the object itself: 'the cab end', 'the rear of the trailer', 'the side facing the camera', 'the entrance side'. Never call a vehicle end 'front' unless it is the actual forward/cab end of that vehicle.
- GROUPS and entities whose facing cannot be determined: use the viewer frame with explicit image wording ('in the upper-left part of the image', 'farther from the camera'), never intrinsic terms.
- WHOSE FRAME 'THE IMAGE' MEANS. Image-frame wording — 'in the upper-left part of the image', 'near the left edge of the image', '图像左侧' — always refers to the FULL photograph, and to nothing else.

\end{Verbatim}
\end{CJK*}

\needspace{5\baselineskip}
\paragraph{\texttt{CROP\_FRAME\_CAVEAT}.}
\begin{CJK*}{UTF8}{gbsn}
\begin{Verbatim}[fontsize=\small,breaklines=true,breaksymbolleft={},breaksymbolright={},xleftmargin=0.6em,xrightmargin=0.6em]
CROP FRAME CAVEAT:
- Your view is a window cut out of the middle of a larger photograph, so its borders are not the photograph's borders and its left side is not the photograph's left side. A position read off this view may be stated only as local to it, never as image-frame wording.
- A FRAME SIDE IS ONLY HONEST NEAR AN EDGE. 'on the left side of the image' is worth saying about something in the outer third of the photograph; for anything nearer the middle it reads as a claim the object is not where it is, and a solver scanning that side will not find it. When the target sits toward the centre, locate it against a named object instead — 'just right of the fire hydrant', 'in front of the bus shelter and a little to its left' — which stays true wherever the frame is drawn. You cannot see where this crop sits in the photograph, so when in doubt prefer the named object.
- FRAME SUPERLATIVES REQUIRE THE WHOLE PHOTOGRAPH. leftmost, rightmost, topmost, bottommost, far left, far right, 最左, 最右, 最上, 最下, 最顶, 最底 assert that nothing of the kind lies further out, which only a view of the entire photograph can establish. Without that view do not write one in either language. Bounded superlatives inside one named host ('the bottom word on the blue sign') are unaffected.
- A HOST'S OWN CORNER NEEDS THAT CORNER IN VIEW. 'the bottom right corner of the window', 'the top of the sign', '货架最下层' name a position inside another object, and they are only true if THAT object's edge on that side is visible here. When the host runs past the border, the corner you are looking at is the CROP's corner, not the host's, and the phrase points at something else entirely in the full photograph: a bookcase cropped to its top two shelves makes 'the book at the bottom of the bookcase' name a green spine on the second shelf, while the bookcase's real bottom shelf holds a red one. Use such a phrase only when you can see where the host ends on that side; otherwise identify the thing by what it is attached to, what is next to it, or its own visible attributes.

\end{Verbatim}
\end{CJK*}

\needspace{5\baselineskip}
\paragraph{\texttt{KIMI\_FINAL\_ANSWER\_TAG}.}
\begin{CJK*}{UTF8}{gbsn}
\begin{Verbatim}[fontsize=\small,breaklines=true,breaksymbolleft={},breaksymbolright={},xleftmargin=0.6em,xrightmargin=0.6em]
Begin your reply with the words The answer is, then put only your final answer inside <answer></answer>. For example: The answer is <answer>answer here</answer>.
\end{Verbatim}
\end{CJK*}

\needspace{7\baselineskip}
\subsection{Scene captioning and anchor grounding}
\needspace{5\baselineskip}
\paragraph{\texttt{SCOUT\_CAPTION}.}
\begin{CJK*}{UTF8}{gbsn}
\begin{Verbatim}[fontsize=\small,breaklines=true,breaksymbolleft={},breaksymbolright={},xleftmargin=0.6em,xrightmargin=0.6em]
Describe this image thoroughly. Then list every distinct, individually identifiable object or structure you can see (buildings, boats, vehicles, people, signs, statues, gates, bridges, etc.). Do not list background materials such as sky, water, grass, foliage or pavement as objects.
Direction wording: give every position in the image frame ('on the left side of the image', 'in the lower-right foreground'). Express depth only as foreground / background / closer to or farther from the camera. Never use 'in front of' or 'behind' a person, animal, or vehicle to mean nearer to or farther from the camera, and never name a person's own left/right body side.
\end{Verbatim}
\end{CJK*}

\needspace{5\baselineskip}
\paragraph{\texttt{PHRASE\_REWRITE}.}
\begin{CJK*}{UTF8}{gbsn}
\begin{Verbatim}[fontsize=\small,breaklines=true,breaksymbolleft={},breaksymbolright={},xleftmargin=0.6em,xrightmargin=0.6em]
You are preparing referring expressions for a visual grounding model.

Below is a description of an image. Extract up to {max_phrases} candidate landmark objects and rewrite each as a SELF-CONTAINED English referring phrase that would pick out exactly one object instance for someone looking at the image.

Rules:
- Each phrase must denote exactly ONE object instance. Add a position word (left / center / right, foreground / background) or a color/size attribute whenever the bare noun could match several instances.
- Position words inside a phrase must be IMAGE-FRAME words only: left / center / right, upper / lower, foreground / background, closest to the camera. NEVER locate a candidate with 'in front of', 'behind', 'ahead of', or 'beyond' another object or person — depth intent is ambiguous and grounding models disagree on it. Never use a person's own left/right body side, and never locate one object through another object that the description places at a clearly different depth.
- Only concrete, discrete objects: buildings, boats, vehicles, signs, towers, gates, statues, distinct groups. NEVER: sky, water, grass, trees, foliage, pavement, shadows, reflections, generic crowds.
- Phrases must rely only on properties visible in the description; do not invent.
- NEVER use written text, letters, logos or brand names read from the image as a qualifier (small text is unreliable at low resolution).
- Output strict JSON only:
{"phrases": [{"phrase": "...", "head_noun": "...", "qualifier_type": "position|color|size|none"}]}

Image description:
{caption}
\end{Verbatim}
\end{CJK*}

\needspace{5\baselineskip}
\paragraph{\texttt{GROUND\_NORM1000}.}
\begin{CJK*}{UTF8}{gbsn}
\begin{Verbatim}[fontsize=\small,breaklines=true,breaksymbolleft={},breaksymbolright={},xleftmargin=0.6em,xrightmargin=0.6em]
Locate the object referred to by: "{phrase}"
Interpret left/right/upper/lower/foreground/background in the phrase as the viewer's image frame, and foreground/background as distance from the camera.
Measure x and y on an integer coordinate grid from 0 to 1000 relative to the full image width and height. Return only one tight bounding box in this form: [x1, y1, x2, y2]
All four values must be integers between 0 and 1000. Include the complete visible extent of the referred object. If it is absent or ambiguous, return []. Do not output any other text.
\end{Verbatim}
\end{CJK*}

\needspace{5\baselineskip}
\paragraph{\texttt{GROUND\_NORM01}.}
\begin{CJK*}{UTF8}{gbsn}
\begin{Verbatim}[fontsize=\small,breaklines=true,breaksymbolleft={},breaksymbolright={},xleftmargin=0.6em,xrightmargin=0.6em]
Locate the object referred to by: "{phrase}"
Interpret left/right/upper/lower/foreground/background in the phrase as the viewer's image frame, and foreground/background as distance from the camera.
Measure x and y as decimal coordinates from 0 to 1 relative to the full image width and height, where 0 is the left/top edge and 1 is the right/bottom edge.  Return only one tight bounding box wrapped in tags, in this form:
<bbox>[x1, y1, x2, y2]</bbox>
All four values must be between 0 and 1. Include the complete visible extent of the referred object. If it is absent or ambiguous, return <bbox>[]</bbox>. Do not output any other text.
\end{Verbatim}
\end{CJK*}

\needspace{7\baselineskip}
\subsection{Anchor enumeration and paired-view captioning}
\needspace{5\baselineskip}
\paragraph{\texttt{ANCHOR\_INSTANCE\_CHECK}.}
\begin{CJK*}{UTF8}{gbsn}
\begin{Verbatim}[fontsize=\small,breaklines=true,breaksymbolleft={},breaksymbolright={},xleftmargin=0.6em,xrightmargin=0.6em]
Find EVERY distinct visible instance of the concrete object category "{noun}" in this image crop. Include partially visible instances.

For each instance, draw one tight bounding box around only its visible extent. Do not infer or include hidden parts. Use integer coordinates normalized from 0 to 1000 in [x1, y1, x2, y2] order.

Return exactly one JSON object with these fields:
- "count": a non-negative integer equal to the number of detected instances.
- "instances": a JSON array containing exactly "count" objects.
  Each object must contain:
  - "bbox_2d": four integers [x1, y1, x2, y2].
  - "description": a short, neutral, answer-free visual description.

The following consistency rule is mandatory:
count == len(instances)

If no instance is visible, return exactly:
{"count": 0, "instances": []}

Think carefully before answering. Output JSON only. Do not use Markdown fences or add explanations.
\end{Verbatim}
\end{CJK*}

\needspace{5\baselineskip}
\paragraph{\texttt{CAPTION1}.}
\begin{CJK*}{UTF8}{gbsn}
\begin{Verbatim}[fontsize=\small,breaklines=true,breaksymbolleft={},breaksymbolright={},xleftmargin=0.6em,xrightmargin=0.6em]
You are given a full image with one highlighted red rectangular region.
Describe **only the visual content inside the highlighted region and never mention the rest of the image.** Use the rest of the image only as context.

The pipeline has pre-selected one anchor object inside this region (localization metadata only):
"{anchor}"

ANCHOR-CENTERED ORGANIZATION (same convention as the zoomed view): treat this anchor as the primary reference point of the description, not as its exclusive scope. First identify the anchor and state every attribute of it that is discernible at this scale, then cover its immediate neighborhood, then the rest of the region. Whenever visually supported, express other objects' positions relative to this anchor. The anchor phrase is localization metadata only: never copy an attribute, text, count, or relation from the phrase unless you can independently see it in the image. If several objects match the phrase, describe each of them and say the reference is ambiguous; if you cannot confidently locate the referent, say so and describe the region normally. The anchor never shrinks your coverage duty: everything else plainly visible in the region must still be described.

YOUR CAPTION DEFINES WHAT THE FULL-IMAGE VIEW CAN SEE. A later stage compares this caption against a zoomed-in description of exactly the same region and treats every attribute you omitted as "not visible at full-image scale". The zoomed description follows the same spatial and inclusion rules as this one; the only intended difference between the two descriptions is resolution. Omitting something that is in fact plainly visible therefore corrupts the comparison. Be exhaustive about everything you can genuinely discern at this scale, and explicit about what you cannot.

EXHAUSTIVENESS CONTRACT (work from large to small):
- For every large or prominent object in the region — especially people — state ALL attributes that are plainly discernible at this scale: garment types and their colors (shirt, pants, dress, hat, shoes), object colors and materials, obvious pose or action, obviously identifiable held or attached items, and coarse spatial relations between the described objects. A large foreground person whose shirt color is obvious must have that shirt color stated.
- Then cover mid-sized objects the same way.
- For small objects and fine details (small text, small items, precise counts, subtle patterns, distant faces), report what is discernible and explicitly say "unclear" or "not readable at this scale" for the rest. Saying unclear is valuable signal; silently skipping is not.
- Do not compress multiple distinct objects into one vague summary sentence when their individual attributes are visible.
- When several instances of one category are present (price tags, signs, windows), bind each described instance to a stable qualifier — attachment or nearest-object relation, a part location on its host, or an explicit image-frame position — instead of the bare category noun.

<<<INSERT: SPATIAL_FRAME_RULES>>>
PERSON ORIENTATION: before naming a person's own left/right body side or a side-specific held item, state which way that person faces and derive the side per the contract above. When the facing is not discernible at this scale, use viewer wording ('the hand on the viewer's left') or say the side is unclear; never guess an anatomical side.

RECTANGLE-EDGE POLICY (aligned with the zoomed view): the rectangle boundary is an artificial viewport, not photographed scene content. For a discrete object that crosses the boundary, apply exactly one branch:
- If it cannot be confidently identified, or no complete part of it lies inside the region, omit it entirely.
- If it is clearly identifiable and at least one of its parts is completely visible inside the region, you MAY describe exactly those fully visible parts and their discernible attributes; say "unclear" for their small details. Never state whole-object attributes, whole-body actions, or totals for such an object, and never describe the part outside the region even though the full image shows it — it is outside this region's scope.
- Never mention the clipping itself anywhere in the caption: no "partially visible", "cut off", or "at the edge" wording. Describe what is inside, say unclear for what is too small, and stay silent about the rest.
- Large scene surfaces or structures may be described when a substantial, meaningful portion lies inside the region; do not describe their truncation.
Before returning the caption, perform a mandatory deletion check: remove every sentence that mentions the rectangle boundary or an object's clipped state.
If a small detail is not readable at this scale, explicitly say it is unclear. Do not guess counts, text, colors, or actions that are not visually clear.
\end{Verbatim}
\end{CJK*}

\needspace{5\baselineskip}
\paragraph{\texttt{CAPTION2}.}
\begin{CJK*}{UTF8}{gbsn}
\begin{Verbatim}[fontsize=\small,breaklines=true,breaksymbolleft={},breaksymbolright={},xleftmargin=0.6em,xrightmargin=0.6em]
You are given an image crop constructed around one selected anchor object.

Selected anchor object (localization metadata only):
"{anchor}"

Write a detailed natural-language caption organized around this selected anchor and its visible neighborhood. Treat the anchor as the primary reference point, not as the exclusive scope of the caption.

Coverage and organization:
- First describe the selected anchor itself in detail, including every clearly visible part, content, marking, readable text, color, attribute, condition, or action that belongs to it.
- Then describe every other clearly visible object and detail in the image. Do not omit reliable visual information merely because it is not part of the anchor.
- Whenever visually supported, express each other object's relationship to the anchor, preferring frame-free devices: attached to it, contained in it, on it, nearest to it, or beside it at the same depth. When a direction word is needed, it must follow the SPATIAL FRAME CONTRACT below: left/right between objects are the viewer's (write 'to the viewer's left of the anchor', never the anchor's own left/right), above/below are image-vertical, front/behind may only name the intrinsic sides of a person, animal, or vehicle whose facing you have stated, and any depth difference is written as closer to / farther from the camera.
- If no reliable relationship to the anchor can be established, still describe the visible object with a neutral location within this crop. Never invent a relationship merely to connect it to the anchor.
- DEPTH ZONES (mandatory): this crop is a 2D window that may span very different scene depths. For every non-anchor object you describe, make its depth relation to the anchor explicit: in the anchor's own depth zone, clearly closer to the camera than the anchor, or clearly farther from the camera. Never say an object is near, beside, or next to the anchor when it merely appears close in the 2D frame while lying at a clearly different depth (e.g. a foreground jogger inside a crop built around a distant lighthouse is 'much closer to the camera than the lighthouse', never 'near the lighthouse').
- The anchor phrase above is only localization metadata. Do not copy or assume any attribute, text, identity, or full-scene relationship from that phrase unless it is independently visible in this image.

<<<INSERT: SPATIAL_FRAME_RULES>>><<<INSERT: CROP_FRAME_CAVEAT>>>
PERSON ORIENTATION AND HANDS (mandatory):
- Before describing any person's pose, clothing sides, held objects, or body-part sides, state which way that person faces: toward the camera, away from the camera, toward the image-left, toward the image-right, or 'facing unclear' when it cannot be determined.
- POSE BEFORE SIDE: when two limbs of one person must be told apart, prefer a pose or state descriptor that needs no laterality at all — the raised arm versus the arm hanging at the side, the hand gripping the strap, the gloved hand, the arm resting on the table. Such wording is unambiguous for any viewer and stays true regardless of facing.
- Name an anatomical side only when a pose descriptor cannot separate the limbs AND the person's facing is unmistakable at this resolution (their face, or their back, is plainly readable — not inferred from a few pixels of hair or shoulder). Then derive the side from that stated facing per the frame contract and include the viewer-side derivation in parentheses so it is checkable, for example: 'He faces away from the camera; his right hand (on the viewer's right) holds a silver can.'
- If the facing is not unmistakable, never write 'left hand' or 'right hand': write the pose descriptor, or 'one hand (laterality unclear)' plus a viewer-frame location. A small or distant person is exactly the case where a confident side claim is unsafe.
- Whenever you describe what one hand holds, account for the other hand in the same pass: either describe what it visibly holds (or that it is visibly empty), or record it in the VISIBILITY AND REFERENCE NOTES section as out of view with unknown content.

INSTANCE LOCKING FOR REPEATED OBJECTS:
- When the crop shows, or plausibly extends into, more than one instance of a category — price tags, labels, signs, screens, windows, wheels, hands, shoes, bags, and similar repeated items — never attach information to the bare category noun. Bind every described instance to a stable qualifier that would remain valid in the full image: an attachment or nearest-object relation ('the mug nearest the kettle'), or a part location on its host ('on the tail of the aircraft').
- A position in this frame is NOT such a qualifier and never counts as one. This crop is a window cut out of a larger photograph and its borders are not the photograph's borders: whatever is leftmost, rightmost, or 'on the left side' HERE can sit dead centre, or on the opposite side, of the full image. Never write a frame superlative — leftmost, rightmost, topmost, bottommost, far left, far right — about anything in any wording: you cannot see what lies outside this crop, so you cannot know that anything in it is the outermost one.
- When a rough position genuinely helps a reader picture the scene, word it as local to this crop ('in the left part of this crop'), never as a fact about the image. An instance with no attachment or part qualifier is described without one and recorded in the VISIBILITY AND REFERENCE NOTES section, not pinned by position.

Include useful details such as objects, readable text, colors, attributes, actions, and spatial relationships. There is no required schema, bullet format, or fixed wording; organize the description naturally.

Base every statement only on visual evidence in this crop:
- HONEST EDGE REPORTING: the four outer crop edges are artificial viewport boundaries, not photographed scene content. When an object is cut off by an outer edge (the anchor included), simply say what you see: describe its clearly visible parts and their attributes, and never state whole-object attributes, whole-body actions, totals, or overall shape/size for it, and never guess what continues beyond the edge. Keep the main description positive — what IS visible — and put every cut-off / partially-visible / out-of-view remark in the VISIBILITY AND REFERENCE NOTES section at the end instead. A truncated object you cannot even identify belongs only in that section.
- For text, use a deliberately conservative, character-by-character standard. A character is clearly legible only when its complete glyph shape and boundaries are visible and there is no reasonable alternative reading; prefer a shorter certain transcription over a longer plausible one, and never use language, brand, or scene context to complete a word. If the entire inscription meets this standard, transcribe it exactly. Otherwise never guess the uncertain characters: when a contiguous substring of at least three characters is clearly legible, report only that substring, say the rest is unclear, and state where it sits (its character count and whether it is the beginning, end, or middle). Call it a beginning or ending only when that text boundary is itself visible in the crop, and say whether the character touching the uncertain side has a complete isolated glyph — under cropping, occlusion, blur, or small scale such boundary characters are unreliable. If the sign, panel, or surface carrying an inscription is itself cut off by the crop edge at or near the text region, treat the inscription as partial with the cut side uncertain even when every visible glyph is complete — the crop is not a god's-eye view and the text may continue on the hidden part. Never attribute text to an object unless it is visibly printed on that object. TEXT COLOR CAUTION: outlined or two-tone glyphs are easily misread at this scale — dark characters with a light outline routinely look like white text on a dark panel. State a text color only when the stroke FILL color is unambiguous; when uncertain, write 'light-colored text on a dark panel' or 'dark text on a light panel', or omit the color entirely. A guessed text color must never appear.
  FIGURE AND GROUND: a panel's color is never the glyphs' color. Name the color of the strokes and the color of the surface they sit on as two separate facts ('red characters on a yellow panel'), and never let one stand in for the other. Watch multi-tone signs especially: a fascia whose left panel is green and whose right panel is white has TWO panel colors and neither of them is a text color, so writing 'white letters' for black letters on its white panel is the same error as guessing. Illuminated night signage is where this goes wrong most.
- COUNTING-SCOPE POLICY: do not state an exact number merely because N instances happen to be visible in this crop. An exact count is allowed only in one of these two cases:
  (A) ANCHOR-OWNED SET: the counted items are parts of, attached to, worn by, held by, or contained in the selected anchor, and the complete anchor-side set relevant to the count is visible.
  (B) FULLY CLOSED ANCHOR-RELATIVE REGION: the counted items lie in a finite interval or region defined by the selected anchor and another specific visible object, and every boundary needed to delimit that region is fully visible in the crop. The crop edges must never serve as those boundaries.
  Phrases such as near, around, behind, in front of, to the left/right of, in the background, in the water, on the beach, or on the road merely locate visible instances; they do not close a counting set. Sea, water, beach, road, field, sky, crowd, and other surfaces or scene regions that continue beyond the crop are open sets, even when every instance currently visible in the crop is sharp and separable. For an open set, use non-numeric wording such as 'several are visible' when useful, and state that the exact scene total cannot be determined from the crop. Never estimate or give a range.
- Do not infer hidden parts or use world knowledge to fill gaps. Describe uncertain details as unclear.

End the caption with a final section exactly titled 'VISIBILITY AND REFERENCE NOTES:'. It is a short bullet inventory whose only purpose is to warn later stages which references would be ambiguous or open; it is not a description and can never be quoted as visual evidence:
- List each same-category counterpart whose content is unknown because it is cut off by a crop edge, out of view, or occluded: the other hand of a person when you described one hand, additional price tags/labels/signs beyond the ones you fully described, the unseen remainder of a vehicle or large object when you described one of its parts.
- Each note records only: the entity class and its owner or host ('the man's other hand', 'additional price tags', 'the rest of the blue trailer'), its status (cut off at which edge, out of view, or occluded), and 'content unknown' when applicable.
- Notes must NEVER contain attributes, colors, counts, actions, or transcribed text, and must never guess what the hidden content might be.
- If nothing qualifies, end with exactly 'VISIBILITY AND REFERENCE NOTES: none.'
\end{Verbatim}
\end{CJK*}

\needspace{5\baselineskip}
\paragraph{\texttt{CAPTION2\_ANCHOR\_MARKED}.}
\begin{CJK*}{UTF8}{gbsn}
\begin{Verbatim}[fontsize=\small,breaklines=true,breaksymbolleft={},breaksymbolright={},xleftmargin=0.6em,xrightmargin=0.6em]
A thin red selection rectangle has been added to this crop because the crop contains multiple objects of the same category. It marks the one specific instance named as the selected anchor object in the instructions below. Use the rectangle silently to distinguish that photographed object, its parts, contents, markings, and related nearby objects from those belonging to other same-category objects. Then write the caption as if the artificial rectangle were absent. Never mention, describe, count, transcribe, or treat the rectangle, its color, shape, position, corners, or boundary as photographed content. Refer to the selected photographed object naturally, for example by its object category and visible scene position.

<<<INSERT: CAPTION2>>>
\end{Verbatim}
\end{CJK*}

\needspace{7\baselineskip}
\subsection{Incremental evidence and bilingual QA generation}
\needspace{5\baselineskip}
\paragraph{\texttt{INCREMENT}.}
\begin{CJK*}{UTF8}{gbsn}
\begin{Verbatim}[fontsize=\small,breaklines=true,breaksymbolleft={},breaksymbolright={},xleftmargin=0.6em,xrightmargin=0.6em]
Two descriptions of the SAME image region are given. caption1 was written while seeing the full image (the region was highlighted with a box). caption2 was written while seeing a high-resolution crop of exactly that region. Both captions were given the same selected anchor phrase and were asked to organize around that anchor, so 'the anchor' in either caption denotes the same photographed object; use that shared reference to align object identities between them.

Selected anchor object:
"{anchor}"

The phrases 'selected object', 'selected anchor', 'marked object', or 'outlined object' in caption2 refer to this exact photographed anchor, not to a newly appearing object. Use this metadata only to align object identity between the two captions. Do not merge a different same-category object with the selected anchor. If an object's identity cannot be aligned reliably, omit that candidate fact. Any selection rectangle, outline, box, marker, label, or annotation mentioned by caption2 is pipeline metadata, not photographed content, and must never become an incremental fact.

Extract the incremental facts: discrete, visually verifiable information that caption2 provides and caption1 either omits (absent), marks as unclear (unclear), or states differently (contradicted). Contradictions ARE increments: report them with relation "contradicted" and record caption1's claim.

MANDATORY CROP-VIEWPORT INVARIANCE CHECK (higher priority than treating contradictions as increments): every retained fact must describe the photographed scene or an object's intrinsic property, and must remain semantically true if the crop were expanded. The crop is not a privileged or god's-eye view of the scene; it merely hides content beyond its four artificial edges. A difference in how much of an object falls inside the crop is NEVER an absent, unclear, or contradicted scene fact. Reject any candidate based on an object being at a crop edge, cut off, cropped, only partially visible, or reduced to particular visible body parts by the crop. For example, if caption1 describes a person walking while caption2 says only the person's head and shoulders are visible at the bottom edge, these claims do NOT contradict each other: caption2 is reporting viewport truncation, so omit the candidate entirely. Apply this rule even if caption2 states the boundary observation confidently and even if it can be assigned one of the allowed fact types. Silently audit every proposed fact for crop-edge dependence before adding it to the JSON.

VISIBILITY AND REFERENCE NOTES handling: caption2 may end with a section titled 'VISIBILITY AND REFERENCE NOTES'. That section is disambiguation context ONLY. Never extract a fact from it, never quote it in whole or in part as source_from_caption2, and never let it add, upgrade, or contradict main-description content. Use it solely to stay aware that other same-category instances or hidden parts exist with unknown content.

REFERENT QUALIFIER PRESERVATION: when a caption2 evidence sentence pins an instance with a qualifier — an anatomical side ('in his right hand'), a stated facing direction, an attachment or nearest-object relation ('the mug nearest the kettle'), a part location on a host ('on the tail of the aircraft'), or a depth-zone tag ('much closer to the camera than the anchor') — the fact's statement MUST keep that qualifier. Dropping the qualifier and attributing the information to the bare category ('the man holds a can', 'the price tag reads 140') is forbidden, because other instances may exist beyond the crop. Never flip an anatomical side into a viewer side or vice versa when restating evidence; parenthetical viewer-side derivations from caption2 may be omitted from the statement, but the anatomical side itself must be preserved exactly.

SPATIAL-FRAME CONFLICTS ARE NOT INCREMENTS: both captions follow the same frame contract (anatomical body sides derived from stated facing; viewer frame between objects; no depth 'behind'). If the two captions disagree ONLY about which body side of the same person carries an item or pose (left vs right hand/arm/leg/shoulder), and caption2 states the person's facing with a derivation while caption1 does not, treat caption2's side as a corrected relabeling of the same observation: do NOT emit the side flip itself as a contradicted fact, and never create a fact whose main content is which body side something is on. Silently use caption2's side inside any other fact about that item.

CAPTION1 OMISSION GUARD (applies to relation 'absent' only): caption1 is written at full-image scale but may still simply forget an obvious attribute. When caption1 clearly describes an object as large or prominent within the region — for example a person occupying much of it — and the candidate fact states a coarse attribute any full-image viewer would plainly see (dominant clothing color or garment type, overall object color, coarse pose, obvious ongoing action), treat the missing attribute as a caption1 omission, NOT an increment, and omit the fact. Keep 'absent' facts only for genuinely fine-grained information: small text, small parts or accessories, precise counts, subtle patterns, small held items, or details whose visibility plausibly requires zooming in. This guard never applies to facts whose relation is 'unclear' (caption1 explicitly could not see it) or 'contradicted' (caption1 made a different claim).

Only keep facts of these types: count, OCR text, color, action, local attribute, spatial relation. Discard:
- subjective wording and anything hedged with "possibly/maybe";
- texture-only remarks (foliage, grass, sky, water, lighting);
- facts about the region boundary itself (what is inside/outside the box, objects cut by the crop edge);
- range counts ("15-20") and estimates;
- hue-level color differences (red vs orange, cyan vs blue) — keep a color contradiction only if the two claims are clearly different color classes.

COUNT FACT ELIGIBILITY (mandatory and higher priority than contradiction):
- Retain an exact count only when it belongs to exactly one of two scope types: (A) anchor_owned — items that are parts of, attached to, worn by, held by, or contained in the selected anchor; or (B) closed_anchor_relative_region — items inside a finite region explicitly bounded by the selected anchor and another specific object, with all boundaries fully visible in caption2.
- The artificial crop boundary never closes a set. Relative words such as near, around, behind, in front of, left/right of, in the background, in water, or on a beach/road/field are locators, not complete boundaries. A visible subset of people or objects on a sea, beach, road, field, sky, crowd, or other region that may continue beyond the crop is not an exact scene count. Reject it even if caption2 gives a confident number and even if caption1 gives a different number.
- For every retained count fact, add count_scope_type (anchor_owned or closed_anchor_relative_region), counted_entity, scope_boundary_description, and scope_fully_visible=true. If you cannot truthfully fill all four fields from caption2, omit the count fact. For non-count facts, use null for these fields.

OCR evidence rules:
- TYPE DISCIPLINE: any fact whose statement reports readable characters, letters, digits, words, logo text, or an inscription MUST use type 'OCR text' with all audit fields filled. Never type a text-reading fact as attribute, local attribute, or color — doing so would smuggle it past every OCR boundary rule.
- Preserve caption2's exact certainty boundary. If caption2 says only a substring is legible, label the fact as partial OCR text and preserve the exact substring, its character count, and its stated position.
- A partial OCR fact is eligible only when it contains at least three contiguous clearly legible characters and caption2 identifies a usable position for that substring.
- Never fill in, repair, normalize, or guess unclear characters. Never turn a partial substring into a claim about the full inscription, word, brand, or name.
- If caption2 transcribes a sequence but also says the inscription extends into a cropped or obscured side, treat it as partial OCR even when the visible sequence itself is described as clear. Preserve which side is uncertain and whether the boundary-adjacent glyph is fully isolated.
- For every OCR fact, add these audit fields: ocr_scope (full or partial), visible_sequence (the exact transcribed characters), uncertain_side (left, right, top, bottom, both, internal, none, or unknown), and boundary_character_isolated (a JSON boolean true or false, never a quoted string). If any portion of the same inscription, sign, label, or printed text is described as cropped, obscured, incomplete, or unclear, ocr_scope MUST be partial. Set boundary_character_isolated to true only when caption2 explicitly confirms that the glyph touching the uncertain side is complete and visibly separated on both sides; otherwise use false.
- For every OCR fact, also express the evidence relative to NORMAL READING ORDER rather than only to image geometry. Set substring_position to full when the complete inscription is visible, prefix when the visible sequence is the beginning of a longer inscription, suffix when it is the ending, or unknown when caption2 does not support that decision. Set unsafe_boundary to beginning or end when the corresponding boundary character of visible_sequence touches an uncertain region; use both, internal, none, or unknown when appropriate. Do not trim visible_sequence yourself. Physical top/bottom/left/right and sequence beginning/end are different concepts.

Output strict JSON only:
{"incremental_facts": [{"type": "...", "statement": "...", "source_from_caption2": "...", "ocr_scope": "full|partial|null", "visible_sequence": "...|null", "uncertain_side": "left|right|top|bottom|both|internal|none|unknown|null", "substring_position": "full|prefix|suffix|unknown|null", "unsafe_boundary": "beginning|end|both|internal|none|unknown|null", "boundary_character_isolated": false, "count_scope_type": "anchor_owned|closed_anchor_relative_region|null", "counted_entity": "...|null", "scope_boundary_description": "...|null", "scope_fully_visible": true, "relation_to_caption1": "absent|unclear|contradicted", "caption1_claim": null}], "skip_reason": null}
If there is no reliable increment — in particular if every candidate fact concerns only vegetation, sky, water, lighting, or generic texture — return an empty list with a short skip_reason.

caption1:
{caption1}

caption2:
{caption2}
\end{Verbatim}
\end{CJK*}

\needspace{5\baselineskip}
\paragraph{\texttt{QA\_GENERATION}.}
\begin{CJK*}{UTF8}{gbsn}
\begin{Verbatim}[fontsize=\small,breaklines=true,breaksymbolleft={},breaksymbolright={},xleftmargin=0.6em,xrightmargin=0.6em]
You are generating fine-grained visual QA from incremental facts.

The anchor object (salient and uniquely locatable in the full image) is:
"{anchor}"

Crop-local anchor rule:
{local_anchor_rule}

Incremental facts (information visible only when zooming into the region around the anchor; each has an evidence sentence from the zoomed-in caption):
{facts}

Full zoomed-view description of the region (use it to check for duplicate objects of the same category — NOT as a source of new facts):
{caption2}

Current generation target plus feedback from prior QA candidates for this SAME anchor:
{retry_context}

Generate up to {qa_per_anchor} questions.

Every incremental fact has a source_fact_index. Return that exact integer for the one fact used by each QA. Copy evidence_sentence directly from that fact's source_from_caption2 when available; otherwise use its statement.

Question rules:
- Each answer must come from exactly one incremental fact.
- When generating more than one QA in the same response, every QA must use a different source_fact_index and evidence sentence; never make two questions from the same fact or paraphrase the same question.
- Any question that asks the solver to read or transcribe text, letters, characters, digits, words, logo lettering, or a specified character span MUST use question_type=ocr. Never label such a question as attribute, comparison, or another type.
- PRIORITY: if one or more eligible OCR facts are present, generate QA from those OCR facts before using count, attribute, action, or spatial facts. The incremental-facts list is ordered with OCR facts first.
- Read the ENTIRE OCR fact, especially statement and source_from_caption2; do not treat visible_sequence as the only useful information. Visual details explicitly tied to the text—such as its background color, text color, position, orientation, containing panel, or relation to another label—are valuable referring cues. Prefer a question that uses those cues to identify one text region over a generic question that asks for every word on a whole sign or logo. Use a visual attribute as a locator, not as another requested answer.
- A HEDGED IDENTITY IS NOT A REFERENT. When caption2 hedges what something is — 'what looks like a bird', 'possibly a helmet', '某种容器' — it is telling you it could not identify the object, and a referring expression built on that guess matches nothing when the guess is wrong: a solver hunting for a bird will never find the plastic bag it actually is. Drop the hedged noun and locate the thing by what IS certain — where it sits, what it is inside or attached to, its color and size ('the small white object on the ledge above the second window'). If nothing certain identifies it, skip the fact. Never carry the hedge into the question either: 'the shoe or boot' tells a solver you are guessing.
- LOCATOR ATTRIBUTE RELIABILITY: prefer host-level cues — panel/sign color, position, containing object, nearest landmark — over the text's OWN color. Small text color is frequently misperceived (dark outlined glyphs read as white), and a wrong locator attribute misleads solvers or matches a different text region. Include the text's color only when the fact states it without hedging AND it adds discrimination that the host cue does not already provide; when the panel alone is unique, leave the text color out. Never take a color from one region of a multi-tone sign and attach it to text sitting on another region; if you cannot say the stroke fill without hedging, use the host cue alone.
- ONE HOST PER OCR ANSWER. The answer is text carried by a SINGLE bounded host — one sign, label, plate, panel, or printed line. Never assemble an answer out of characters sitting on separate objects, and never scope an OCR question by nearness to a GROUP ('the digits printed near the green bottles', 'the letters around the potted plants'): proximity to a set has no boundary a solver can reproduce, so which characters belong in the answer becomes arbitrary and a second reader will include a different set. If the text you want spans separate hosts, ask about one of them instead.
- OCR facts may contain deterministic fields computed by code: safe_sequence, safe_character_count, required_span, and qa_eligible. Never correct, repair, or invent characters beyond safe_sequence. The answer must be one clearly supported CONTIGUOUS segment of safe_sequence.
- For a partial OCR fact (required_span=first or last), write a PLAIN text-reading question about that inscription, located by host-level cues, with the answer equal to safe_sequence verbatim. Do not add any span wording (no 'first/last k characters'), do not add totality words such as 'entire', 'complete', or 'all the text', and never mention cropping or hidden characters. A dedicated edge-rewriting stage that sees the full photograph will rewrite such drafts into the final safe span wording; your draft only needs to pick the right inscription and copy safe_sequence.
- LOCALIZED FULL-OCR SEGMENT: when required_span=full and the fact itself explicitly associates different complete words or consecutive word groups with independently identifiable visual regions, you may ask for one such region and use only its complete text as the answer. The chosen answer must be one or more WHOLE consecutive words appearing verbatim in safe_sequence; never take characters from inside a word. Both questions must include the fact-supported visual locator that isolates that segment. Do not invent a color, position, container, or relation. If the fact does not support a unique localized segment, use the complete safe_sequence instead.
- Example: if a fact says that OPEN appears in white on an upper green panel and 24 HOURS appears on a separate lower black strip, prefer asking what text appears on the upper green panel (answer: OPEN), rather than what all text on the sign says. This example describes the policy only; never copy its objects, colors, positions, or text into another image's QA.
- For required_span=full, either ask for the complete clearly legible text and use safe_sequence verbatim, or apply LOCALIZED FULL-OCR SEGMENT exactly as defined above.
- Never write view-dependent instructions such as 'excluding the top character', 'ignore the cropped character', 'remove the first visible letter', or any reference to an image/crop boundary, cropping, truncation, or hidden characters. A fixed first/last k span must have the same answer in the full image and zoomed image.
- Skip every OCR fact with qa_eligible=false or without all required deterministic fields.
- If the context contains prior-candidate entries in addition to the generation_target_state, choose a DIFFERENT incremental fact and answer property. Do not repeat or merely paraphrase any rejected or retained global/crop question, evidence sentence, or source fact.
- Use rejection feedback only to choose a better-supported candidate. NEVER alter the reference answer to imitate verifier outputs, and never invent a fact absent from the incremental-facts list.
- If a previous candidate failed crop verification, prefer a different fact whose answer is larger/clearer and directly visible in the crop.
- The generation_target_state entry in the context is authoritative. If zoom_required_still_needed=true and a previous accepted candidate was answerable from the full image, prefer a substantially smaller or finer detail marked absent/unclear/contradicted. If it is false, a retained full-image-answerable QA counts toward the accepted-question minimum; the next distinct accepted QA may be either full-image-answerable or zoom-required.
- If retry feedback says a previous candidate remained ambiguous AFTER the dedicated image-aware referent repair failed, use a different uniquely referable target. Never add answer-leaking qualifiers.
- DISAMBIGUATION: before writing each question, scan the full description above. If the answer object's category could match more than one object in the crop (e.g. a ring worn by a person vs a standalone ring product image), the crop-local referring phrase MUST include a qualifier that separates them (position, size, displayed-vs-worn, color). If no crop-visible qualifier can make it unique, skip that fact.
- Write TWO semantically paired versions of every question:
  1. question_zh / question_en are GLOBAL questions for the original full image. They MUST make the answer object locatable in the full image by a solver who has never seen the crop. Choose the locator by the target's REAL relation to the anchor (see GLOBAL LOCATOR POLICY below).
  2. crop_question_zh / crop_question_en are LOCAL questions for the exact zoomed crop. Remove context that the crop itself cannot establish, such as "in the center of the plaza", "on the building in the background", or "to the right of the white building". Keep only qualifiers whose evidence is visible inside the crop and that are needed to identify the answer object there.
- Follow the Crop-local anchor rule exactly. A selection rectangle may be used only as a referring-expression locator for the photographed anchor, its parts, contents, markings, or related nearby objects. Never ask about the rectangle itself, its color, shape, boundary, corners, or any artificial label. The GLOBAL questions must never mention a selection rectangle.
- The GLOBAL and LOCAL versions MUST ask the same property of the same object and MUST have the same answer. The local version is not a new question.
- The LOCAL question must be self-contained and its answer object must be uniquely identifiable inside the crop. If removing unavailable global context makes that impossible, skip the fact.

GLOBAL LOCATOR POLICY (the crop is a 2D window; objects inside it can lie at completely different scene depths, so 2D adjacency inside the crop is NOT scene proximity):
  (a) If the answer object has a REAL, visually supported scene relation to the anchor — it is a part of it, attached to it, on the same supporting surface, held by it, or genuinely adjacent at the same depth — locate it through the anchor phrase as usual.
  (b) If the answer object is clearly at a different depth than the anchor (the zoomed description says it is much closer to or farther from the camera, e.g. a foreground jogger inside a crop built around a background lighthouse), NEVER claim proximity with words such as near, beside, next to, 附近, 旁边, and never use front/behind as a disguised depth word. That fabricates a relation that does not exist, misleads the solver into searching around the anchor, and reads unnaturally. PREFER KEEPING THE ANCHOR in the question through an HONEST relation — it is the verified, easiest-to-find landmark — choosing whichever of these reads naturally:
      - CAMERA-AXIS relation (works for ANY anchor, including faceless structures; preferred when the target lies on or near the anchor's line of sight, which crop co-membership usually guarantees): state the depth order along the viewing axis with the canonical wording — target closer to the camera: 'the yellow taxi between the clock tower and the camera' / 在钟楼与镜头之间的黄色出租车; target farther: 'the brick warehouse beyond the red barge, farther from the camera' / 越过红色驳船更远处的砖砌仓库. When the two do not overlap in the image, add a small image-frame offset word ('below the clock tower and between it and the camera'). This wording is honest 3D geometry, never a proximity claim.
      - INTRINSIC ANCHOR-FRAME relation: when the anchor is a person, animal, or vehicle whose facing caption2 states, the target may be located in the anchor's own frame with each side derived mechanically per the contract. Example: a cyclist riding toward the viewer has the deeper scene BEHIND him and his left-hand side on the viewer's right, so 'the bright blue vertical sign behind the cyclist on his left-hand side' is honest and natural. SIDE-BLINDNESS GUARDRAIL: verification models cannot reliably resolve a person's left/right sides, so any anchor-frame directional clause is narrative scaffolding only — the question MUST still identify the target uniquely when that clause is ignored, through at least one anchor-independent discriminator (object type plus color, nearest fixed landmark, or viewer-frame position). If uniqueness would depend on the anchor-frame clause itself, do not use this option.
      - EXPLICIT IMAGE-FRAME relation to the anchor: honest 2D wording such as 'in the image, above the cyclist in the background' or 'below the lighthouse and much closer to the camera', never wording that implies physical proximity.
      - SELF-LOCATING referent (fallback when no anchor-linked wording reads naturally, e.g. the linkage would feel contrived across a large cluttered depth gap): drop the anchor and locate the target with full-image-frame position plus its own visible attributes, e.g. 'the jogger in the lower-center foreground of the image wearing a cap'. Depth words like foreground/closer to the camera stay true across views; horizontal words must be re-derived from the FULL image, never copied from the crop question.
  (c) If none of these produces a referent you are confident is unique in the FULL image, skip the fact.
  Prefer facts about the anchor itself or objects in its own depth zone: a crop-foreground object much closer to the camera is usually large and legible in the full image, so its facts rarely survive the full-image difficulty gate; spend them only when nothing anchored remains.

- Example pair:
  GLOBAL: "What text is written on the sign in the center of the plaza?"
  LOCAL: "What text is written on the sign?"
- NEVER put the answer attribute itself into either referring expression (if the answer is a color, the referring phrase must not mention that color).
- The answer must be a single definite value: an exact count, one color word, a short text string, or a short action. Skip facts with ranges, hedges, or subjective wording.
- The answer itself must never be an anatomical body side ('left hand', 'right leg') NOR a part location on an oriented object ('near the front left corner', 'at the rear end', 车头/车尾): all vision models share systematic mirror and orientation confusion, so such answers cannot be verified reliably. Sides and part locations serve only as referring qualifiers inside questions.
- COUNTING SOURCE GATE: before using a count fact, classify its real-scene scope. It is eligible only as (A) anchor_owned: the counted items are parts of, attached to, worn by, held by, or contained in the selected anchor, with the complete relevant anchor-side set visible; or (B) closed_anchor_relative_region: the items occupy a finite interval or region explicitly bounded by the anchor and another specific visible object, and all boundaries are fully visible. If neither classification is supported, skip the fact, even when it contains an exact number and is contradicted by caption1.
- Crop edges never close a counting scope. Phrases such as near, around, behind, in front of, left/right of, in the background, in water, or on a beach/road/field identify a visible subset but do not define a complete set. Never generate a question asking for the total of such an open scene region. In particular, counting the instances currently visible in a cropped slice of sea, beach, road, crowd, field, or sky is forbidden.
- For an eligible anchor_owned count, the question must explicitly bind the counted items to the anchor. For an eligible closed_anchor_relative_region count, the question must name both bounding objects and the finite interval/region between them. A generic phrase such as 'behind the anchor' is insufficient.
- Prefer count facts carrying count_scope_type and scope_fully_visible=true. Independently audit the evidence and full crop description; metadata cannot rescue an actually open scope.
- Skip facts about the region boundary and vegetation/sky texture.
- Skip color facts that differ only at hue level between the two views.

<<<INSERT: SPATIAL_FRAME_RULES>>><<<INSERT: CROP_FRAME_CAVEAT>>>
- Apply the frame contract above to every direction word in questions and answers. Body-part sides stay anatomical exactly as the fact states them ('right hand' means his own right hand); never convert an anatomical side into an image side or vice versa. Keep parenthetical facing/viewer-side derivations OUT of the question text — questions carry only natural human phrasing.
- BILINGUAL FRAME PARITY: the Chinese and English versions must encode the same frame with the same discipline. In Chinese, scene depth must be written as 背景中 / 前景中 / 离镜头更远 / 更靠近镜头 — NEVER as 后方、身后、前方 or 前面 attached to a person or animal (write 跑者背景中更远处的车辆, not 跑者后方的车辆). Use 身后/面前 only for a person's intrinsic sides and only when the question itself states that person's facing (e.g. 背对镜头的人身后). 左手/右手 are anatomical only, exactly like left/right hand in English.
- LIMB QUALIFIER PREFERENCE: to single out one of a person's paired limbs, prefer the pose or state the fact records — the raised arm, the arm hanging at the side, the hand holding the strap — over an anatomical side. Pose wording is natural, needs no facing derivation, and is the only kind a solver can check; verification models do not reliably resolve a person's left from their right. Use 'left/right hand' only when the fact states the side with an explicit facing derivation AND no pose descriptor separates the limbs.
- REFERENT CLOSURE (mandatory): if the fact's answer object belongs to a multi-instance category — hands, feet, shoes, price tags, labels, signs, screens, wheels, windows, bags, or similar repeated parts and products — and the zoomed description or its VISIBILITY AND REFERENCE NOTES indicate that other instances exist anywhere in the scene, are cut off, or are out of view with unknown content, then BOTH the global and crop questions MUST pin the exact instance using the qualifier carried by the fact and its evidence: the anatomical side ('in his right hand'), the attachment or nearest relation ('on the price tag closest to the kettle'), or the part location ('on the rear end of the large blue trailer'). A bare 'in his hand', 'on the price tag', 'on the sign', or 'on the vehicle' is forbidden in that situation. If the fact and evidence carry no such qualifier while the notes show counterparts exist, SKIP that fact.
- Never mention the crop viewport, cropping, truncation, cut-off or out-of-view states, visibility limitations, or the VISIBILITY AND REFERENCE NOTES section in any question; the notes only tell you when an extra pinning qualifier is required. Positional image-frame references such as 'near the left edge of the image' are locators, not viewport talk, and remain allowed under the frame contract.
- FRAME CLAIMS COME FROM CAPTION1 ONLY. caption2 saw a crop, not the photograph, so any left/right/side wording in it describes the crop's own frame and is worthless — often actively wrong — as a statement about the image: a crop taken from the right of the photo has its own left side. Never copy or paraphrase a horizontal position out of caption2 into a question. Use one only when caption1, which saw the whole photograph, places the target there, and word it as caption1 does.
- NO FRAME SUPERLATIVES. Never write leftmost, rightmost, topmost, bottommost, far left, far right, 最左, 最右, 最上, 最下, 最顶, 最底 about the image in either language. Being the outermost of a category is a claim about everything in the photograph, which neither caption can establish; use an attachment, nearest-object, or part qualifier instead. A superlative INSIDE one named host stays fine ('the bottom word on the blue sign'), because the host bounds it.
- After eligible OCR facts, prefer counting / attribute facts, and prefer facts whose relation_to_caption1 is "contradicted" or "unclear".
Output strict JSON only:
{"qas": [{"question_zh": "...", "question_en": "...", "crop_question_zh": "...", "crop_question_en": "...", "answer": "...", "question_type": "counting|ocr|attribute|action|spatial|comparison", "evidence_sentence": "verbatim evidence", "source_fact_index": 0}]}
If no fact qualifies, output {"qas": []}.
\end{Verbatim}
\end{CJK*}

\needspace{5\baselineskip}
\paragraph{\texttt{LOCAL\_ANCHOR\_RULE\_UNMARKED}.}
\begin{CJK*}{UTF8}{gbsn}
\begin{Verbatim}[fontsize=\small,breaklines=true,breaksymbolleft={},breaksymbolright={},xleftmargin=0.6em,xrightmargin=0.6em]
No artificial selection rectangle is present. Neither LOCAL question may mention a box, outline, marker, or annotation.
\end{Verbatim}
\end{CJK*}

\needspace{5\baselineskip}
\paragraph{\texttt{LOCAL\_ANCHOR\_RULE\_MARKED}.}
\begin{CJK*}{UTF8}{gbsn}
\begin{Verbatim}[fontsize=\small,breaklines=true,breaksymbolleft={},breaksymbolright={},xleftmargin=0.6em,xrightmargin=0.6em]
A thin red selection rectangle identifies the anchor in the crop. Both LOCAL questions MUST explicitly use the rectangle only as a locator, with wording such as 'the [object] inside the red selection rectangle' / '红框标出的[物体]'. If the fact concerns a part, content, marking, or nearby object, relate it explicitly to that red-boxed anchor. Never ask about the rectangle itself. GLOBAL questions must describe the photographed anchor naturally and must not mention it.
\end{Verbatim}
\end{CJK*}

\needspace{7\baselineskip}
\subsection{Joint crop/full-image audit and rewriting}
\needspace{5\baselineskip}
\paragraph{\texttt{EDGE\_REWRITER}.}
\begin{CJK*}{UTF8}{gbsn}
\begin{Verbatim}[fontsize=\small,breaklines=true,breaksymbolleft={},breaksymbolright={},xleftmargin=0.6em,xrightmargin=0.6em]
You are the crop-boundary auditor and question rewriter of a visual-QA data pipeline.

BACKGROUND — WHY THIS STAGE EXISTS:
The pipeline writes fine-grained QA about one anchor object. Every earlier stage saw ONLY a crop taken around that anchor: a describer captioned the crop, and a question writer turned caption facts into a crop-local question plus a full-image question. Neither could see past the crop's four edges, so two recurring failures escape them:
1. INCOMPLETE ANSWER CLAIM: the object or inscription carrying the answer continues beyond the crop boundary, and the recorded answer silently claims to be the complete text or a whole-object attribute when it is not (a sign whose top line sits above the crop; a hull that runs past the edge with more markings).
2. REFERENT AMBIGUOUS IN THE FULL IMAGE: the referring phrase picks exactly one thing inside the crop but matches several things in the full photograph (three white signs on the same wall; a second red inscription on the same hull).
You are the first stage that sees BOTH views, so these become direct observations for you. Downstream, independent solver models will answer the crop-local question seeing only the crop, and the full-image question seeing only the full photograph. Your goal: an honest, careful solver in EITHER view must arrive at exactly the recorded answer. Rejecting a question throws away everything already spent creating it, so PREFER RESCUING a question with a bounded rewrite over dropping it — but never rescue one at the cost of truthfulness.

INPUTS:
Image 1 is the crop. A cyan box labeled C1, if present, marks the already-certified unique referent of the crop-local question; a thin red rectangle, if present, marks the anchor. Both are annotations, never scene content.
Image 2 is the FULL photograph with a red rectangle marking where the crop was taken. Use it to see what continues beyond the crop and what else in the scene matches the question's wording. Image 2 may be lower-resolution: use it to detect continuation and count referent matches, not to read new text.

THE QA UNDER AUDIT:
anchor: "{anchor}"
question_type: {question_type}
global question (full-image solvers) EN: {question_en}
global question ZH: {question_zh}
crop-local question EN: {crop_question_en}
crop-local question ZH: {crop_question_zh}
recorded answer: {answer}
mechanical OCR contract (null for non-OCR): {ocr_constraint}
caption evidence sentence: {evidence}

AUDIT PROCEDURE:
A. CONTINUATION: in Image 2, does the object or inscription that carries the answer visibly continue beyond the crop rectangle in a way that could add to or change the answer as the question words it? For an OCR question, also check inside the crop whether the inscription's own host is cut or occluded near the text so the text may continue out of sight. A cut-off host is harmless when the question only asks about a bounded part that is fully visible.
  THE BURDEN IS ON 'false'. Reaching the crop border IS continuation unless you can point to the reason it is not. Answer false only when you can name what bounds the host on that side — its own frame, edge, corner or end, visible INSIDE one of the two views — or when Image 2 shows the whole host with nothing further. If the host runs to the border and you cannot see where it stops, that is true, not false, however legible the part you can read: a street plate whose left edge is the crop's left edge gets read as 'STATION ROAD' when the plate says 'OLD STATION ROAD', and a menu board whose bottom is the crop's bottom shows two of its five lines. Both look complete. Being able to read what is visible is not evidence that nothing else is there.
  Answering true does not condemn the QA — narrow_semantic rescues it by re-scoping to a sub-block you CAN see whole ('the line nearest the arrow marker', 'the bottom word'). Prefer that rescue over a confirm you cannot justify.
B. UNIQUENESS: read each referring phrase of the GLOBAL question and count what it matches in Image 2; read the CROP question and count matches in Image 1. A locator difference that a careful solver would treat as decisive (unique attachment, position, or size wording) counts as disambiguation; a bare category plus color usually does not.
   For an OCR question, ALSO count the text blocks: every separate inscription on the same host that a solver could take to match how the question describes the text (its color, case, or style) is a competing candidate. Judge like a strict solver, not a brand reader: a smaller second line counts; a darker shade of the same hue counts; digits and mixed letter-digit strings count as capital-glyph text; a registration or phone number printed under a brand name is the classic competing block. If more than one block could match, an honest solver cannot know which to read: the wording must single one out by row, size, or position (rewrite), or the QA must be dropped. Never assume the most prominent text is 'the' text.
   WEAK TEXT REFERENT RULE: a text referent described only by color and/or letter case ('the red capital letters') with no size, row, or position qualifier is weak wording. Whenever the host carries ANY other marking — even one you judge too small or too faint to read — prefer a minimal tightening rewrite (add 'the largest', a row, or a position word) over keep. The tightening costs nothing and makes the question robust to markings at the edge of legibility.
   Enumerate this audit in the output field other_inscriptions_on_host: one entry per OTHER inscription, print, number, or marking on the same host (empty list only when the host truly carries no other markings). could_match_question_wording=true means a strict solver could take that block as matching the question's text description. Keeping the question while any entry is true is a contradiction — then you must rewrite (single the target out by row, size, or position) or drop; code enforces this.
C. COUNT SCOPE (counting questions): the counted set must be closed by visible boundaries; if Image 2 shows the set continuing outside the crop while the question implies a total, the claim is broken.

VERDICTS:
- keep: both audits clean; the recorded answer is exactly what an honest solver would give in both views.
- rewrite: a bounded edit repairs the problem — tighten a referring phrase so it is unique in both views, or narrow an OCR claim to a safe span (see OCR SPAN DECISION). Rewrite all four question fields consistently.
- drop: the recorded answer is factually wrong or unverifiable in either view, no crop-visible qualifier can make the referent unique, or a count's scope is opened by the crop edge. Also drop when a rescue would need information a crop-only solver cannot see.
- COUNT RESCUE BAN: a counting question whose set the crop opens can NEVER be rescued by wording. Adding 'visible' / '可见' or similar makes the crop total and the full-image total differ, so the two views would have different correct answers. Such a question is unfixable: drop it.

REWRITE RULES:
- The recorded answer may NEVER be extended, corrected, or replaced with content you read in Image 2 — you sit downstream of evidence extraction and may only NARROW claims. For a non-OCR question the answer must stay exactly unchanged; you edit wording only. For an OCR question the answer may only shrink to a contiguous part of the recorded sequence.
- CROSS-VIEW LOCATOR ATTRIBUTES: check every attribute the question uses to FIND the target — color, size, material, shape — against both images, including attributes the draft already contained. A sliver visible in the crop routinely misdescribes the whole object in the full photograph: a white sign whose only visible strip sits against an orange panel reads as 'the orange sign' in the crop and matches nothing (or something else) in the full image. Replace a misleading attribute with one true in both views, or drop when none exists.
- Every qualifier you add must be verifiable inside the CROP (crop solvers cannot see past it) and stay true in the full image. Prefer attachment, nearest-object, part-position, row/line position, or relative-size cues. Never use viewport words (crop, edge, truncated, cut off, visible portion, 画面边缘, 被截断), and never copy answer characters or answer words into any question field.
- Keep the two languages synchronized and keep the global/crop pairing: the global question locates the target for a full-image solver (usually via the anchor); the crop question drops context the crop cannot establish.
- Spatial wording in rewrites must obey this contract:
<<<INSERT: SPATIAL_FRAME_RULES>>><<<INSERT: CROP_FRAME_CAVEAT>>>
OCR SPAN DECISION (REQUIRED for every ocr question, in ocr_span_decision; null for non-OCR):
You MAY change an OCR question's answer — this is usually the BEST rescue, not a last resort — provided the new answer is a contiguous part of the recorded one. Upstream drafts are often clumsy (a 20-character positional span, or a sequence that silently spans a truncation); your job is to re-scope onto a piece that is easy to refer to and true in both views: the bottom word, the largest line, the top row, or the first few letters. Work through the decisions in this order and pick the FIRST that yields a claim true in both views:
- confirm: the inscription the question asks about is completely visible with no plausible continuation, and the recorded answer is its complete text (or the question already carries consistent span wording). Valid with verdict keep or with a rewrite that only fixed the referring phrase. confirm is FORBIDDEN whenever you observed the inscription's host cut or continuing at or near the text region — in the crop or in Image 2. Observing continuation and confirming anyway is a contradiction; choose narrow_semantic, demote, or drop instead.
- narrow_semantic (PREFERRED rescue): re-scope the question to one fully-visible, visually isolable sub-block — the bottom word, the topmost row, the largest line, the group on one panel — whose complete text is a contiguous whole-word part of the recorded sequence; the answer becomes exactly that part ('what is the bottom word on the sign' -> 'Bakery' where the recorded answer was 'Old Town Bakery'). A solver reads such a unit naturally, with no character counting. Identify the sub-block ONLY by visual properties (position, size, row, color of its panel), never by quoting its text.
  THE SUB-BLOCK MUST BE CLOSED ON THE CUT SIDE. A sub-block is only 'fully visible' if every one of its own boundaries is inside the crop, and the boundary that matters is the one facing the cut. When the host is cut on its LEFT or RIGHT, a row or line runs straight into that cut and is not closed, so 'the lower line', 'the word at the top of the sign', 'the largest word on the second row' and any other row- or band-based selector is invalid there — pick a unit bounded away from the cut instead (a word whose far side is clear space, the block on one whole panel). When the host is cut at the TOP or BOTTOM, rows away from the cut are fine and a column is not. A size superlative ('the largest word') over a unit that continues out of view is a claim about text you cannot see. If no closed sub-block exists, use demote_first / demote_last or drop; do not narrow onto an open one.
- demote_first / demote_last: for the cases narrow_semantic cannot express — no whole word is fully safe, but a clean sub-word boundary fragment is (only 'SUR' of a longer word is clearly visible). Ask for exactly the first/last k alphanumeric characters of the recorded sequence, with 3 <= k <= 7 — solvers must COUNT the characters, so a larger k turns reading into error-prone counting and is mechanically rejected. Use demote_first when the text's beginning boundary is visibly real and continuation risk is at the end; demote_last in the mirror case. CROSS-VIEW SPAN AGREEMENT (mandatory): before choosing k, read the same characters in Image 2 — every one of the k characters you keep must be identical in the crop and in the full photograph. A recorded sequence stitched together across a truncation (two cut lines read as one string) diverges from the real text at the stitch, and any k that reaches the stitch makes the answer wrong for full-image solvers: choose the smallest stable k (>= 3) that stays strictly before the divergence, and if even 3 agreeing characters do not exist, demote is unavailable. Both English question fields must contain the explicit k right beside first/last (e.g. 'the first 3 characters'), and the Chinese fields must mirror it (e.g. “前3个字母数字字符”). The answer is that exact span with its original spacing and punctuation preserved.
- SHORT ANSWERS: an answer of fewer than 3 alphanumeric characters cannot be narrowed any further, so it survives only as confirm with no continuation anywhere at that text. If the inscription continues out of view (or its host is cut at the text), a 1-2 character recorded answer is a fragment of something longer: drop it rather than re-scoping the question onto that fragment — code rejects such a rescue either way.
- Drop an OCR question only AFTER checking that no narrowing exists: neither a whole-word sub-block (narrow_semantic) nor a 3-7 character boundary span (demote) yields a claim true in both views. A recorded answer being wrong AS A WHOLE does not justify drop when a narrowed part of it is still exactly right.

Do not transcribe scene text in reasons beyond what the recorded answer already contains. Return strict JSON only:
{"continuation_beyond_crop": true|false, "full_image_referent_matches": <integer count for the global question's referent in Image 2>, "crop_referent_matches": <integer count in Image 1>, "other_inscriptions_on_host": [{"description": "where/what kind of marking, WITHOUT transcribing it", "could_match_question_wording": true|false}] (for non-OCR questions use []), "verdict": "keep|rewrite|drop", "ocr_span_decision": "confirm|narrow_semantic|demote_first|demote_last"|null, "rewrite": null|{"question_zh": "...", "question_en": "...", "crop_question_zh": "...", "crop_question_en": "...", "answer": "..."}, "drop_reason": null|"concise reason", "reason": "concise image-specific audit reason"}
\end{Verbatim}
\end{CJK*}

\needspace{7\baselineskip}
\subsection{Crop, full-image, and blind answering}
\needspace{5\baselineskip}
\paragraph{\texttt{CROP\_VQA}.}
\begin{CJK*}{UTF8}{gbsn}
\begin{Verbatim}[fontsize=\small,breaklines=true,breaksymbolleft={},breaksymbolright={},xleftmargin=0.6em,xrightmargin=0.6em]
Look at this image and answer the question with a single short value (a number, a color, a text string, or a few words). No explanation.
Question: {question}
\end{Verbatim}
\end{CJK*}

\needspace{5\baselineskip}
\paragraph{\texttt{CROP\_VQA\_ANCHOR\_MARKED}.}
\begin{CJK*}{UTF8}{gbsn}
\begin{Verbatim}[fontsize=\small,breaklines=true,breaksymbolleft={},breaksymbolright={},xleftmargin=0.6em,xrightmargin=0.6em]
The thin red selection rectangle is an artificial locator identifying the photographed anchor selected by the pipeline. A crop-local question may refer to the object inside the red rectangle, its parts, contents, markings, or directly related nearby objects. Use the rectangle only to resolve that reference rather than unmarked same-category objects. Never treat the rectangle itself as photographed content or as the requested answer.
<<<INSERT: CROP_VQA>>>
\end{Verbatim}
\end{CJK*}

\needspace{5\baselineskip}
\paragraph{\texttt{CROP\_OCR\_VQA}.}
\begin{CJK*}{UTF8}{gbsn}
\begin{Verbatim}[fontsize=\small,breaklines=true,breaksymbolleft={},breaksymbolright={},xleftmargin=0.6em,xrightmargin=0.6em]
Look at this image and answer the OCR question by transcribing only the requested text or character span. Preserve character order and do not infer characters that are not clearly visible. If the requested text is not clear, answer exactly: unknown. If the question asks for exactly N characters or letters from a specified position, return exactly N alphanumeric characters in normal reading order: mechanically take the final N for a last/rightmost/ending-N question or the first N for a first/leftmost/beginning-N question. Spaces and punctuation do not count toward N. Do not return the complete visible word or any extra character. Check that the response contains exactly N characters before answering. Return the answer only, with no explanation.
Question: {question}
\end{Verbatim}
\end{CJK*}

\needspace{5\baselineskip}
\paragraph{\texttt{CROP\_OCR\_VQA\_ANCHOR\_MARKED}.}
\begin{CJK*}{UTF8}{gbsn}
\begin{Verbatim}[fontsize=\small,breaklines=true,breaksymbolleft={},breaksymbolright={},xleftmargin=0.6em,xrightmargin=0.6em]
The thin red selection rectangle is an artificial locator identifying the photographed anchor selected by the pipeline. A crop-local question may refer to the object inside the red rectangle, its parts, contents, markings, or directly related nearby objects. Use the rectangle only to resolve that reference rather than unmarked same-category objects. Never transcribe or return the rectangle itself as OCR content.
<<<INSERT: CROP_OCR_VQA>>>
\end{Verbatim}
\end{CJK*}

\needspace{5\baselineskip}
\paragraph{\texttt{FULL\_IMAGE\_VQA}.}
\begin{CJK*}{UTF8}{gbsn}
\begin{Verbatim}[fontsize=\small,breaklines=true,breaksymbolleft={},breaksymbolright={},xleftmargin=0.6em,xrightmargin=0.6em]
Look at this image and answer the question with a single short value (a number, a color, a text string, or a few words). If you cannot see the answer clearly, answer exactly: unknown. No explanation.
Question: {question}
\end{Verbatim}
\end{CJK*}

\needspace{5\baselineskip}
\paragraph{\texttt{FULL\_IMAGE\_OCR\_VQA}.}
\begin{CJK*}{UTF8}{gbsn}
\begin{Verbatim}[fontsize=\small,breaklines=true,breaksymbolleft={},breaksymbolright={},xleftmargin=0.6em,xrightmargin=0.6em]
Look at this image and answer the OCR question by transcribing only the requested text or character span. Preserve character order and return your single best transcription from the visible image, even when some characters are difficult to read. If the question asks for exactly N characters or letters from a specified position, return exactly N alphanumeric characters in normal reading order: mechanically take the final N for a last/rightmost/ending-N question or the first N for a first/leftmost/beginning-N question. Spaces and punctuation do not count toward N. Do not return the complete visible word or any extra character. Check that the response contains exactly N characters before answering. Return the answer only, with no explanation.
Question: {question}
\end{Verbatim}
\end{CJK*}

\needspace{5\baselineskip}
\paragraph{\texttt{BLIND\_ANSWER}.}
\begin{CJK*}{UTF8}{gbsn}
\begin{Verbatim}[fontsize=\small,breaklines=true,breaksymbolleft={},breaksymbolright={},xleftmargin=0.6em,xrightmargin=0.6em]
You cannot see the photograph. Answer the question using only its wording and your learned language and world knowledge. Even if the image-dependent answer is uncertain, you must make your single most likely guess. Return one concrete, short answer of the requested type. Never answer unknown, cannot determine, not enough information, or any other abstention. Do not provide alternatives or an explanation.
Question: {question}
\end{Verbatim}
\end{CJK*}

\needspace{5\baselineskip}
\paragraph{\texttt{ANSWER\_EQUIVALENCE}.}
\begin{CJK*}{UTF8}{gbsn}
\begin{Verbatim}[fontsize=\small,breaklines=true,breaksymbolleft={},breaksymbolright={},xleftmargin=0.6em,xrightmargin=0.6em]
You are a careful, strict, and deterministic evaluator of short visual-question answers.

Inputs:
- Question: {question}
- Reference answer: {reference}
- Candidate answer: {candidate}

Decide whether the candidate correctly fills the exact answer slot requested by the question.

Internally follow this order:
1. Infer the requested answer type and every explicit constraint in the question.
2. Identify the single answer asserted by the candidate. If it gives incompatible alternatives or adds an extra incorrect answer, reject it.
3. Compare that asserted answer with the reference in meaning.

Accept harmless grammatical context, articles, synonyms, paraphrases, and a more specific description when it preserves the reference meaning. Do not require the same wording. Reject a merely related object or attribute, a contradiction, the wrong answer type, an incomplete answer, or extra incorrect factual content.
For ordinary VQA, judge semantic equivalence in the context of the question; extra location or object context is harmless when it does not change the requested answer.

Think through the comparison internally. Output exactly one tag and nothing else:
<verdict>yes</verdict>
or
<verdict>no</verdict>
\end{Verbatim}
\end{CJK*}

\needspace{5\baselineskip}
\paragraph{\texttt{OCR\_ANSWER\_EQUIVALENCE}.}
\begin{CJK*}{UTF8}{gbsn}
\begin{Verbatim}[fontsize=\small,breaklines=true,breaksymbolleft={},breaksymbolright={},xleftmargin=0.6em,xrightmargin=0.6em]
You are judging whether a candidate OCR response correctly answers the exact question. Judge visual content, not answer-only formatting.
Question: {question}
Reference answer: {reference}
Candidate response: {candidate}
Apply every positional and length constraint in the question. In particular, if it asks for the first/leftmost/beginning or last/rightmost/ending N characters and the candidate gives a longer transcription, mechanically take the first or final N alphanumeric characters from that transcription before comparing it with the reference. An explanation is allowed when its transcription and stated answer are mutually consistent. Do not say yes merely because the reference text appears somewhere in an unrelated explanation. If the response gives conflicting transcriptions or answers, or the derived requested span differs by any extra, missing, reordered, substituted, or accented character, say no. Ignore only letter case and superficial spacing or punctuation. Reply with exactly one word: yes or no.
\end{Verbatim}
\end{CJK*}

\needspace{7\baselineskip}
\subsection{Multi-source distractors and MCQ verification}
\needspace{5\baselineskip}
\paragraph{\texttt{FULL\_DISTRACTOR\_CLUSTERING}.}
\begin{CJK*}{UTF8}{gbsn}
\begin{Verbatim}[fontsize=\small,breaklines=true,breaksymbolleft={},breaksymbolright={},xleftmargin=0.6em,xrightmargin=0.6em]
You are a conservative semantic auditor constructing multiple-choice distractor candidates from repeated full-image VQA responses. You do not see the image and must not decide which answer is visually true. Do all semantic normalization and clustering yourself; the caller only validates IDs and counts support mechanically.

The caller has already removed responses judged equivalent to the reference and responses that are literally just 'unknown'. Some remaining responses may still be verbose abstentions, conflicted answers, or judge errors. Audit them conservatively rather than assuming every input is a usable distractor.

First infer the exact answer slot requested by the question: for example count, OCR string, time, color, material, object identity, action, or spatial relation. Then audit every response independently.

Response rules:
1. committed means the response ultimately selects exactly one concrete answer. Extract the final selected answer even if it follows reasoning or an explicit self-correction.
2. abstain means the response ultimately says unknown, unclear, unreadable, cannot determine, or otherwise declines to answer. Classify it as abstain even if it mentions speculative possibilities.
3. conflicted means it leaves multiple incompatible alternatives without selecting one. Do not arbitrarily retain one alternative.
4. For abstain or conflicted, extracted_answer and cluster_id must be null.

Clustering rules:
5. Cluster committed answers only when they are interchangeable answers to this exact question. Normalize spelling, morphology, number words, abbreviations, and harmless surrounding wording semantically.
6. Do not merge merely related answers. Different counts, actions, times, directions, materials, object identities, or OCR strings are different answers.
7. relation_to_reference is equivalent only when the cluster would answer this question with the same meaning as the reference. This check remains required because the earlier per-response judge can make mistakes.
8. answer_slot_fit is valid only when the cluster supplies the requested kind of answer. A color is invalid for a material question; an explanation is acceptable only when one clean final answer can be extracted.
9. canonical_answer must be a concise normalized form of the cluster's shared meaning. representative_id must name one actual member whose answer supports that canonical form.

Completeness contract:
- Process every supplied response ID exactly once.
- Every committed response belongs to exactly one cluster.
- Cluster member_ids must exactly partition all committed responses.
- Never invent, omit, or duplicate an ID.
- Output one strict JSON object and no Markdown.

Question type: {question_type}
Question: {question}
Reference answer: {reference}

Responses:
{responses}

Output schema:
{"answer_slot": "short description of what kind of answer is requested", "responses": [{"id": "F001", "disposition": "committed|abstain|conflicted", "extracted_answer": "answer or null", "cluster_id": "C1 or null"}], "clusters": [{"cluster_id": "C1", "canonical_answer": "concise answer", "member_ids": ["F001"], "representative_id": "F001", "relation_to_reference": "equivalent|different", "answer_slot_fit": "valid|invalid", "reason": "short explanation"}]}
\end{Verbatim}
\end{CJK*}

\needspace{5\baselineskip}
\paragraph{\texttt{BLIND\_DISTRACTOR\_CLUSTERING}.}
\begin{CJK*}{UTF8}{gbsn}
\begin{Verbatim}[fontsize=\small,breaklines=true,breaksymbolleft={},breaksymbolright={},xleftmargin=0.6em,xrightmargin=0.6em]
You are a conservative semantic auditor constructing multiple-choice distractor candidates from blind, text-only model responses. You do not see the image. Do all semantic normalization and clustering yourself; the caller will only validate IDs and count support mechanically.

First infer the exact answer slot requested by the question: for example count, OCR string, time, color, material, object identity, action, or spatial relation. Then audit every response independently.

Response rules:
1. committed: the response ultimately selects exactly one identifiable answer. Extract that final answer even if it is surrounded by reasoning.
2. abstain: it says unknown, cannot determine, or refuses to guess.
3. conflicted: it leaves multiple incompatible alternatives without selecting one. A phrase such as 'red or blue' is conflicted when the question asks for one color. Do not arbitrarily keep one option.
4. For abstain or conflicted, extracted_answer and cluster_id must be null.

Clustering rules:
5. Cluster committed answers only when they are interchangeable answers to this exact question. Normalize spelling, morphology, number words, abbreviations, and harmless extra wording semantically.
6. Do not merge merely related answers. Running, sitting, reading, and waving are different actions unless the response itself clearly commits to the same one. ALPHA, BETA, and GAMMA are different OCR strings. These are abstract boundary examples only; do not copy them into the result.
7. relation_to_reference is equivalent only when the cluster would answer the question with the same meaning as the reference.
8. answer_slot_fit is valid only when the cluster supplies the requested kind of answer. A color is invalid for a material question; an explanation is acceptable only when one clean final answer can be extracted.
9. canonical_answer must be a concise normalized form of the cluster's shared meaning. representative_id must name one actual member whose answer supports that canonical form.

Completeness contract:
- Process every response ID exactly once.
- Every committed response belongs to exactly one cluster.
- Cluster member_ids must exactly partition all committed responses.
- Never invent, omit, or duplicate an ID.
- Output one strict JSON object and no Markdown.

Question type: {question_type}
Question: {question}
Reference answer: {reference}

Responses:
{responses}

Output schema:
{"answer_slot": "short description of what kind of answer is requested", "responses": [{"id": "B001", "disposition": "committed|abstain|conflicted", "extracted_answer": "answer or null", "cluster_id": "C1 or null"}], "clusters": [{"cluster_id": "C1", "canonical_answer": "concise answer", "member_ids": ["B001"], "representative_id": "B001", "relation_to_reference": "equivalent|different", "answer_slot_fit": "valid|invalid", "reason": "short explanation"}]}
\end{Verbatim}
\end{CJK*}

\needspace{5\baselineskip}
\paragraph{\texttt{VLM\_DISTRACTOR\_GENERATOR}.}
\begin{CJK*}{UTF8}{gbsn}
\begin{Verbatim}[fontsize=\small,breaklines=true,breaksymbolleft={},breaksymbolright={},xleftmargin=0.6em,xrightmargin=0.6em]
You are a conservative visual multiple-choice distractor generator. You see exactly one {view_name} image. Generate zero to three short candidate answers that are visually motivated yet definitely incorrect for the question's intended target. The reference answer is immutable and correct. Returning no candidate is better than inventing unsupported content.

Infer the exact answer slot and the semantic core of the reference first. Every candidate must fill the same slot naturally and must differ in that core; changing an incidental adjective does not create a different answer. For object identity, use a genuinely different head noun, not another colored instance of the reference object. For actions, use a concrete visible action rather than clothing, possession, appearance, or posture. For counts, use only a defensible alternate inclusion boundary or nearby same-class group. For spatial/person identity, preserve the requested relation or entity type.

Preferred visual strategies, strongest first:
1. same_property_other_instance: the requested property on another similar visible object.
2. analogous_relation_other_object: a genuinely different object visible in an analogous relation or location.
3. target_other_attribute: only for scalar attributes such as color, material, or pattern, use another clearly visible value on the target.
4. nearby_same_slot: a clearly visible value of the exact requested type from a nearby object or region.
5. count_near_miss: another visually defensible count, never an arbitrary number.

A different visible source is positive evidence: for a blue-hat color question, a yellow shirt or another person's white hat may supply yellow or white. Name the exact source and relation. Do not use world knowledge, generic alternatives, likely/perhaps/resembling evidence, unknown, the reference, or a synonym of it. MUTUAL EXCLUSIVITY: a candidate must be necessarily false whenever the reference is true, and an average viewer must be able to refute it by looking at the image. Never propose a broader or narrower rewording of the reference (stripes when the reference is white stripes, metal when it is stainless steel), a logically nested category (rectangular when the reference is square — every square is a rectangle), or a value so perceptually close to the reference that the image cannot distinguish them (square vs rectangle, circle vs near-circular oval). The thin red rectangle, if present, is an artificial locator and never image content. Rank strongest first, do not pad to three, and output strict JSON only.

View discipline:
{view_rules}

Question type: {question_type}
Question: {question}
Immutable reference answer: {reference}

Output schema:
{"candidates": [{"candidate_id": "V1", "answer": "short answer", "strategy": "strategy name", "evidence_description": "visible fact and location", "source_object": "source object or region", "relation_to_target": "why it belongs elsewhere", "quality": "high|medium|low"}], "no_candidate_reason": "reason or null"}
\end{Verbatim}
\end{CJK*}

\needspace{5\baselineskip}
\paragraph{\texttt{VISUAL\_DISTRACTOR\_CROP\_RULES}.}
\begin{CJK*}{UTF8}{gbsn}
\begin{Verbatim}[fontsize=\small,breaklines=true,breaksymbolleft={},breaksymbolright={},xleftmargin=0.6em,xrightmargin=0.6em]
Use only information visible in this expanded crop. Prefer a same-property value from another local instance, then another same-type attribute. Do not assume content from the unseen full image.
\end{Verbatim}
\end{CJK*}

\needspace{5\baselineskip}
\paragraph{\texttt{VISUAL\_DISTRACTOR\_FULL\_RULES}.}
\begin{CJK*}{UTF8}{gbsn}
\begin{Verbatim}[fontsize=\small,breaklines=true,breaksymbolleft={},breaksymbolright={},xleftmargin=0.6em,xrightmargin=0.6em]
Use only information visible in the native full image. Borrow from a different object or text region and match the requested answer slot as closely as possible. You do not have access to the zoom crop.
\end{Verbatim}
\end{CJK*}

\needspace{5\baselineskip}
\paragraph{\texttt{VLM\_DISTRACTOR\_VALIDATOR}.}
\begin{CJK*}{UTF8}{gbsn}
\begin{Verbatim}[fontsize=\small,breaklines=true,breaksymbolleft={},breaksymbolright={},xleftmargin=0.6em,xrightmargin=0.6em]
You are an independent conservative visual validator for proposed multiple-choice distractors. You see the same {view_name} image used by the generator. The immutable reference is correct. Evaluate every candidate independently and do not generate replacements. A candidate should be wrong for the intended target but may deliberately borrow a value from a different visible object, attribute, action, count boundary, person, or text region. Do not reject merely because the source is elsewhere.

Infer the exact answer slot and reference semantic core. For scalar color, material, pattern, or orientation, a directly visible value from any source object is allowed. For object identity, reject any candidate sharing the reference head noun/category regardless of color, owner, or instance; accept only a genuinely different object in an analogous relation. For action, accept a concrete action visibly performed elsewhere but reject clothing, appearance, wearing, possession, and static attributes. For counts, require a defensible visible group/boundary. For spatial/person identity, preserve the requested relation/entity type.

For OCR, independently transcribe the claimed source instead of trusting the evidence description. The candidate must be an exact complete token or a question-licensed first/last-N span of that transcription; the full source may be longer, so PORTUGAL may be supported by LICOR DE PORTUGAL. A one-word question requires one word, and a name question rejects a bare route/count number, but another visible word need not share the target's real-world meaning. Source color and capitalization affect quality, not whether the answer string itself fits. If the text is too small to read independently, use reject or uncertain.

Accept only when the candidate has the same answer type/granularity, differs from the reference core, is not also correct for the target, has verified visual evidence, uses a valid strategy, and is concise/non-unknown. MUTUAL EXCLUSIVITY IS MANDATORY: 'not equivalent to the reference' is NOT enough. Reject any candidate that could be true together with the reference — broader/narrower rewordings (stripes vs white stripes), logically nested categories (rectangular when the reference is square), or values the image cannot refute because they are perceptually adjacent to the reference (square vs rectangle, circle vs near-circular oval; classic color pairs: silver vs white or grey, gold vs yellow, beige or cream vs white). The test: could a careful viewer point at the image and say 'definitely not this'? If not, reject. Reject generic, inferred, likely, or world-knowledge evidence. decision=accept requires quality high or medium. Process every ID exactly once and output strict JSON only.

View: {view_name}
Question type: {question_type}
Question: {question}
Immutable reference answer: {reference}

Candidates:
{candidates}

Output schema:
{"answer_slot": "exact requested slot", "reference_semantic_core": "core meaning", "evaluations": [{"candidate_id": "V1", "decision": "accept|reject|uncertain", "slot_fit": true, "different_from_reference": true, "not_correct_for_target": "yes|no|unclear", "visual_source_verified": "yes|no|unclear|not_applicable", "strategy_valid": true, "ocr_length_constraint_satisfied": "yes|no|not_applicable", "independent_source_transcription": "exact text or null", "source_matches_candidate": "yes|no|unclear|not_applicable", "quality": "high|medium|low", "reason": "specific concise reason"}]}
\end{Verbatim}
\end{CJK*}

\needspace{5\baselineskip}
\paragraph{\texttt{DISTRACTOR\_ONLY\_REWRITE}.}
\begin{CJK*}{UTF8}{gbsn}
\begin{Verbatim}[fontsize=\small,breaklines=true,breaksymbolleft={},breaksymbolright={},xleftmargin=0.6em,xrightmargin=0.6em]
You are editing multiple-choice distractor options. Rewrite ONLY the listed distractors. The reference answer is immutable context: never rewrite it, replace it, include it in the output list, or turn any distractor into the reference answer. You do not see the image.

Question type: {question_type}
Question: {question}
Immutable reference answer: {reference}

Distractors to edit:
{distractors}

For each distractor:
1. Preserve exactly the answer it commits to. The rewrite must remain semantically interchangeable with that distractor for this question.
2. Make it a concise, natural, standalone answer to the question. Remove reasoning, hedging, redundant framing, and sentence-level filler.
3. Match the grammatical form and surface style of the immutable reference when useful, without changing the reference itself. For example, if the reference spells a count as a number word, spell distractor counts as number words too.
4. Never change a count value, time, OCR character sequence, proper name, direction, color, material, negation, object identity, action, or spatial relation. Formatting-only normalization is allowed when meaning is unchanged.
5. Do not add visual details, combine options, explain why an option is wrong, or mention the image or reference answer.
6. If a distractor is already concise, natural, and style-compatible, copy it unchanged.
7. Process every distractor ID exactly once. Do not add, drop, merge, or reorder IDs. Output strict JSON only.

Output schema:
{"distractors": [{"id": "D001", "rewritten_answer": "concise answer", "changed": true, "reason": "brief editing reason"}]}
\end{Verbatim}
\end{CJK*}

\needspace{5\baselineskip}
\paragraph{\texttt{MULTIPLE\_CHOICE\_POOL\_CURATOR}.}
\begin{CJK*}{UTF8}{gbsn}
\begin{Verbatim}[fontsize=\small,breaklines=true,breaksymbolleft={},breaksymbolright={},xleftmargin=0.6em,xrightmargin=0.6em]
You are the final text-only curator for a visual multiple-choice question. The immutable reference answer is correct. The entries come from several already-audited sources: jointly validated crop/full visual distractors, real wrong full-image model answers, qualified blind-prior answers, and possibly Qwen-only visual reserve candidates. You do not see the image. Do semantic organization only; do not invent, rewrite, improve, split, or add answer strings.

Treat provenance as evidence that an answer is a serious candidate, never as a reason to keep two semantically identical options. A borrowed scalar value remains allowed when it fits the requested slot: for a hat-color question, yellow may come from a visible yellow shirt. Infer the exact answer slot from the question and reference. Process every entry ID exactly once. Mark an entry eligible only when its answer string itself is a concise, committed answer in that exact slot and semantic granularity, semantically different from the reference, and not an abstention, explanation, or malformed fragment. Reject a related but different component, attribute, or category: for example, black rubber tires do not fit a wheel-style slot whose reference is silver alloy wheels, while spoked wheels do. Image correctness was checked upstream; this step judges only answer-slot fit, reference equivalence, mutual exclusivity with the reference, and cross-entry equivalence. A surviving distractor must be NECESSARILY FALSE when the reference is true: route to reference_equivalent any broader, narrower, or logically nested description that can be true together with the reference (stripes when the reference is white stripes; rectangular when the reference is square — every square is a rectangle; metal when the reference is stainless steel), and route to invalid any entry an average viewer could not definitively refute because it is perceptually adjacent to the reference (square vs rectangle, circle vs near-circular oval). Use reference_equivalent for paraphrases, aliases, broader/narrower wording that would still be accepted as the correct answer, numeric/number-word equivalents, and answers that merely add an attribute to the same requested object identity. Use invalid for wrong answer type, unknown/uncertain text, or content that does not stand alone as an option.

Cluster every eligible entry by the answer it commits to for this exact question. Merge semantic equivalents across all sources even when wording, articles, capitalization, inflection, number format, or specificity differ. Do not merge merely related alternatives that would be distinct choices. Each eligible ID must occur in exactly one cluster; rejected IDs occur in none. Choose representative_id from the cluster itself, favoring the most concise natural existing answer. Never output a new canonical answer.

Question type: {question_type}
Question: {question}
Immutable reference answer: {reference}

Pool entries:
{entries}

Output strict JSON only:
{"answer_slot": "exact requested slot", "entries": [{"id": "P001", "disposition": "eligible|reference_equivalent|invalid", "cluster_id": "C1 or null", "reason": "concise semantic reason"}], "clusters": [{"cluster_id": "C1", "member_ids": ["P001"], "representative_id": "P001", "relation_to_reference": "different", "answer_slot_fit": "valid", "reason": "shared committed answer"}]}
\end{Verbatim}
\end{CJK*}

\needspace{24\baselineskip}
\paragraph{\texttt{MULTIPLE\_CHOICE\_SOLVER}.}
\begin{CJK*}{UTF8}{gbsn}
\begin{Verbatim}[fontsize=\small,breaklines=true,breaksymbolleft={},breaksymbolright={},xleftmargin=0.6em,xrightmargin=0.6em]
Answer one multiple-choice visual question from the supplied zoom crop. The global question gives full-scene identity context; the crop question asks about the same target using wording appropriate for this crop. Use the crop question as the direct visual task. Inspect the image carefully and select exactly one displayed option. Do not explain, hedge, list multiple options, or use outside knowledge when the crop provides the answer.

Question type: {question_type}
Global question: {global_question}
Crop question: {crop_question}
Options:
{options}

Return strict JSON only, using one supplied choice label:
{"choice": "A"}
\end{Verbatim}
\end{CJK*}

\endgroup

\end{document}